\pdfoutput=1
\documentclass[11pt]{article}

\usepackage[]{acl}

\usepackage{times}
\usepackage{latexsym}
\usepackage{colortbl}
\usepackage[T1]{fontenc}
\usepackage[utf8]{inputenc}

\usepackage{microtype}

\usepackage{array}
\usepackage{enumitem}
\usepackage{float}
\usepackage{graphicx}
\usepackage{setspace}
\usepackage{bm}
\usepackage{amsfonts}
\usepackage{amssymb}
\usepackage{booktabs}
\usepackage{bbold}

\usepackage{multirow}

\usepackage{soul}
\graphicspath{ {./} }
\usepackage{url}
\usepackage{xcolor}
\usepackage{color}
\usepackage{amsmath}
\usepackage{dsfont}
\usepackage{enumitem}
\usepackage{courier}
\usepackage{kantlipsum}

\usepackage[most]{tcolorbox}

\usepackage{algorithm}
\usepackage{algorithmic}
\usepackage{float}  
\usepackage{lipsum}
\usepackage{hyperref}
\usepackage{inconsolata}
\newcommand{\inlineSubsection}[1]{
  \par\noindent\textbf{#1}\quad
}

\newcommand{\todo}[1]{\textcolor{red}{[#1]}}

\newcommand{\vc}[1]{\textcolor{black}{#1}}
\newcommand{\vic}[1]{\textcolor{blue}{#1}}
\newcommand{\vico}[1]{\textcolor{blue}{#1}}

\newcommand{\paperComment}[3]{{\color{#2}\par\noindent\fbox{\parbox{0.97\linewidth}{{\sf #1:} #3}}\newline}}

\newcommand{\victor}[1]{\paperComment{Victor}{purple!50!black}{#1}}
\newcommand{\jie}[1]{\paperComment{Jie}{purple!50!black}{#1}}

\newcommand{\simon}[1]{\textcolor{violet}{#1}}

\renewcommand{\vic}[1]{#1}
\renewcommand{\jie}[1]{}
\renewcommand{\victor}[1]{}

\renewcommand{\todo}[1]{}

\renewcommand{\simon}[1]{#1}
\renewcommand{\vico}[1]{#1}
\title{N2C2: Nearest Neighbor Enhanced Confidence Calibration for  Cross-Lingual In-Context Learning}

\author{
    Jie He$^1$ \quad 
    Simon Yu$^1$ \quad 
    Deyi Xiong$^3$ \quad \\ 
    \textbf{Víctor Gutiérrez Basulto}$^2$\footnotemark[2] \quad and \; 
    \textbf{Jeff Z. Pan}$^1$\thanks{ \, Corresponding author}  \\
     $^1$ School of Informatics, University of Edinburgh, UK  \\
    $^2$ School of Computer Science and Informatics, Cardiff University, UK \\
    $^3$ College of Intelligence and Computing, Tianjin University, Tianjin, China\\
    \normalsize{\texttt{ \{j.he, c.l.u, j.z.pan\}@ed.ac.uk, dyxiong@tju.edu.cn, }} \normalsize{\texttt{gutierrezbasultov@cardiff.ac.uk}} \\
}

\begin{document}
\maketitle
\begin{abstract}

Recent advancements of in-context learning (ICL) show language models can significantly improve their performance when demonstrations are provided. % compared to zero-shot learning. 
However, little attention has been paid to model calibration and prediction confidence of ICL in cross-lingual scenarios. 
%there has been limited exploration of ICL in cross-lingual scenarios; in particular, insufficient attention has been given to model calibration and prediction confidence. 
To bridge  this gap, we conduct a thorough analysis of ICL for cross-lingual sentiment classification. Our findings suggest  that ICL performs poorly in  cross-lingual scenarios, exhibiting low accuracy and presenting high calibration errors. In response, we propose a novel approach, \textbf{N2C2}, which employs a $k$-nearest neighbors augmented classifier for prediction confidence calibration. N2C2 narrows the prediction gap by leveraging a datastore of cached few-shot instances. Specifically, N2C2 integrates the predictions from the datastore and incorporates 
%retrieval representation construction\nb{retrieval representation construction?}, 
 confidence-aware distribution, semantically consistent retrieval representation, and adaptive neighbor combination modules to effectively utilize the limited number of supporting instances. Evaluation on two multilingual sentiment classification datasets demonstrates that N2C2 outperforms traditional ICL. It surpasses fine tuning, prompt tuning and recent state-of-the-art methods in terms of accuracy and calibration errors. 
\end{abstract}

%\victor{Check that uniformly we use $k$NN instead of $k$NN, as that seems to be the most commonly used because $k$ is a number. Maybe in the name of the model as well.}
\section{Introduction}
\vic{In-context learning (ICL) has significantly enhanced the performance of pre-trained language models (PLMs) on few-shot tasks~\cite{NEURIPS2020_1457c0d6,hu-etal-2022-context,lyu2023zicl}. 
% Particularly in low-resource settings, ICL has shown promise by delivering commendable results with minimal  data~\cite{gao-etal-2021-making,hu-etal-2022-context,lyu2023zicl}.
%Several recent works have explored the application of ICL in low-resource settings \cite{gao-etal-2021-making,hu-etal-2022-context,lyu2023zicl}.
For instance, in cross-lingual learning, high-resource languages (e.g.\ English) \vc{can be explored to address} %has proven effective in addressing the challenges posed by 
data-scarce challenges in low-resource languages~\cite{long-etal-2020-ted,long-etal-2020-shallow,kim2023boosting,tanwar2023multilingual,shi-etal-2022-xricl}.} 

\vic{Despite the progress made in ICL and its successful applications in cross-lingual  scenarios, \vc{there is a noticeable gap in understanding  the reliability of ICL methods in such tasks, i.e. how reliable their confidence predictions  are. Specifically, there is a gap in assessing the consistency between model confidence and accuracy across different multilingual models \cite{pmlr-v139-zhao21c,li-etal-2023-distinguishability}}. Previous works in multilingual ICL often construct prompt-contexts using randomly selected input-label pairs \cite{winata-etal-2021-language,lin-etal-2022-shot}.  %\nb{I  commented out some part that did not make too much sense}% with few attempts made to optimize this process \vc{to improve the model's reliability.} 
While some efforts  have been made to improve context selection in cross-lingual ICL \cite{tanwar2023multilingual}, the focus has predominantly been on performance enhancement rather than reliability. One exception is the study conducted by \citet{ahuja-etal-2022-calibration}, which investigates classical calibration techniques, such as temperature scaling, to mitigate calibration errors in multilingual classification tasks. However, that work concentrates on the calibration of fine-tuned multilingual models, \vc{yet it does not explore how this impacts the efficacy of cross-lingual ICL.} %yet it does not explore how this compares in efficacy to ICL methods.

% within a classification head 
%\nbsimon{It's unclear what does it mean CLS FT here (masked model FT?); and since later CLS is also used to represent dataset, might be better changed to something else?} 
% fine-tuning framework. 
%and \nb{say what is the part that they didnt do, but that we do. It is not clear. \jieh{J: They just investigated using temperature scaling to calibrate cls-based fine tuning, but what we do is calibrating icl and propose a retrieval-based method.} }
}
\begin{figure}[t]
    \begin{center}
        \includegraphics[width=0.45\textwidth]{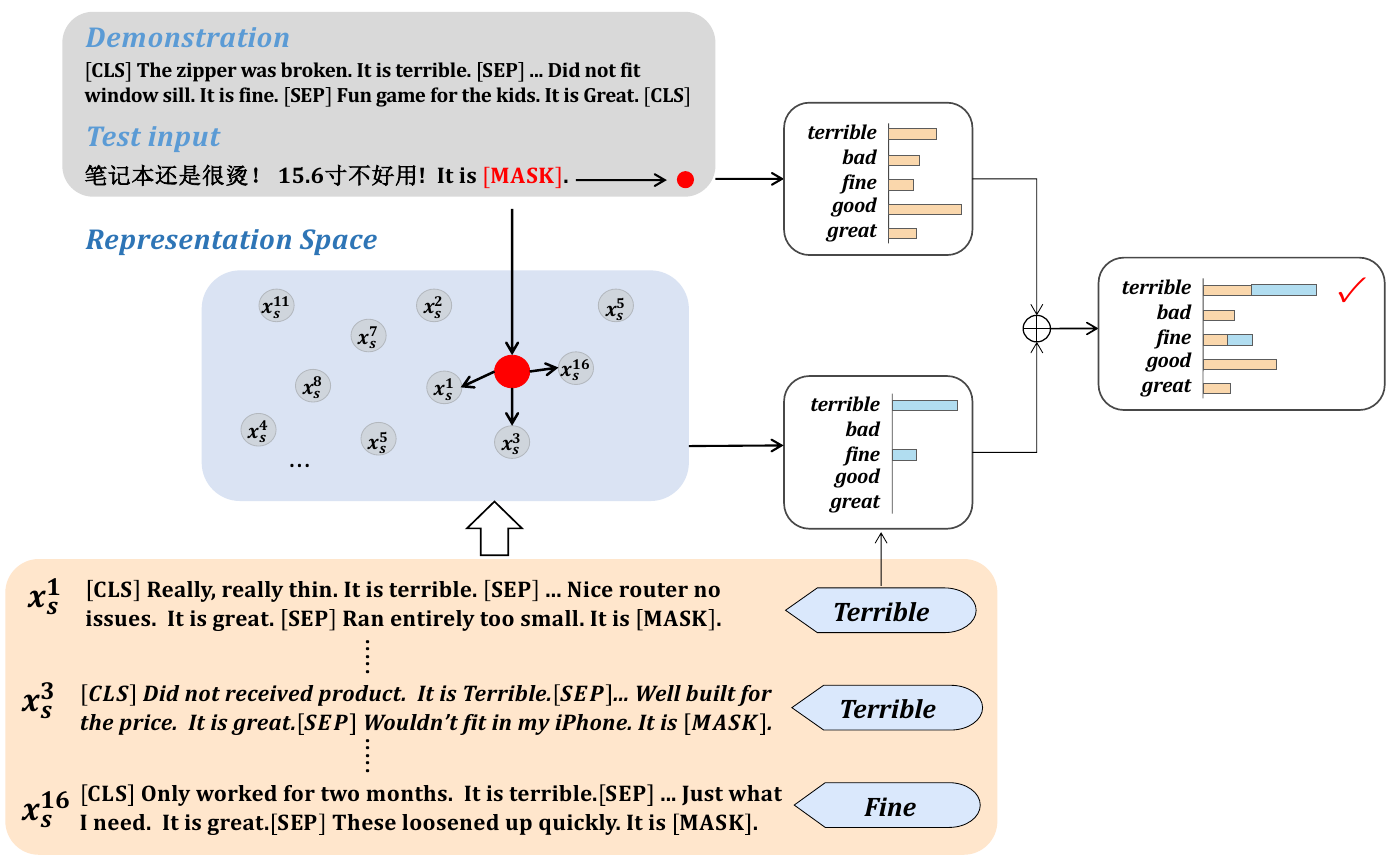}
    \end{center}
    \caption{Illustration of cross-lingual nearest neighbor inference with $k$ = 3, which makes
a prediction from 5 candidate words.}
    \label{fig1}
\vspace{-0.6cm}
\end{figure}
%Despite the progress made in in-context learning (ICL) and zero-shot cross-lingual applications, there has been insufficient consideration given to the reliability of ICL in zero-shot cross-lingual. I.e, How reliable are different multilingual models in zero-shot cross-lingual in-context learning?. Previous endeavors in multilingual ICL \cite{winata-etal-2021-language,lin-etal-2022-shot} have employed randomly selected input-label pairs for constructing prompt-context. \citet{tanwar2023multilingual} attempted to find better context for cross-lingual in-context learning, yet these approaches have primarily focused on enhancing performance while disregarding the reliability aspect of cross-lingual in-context learning.
%The sole work that has explored reliability in a cross-lingual setting is \cite{ahuja-etal-2022-calibration}. However, this work predominantly investigates the calibration of large-scale multilingual models within a fine-tuning framework and explores how classical calibration (e.g, temperature scaling) methods can mitigate calibration errors in a multilingual context. With the emergence of large language models (such as ChatGPT \cite{openai2023gpt4}) capable of handling multiple languages with a single model, the urgency for research on cross-lingual in-context learning has intensified.

\vic{To bridge this gap, we start by assessing the performance of ICL on \vc{ the cross-lingual sentiment classification (CLS) dataset \cite{prettenhofer-stein-2010-cross}.} %in a cross-lingual scenario. 
In this setting, the context is provided in English while  \vc{accuracy and calibration errors are evaluated on unseen languages.} To measure the calibration error, we leverage \vc{the \emph{expected calibration error} (ECE) \cite{10.5555/2888116.2888120}}, which measures the difference between confidence and accuracy,
thus evaluating the model's reliability in cross-lingual ICL  scenarios. Our study on \vc{CLS} %a multilingual dataset
reveals subpar performance achieved by conventional ICL in both accuracy and calibration errors. 

To tackle this challenge, we propose N2C2, a  method that %markedly 
enhances the accuracy of conventional ICL while %substantially 
reducing the expected calibration error. N2C2 identifies examples in the source language, which support predictions in the target language (Fig.\ \ref{fig1}) based on three criteria: (a) accurate retrieval, (b) robustness of retrieved examples with limited training data, and (c) preference for retrieved examples with higher confidence. 

To achieve criterion (a), N2C2 stores masked representations of each example in the source language, which are then transformed into \simon{retrieval-specific representations of lower dimensionality}. For criterion (b), we develop a dynamic weighting mechanism in a trained neural network to adjust the importance of rertrieved examples, rather than relying on a fixed top-$K$ retrieval approach. To satisfy criterion (c), we integrate confidence into the probabilities derived from  retrieval, \vc{rather than} solely relying on distances obtained from retrieval. We evaluate N2C2 on \vc{the} MARC and CLS datasets, with experimental results showcasing its superiority over \vc{existing} baselines across various settings.}

\vic{Our main contributions are as follows:}
%To summarize, the contributions of this paper can be outlined as follows:
\begin{itemize}[leftmargin=5mm]
\setlength{\itemsep}{0pt}
\setlength{\parsep}{0pt}
\setlength{\parskip}{0pt}
\item Our findings uncover significant performance gaps and calibration challenges in cross-lingual in-context learning, particularly in non-English languages. %(See section \ref{sec3})
\item We introduce N2C2, a method designed to retrieve supportive examples aiding predictions in cross-lingual contexts. To our knowledge, this is the first attempt to refine ICL calibration under cross-lingual settings through retrieval-based augmentation.
% \item We provide the first comprehensive analysis of the reliability of cross-lingual ICL, with a specific focus on expected calibration error.
\item Compared to strong baselines and classical calibration methods, extensive experiments on two multilingual sentiment classification datasets demonstrate that our method significantly reduces the expected calibration error while improving the accuracy of predictions. 

% We provide the first comprehensive analysis of the reliability of cross-lingual ICL, with a specific focus on expected calibration error. Through in-depth analysis, we reveal that in few-shot ICL cross-lingual scenarios, the multilingual model exhibits severe deficiencies in both prediction accuracy and calibration capability. Additionally, monolingual calibration methods are not applicable to cross-lingual scenario.
\end{itemize}

\section{Related Work}
%\subsection{Calibration}
\noindent \textbf{Calibration}
%
% \jieh{revise}
% \vic{\textcolor{orange}{%Accurate estimations are essential for challenging or sensitive prediction tasks \cite{PlattProbabilisticOutputs1999}. 
% In NLP, probability calibration particularly  plays a vital role in estimating uncertainty. 
% There exist various tasks that deal with uncertainty estimation in areas like out-of-domain detection or selective inference, but calibration specifically deals with measuring aleatory uncertainty through prediction probabilities and adjustment of the model's overall confidence level~\cite{hendrycks2017a,pereyra2017regularizing,pmlr-v70-guo17a,NEURIPS2021_78421a2e}.
Recent efforts have been made  on the calibration of pre-trained language models  \cite{pmlr-v97-hendrycks19a,desai-durrett-2020-calibration,jung-etal-2020-posterior,he-etal-2021-joint,long-webber-2022-facilitating,park-caragea-2022-calibration,bose-etal-2022-dynamically,chen2023close,long-etal-2024-multi}. Particularly relevant are investigations by \citet{ahuja-etal-2022-calibration, jiang-etal-2022-calibrating}, which explore the performance of various existing post-training calibration methods in cross-lingual classification and structure prediction tasks. Additionally, the study on contextual calibration by \citet{pmlr-v139-zhao21c} calibrates ICL predictions through bias probing and conditional prediction reversal. However, all these works do not evaluate multilingual models' calibration under the ICL setting.

\smallskip \noindent  \textbf{Multilingual Prompt Learning}
%Prompting in English has proven successful, but there is only limited work on its application in multilingual tasks. 
\citet{winata-etal-2021-language} demonstrate the multilingual capabilities of language models trained on English data by using a few English examples as context and evaluating their performance on non-English data. Recent studies optimize prompts for cross-lingual ICL with multilingual PLMs~\cite{huang-etal-2022-zero,zhao-schutze-2021-discrete,nie2023crosslingual,tanwar2023multilingual,long2024leveraginghierarchicalprototypesverbalizer}. However, their focus is on retrieving useful demonstrations from the source language and concatenate them with target examples to enhance cross-lingual ICL performance. In contrast, our approach retrieves and utilizes labeled training examples to aid predictions on target samples. %Moreover, our method does not require retrieval from a large-scale corpus.}

\smallskip \noindent \textbf{Retrieval in In-Context Learning}
Previous studies show the benefits of selecting demonstration examples closely resembling the test input, particularly when ample training data is available \cite{gao-etal-2021-making,liu-etal-2022-makes,liu2022semanticoriented}. \citet{liu-etal-2022-makes} retrieve the nearest training examples to the test input, employing unsupervised and supervised methods. \citet{shi-etal-2022-nearest} and \citet{lyu2023zicl} use nearest neighbor search to incorporate additional data for zero-shot inference by retrieving sentences closely related to the test input. Closer to our approach, \citet{xu2023kNN} and \citet{nie2022improving} enhance monolingual classification using $k$NN retrieval, but they do not address calibration in retrieval-augmented multilingual models. In contrast, our proposed retrieval-augmented approach is specifically designed for cross-lingual in-context learning.

\section{Background and Preliminary Experiments }
\label{sec3}
\inlineSubsection{Task Formulation}\vic{Our key interest is  cross-lingual ICL.
Let $s$ be a \vico{source} language, we use $X_s$ and $Y_s$ to respectively denote the sets of input examples and their corresponding labels in $s$.
We consider a monolingual labeled dataset $D_s = \{(x_i^s, y_i^s)\}_{i=1}^m$ with $m$ \vico{sampled examples}, where $x_i^s \in X_s$ and  $y_i^s \in Y_s$. 
\vico{Let $t$  be a target language, with $X_t$ as above. We consider the  set of sampled examples $D_t = \{x_j^t\}_{j=1}^n$,  with $n$ samples from $X_t$.}
%We also consider an additional set of input examples $D_t = \{x_j^t\}_{j=1}^n$ in language $t$ with $n$ samples. 
For cross-lingual ICL, we randomly choose \textbf{1}-shot per-class input-label pairs from $D_s$ as a prompt-context $C_s$: 
% $C_s = x_{1}^{s} \oplus y_{1}^{s} $
$C_{s} = \pi(x_{1}^{s}, y_{1}^{s})$
, where}
% \oplus [sep] \oplus \cdots x_{s}^{k} \oplus y_{s}^{k} $,}
% \subsection{Task Formulation}
% In this scenario, our focus is on zero-shot cross-lingual in-context learning (ICL). We consider a monolingual labeled dataset, denoted as $D_s = \{(x_s^i, y_s^i)\}_i$, which comprises input examples $x_s^i \in X_s$ and their corresponding labels $y_s^i \in Y_s$ in language $s$. Additionally, we have another collection of input examples, $D_t = \{x_t^i\}_i$, containing examples in language $t$. For zero-shot cross-lingual ICL, we randomly choose \textbf{1}-shot per-class input-label pairs from $D_s$ to form the prompt-context, $C$:
% \begin{equation*}
%     C = x_{s}^{1} \oplus y_{s}^{1} \oplus [sep] \oplus \cdots x_{s}^{k} \oplus y_{s}^{k} 
% \end{equation*}
%
% where the token $[sep]$ is used to separate tokens, such as newlines.
% $\oplus$ is the concatenation operator. 
$\pi$ is the prompt template.
The cross-lingual context $C_s$ is concatenated with the prompted input to form the input $I_j$ %\nb{V: I think this should be Ij, right? J: yes} 
for the multilingual PLM:%\nb{What is the first equation. Why is there a dot? Is this done for a fixed $ x_j^t$ of for all the $n$ ones \jieh{J: just add a prompt for test example. Like x: It was a great movie. P(x)= It was a great movie. It is [mask].}}  %for the MPLM.}
%\texttt{[CLS] [SEP] }
\begin{equation*}
    I_{j} = C_{s} \oplus  \pi(x_{j}^{t}, \texttt{[MASK]})
\end{equation*}
\begin{equation}
    \pi(x_{j}^{t}, \texttt{[MASK]}) =  x_j^t. \texttt{ It is [MASK]. }
    \label{instance}
\end{equation}
where $\oplus$ is the concatenation operator.
%The  token $[sep]$ is used to denote a separation between tokens, such as newlines. $\oplus$ is the concatenation operator.   Further, the cross-lingual context $C$ is concatenated with the prompted  input to form the input $I_i$. %for the MPLM.
% \begin{equation}
%     I_{i} = C_{s} \circ P(X_{t}^{i})
%     \label{vec}
% \end{equation}
\vic{The multilingual PLM is responsible of predicting masked tokens in the input $I_j$ and providing probability estimates $p$ for all possible candidate words. \vico{For  a candidate label $y$ over the label space $\mathcal Y$}, we determine the predicted class $\hat y$ by selecting the verbalizer $v(y)$ 
%\nb{What is $y$? Where does it come from? \jieh{y means the candidates label. Like y = [0,1] and verbalizer v(y)=[terrible, great]}}  
with the highest probability.}

%The multilingual PLM is responsible for predicting masked tokens in the input $I_i$ and providing probability estimates $p$ for all possible candidate words. We determine the predicted class $\hat y$ by selecting the verbalizer $v(\hat y)$ associated with the highest probability assigned by the Multilingual pre-trained language model.
% \begin{equation} --miss this equation
%     \hat{y} = arg \max_{y \in Y} p(v(y))
% \end{equation}
\begin{figure}[t]
    \begin{center}
        \includegraphics[width=0.38\textwidth]{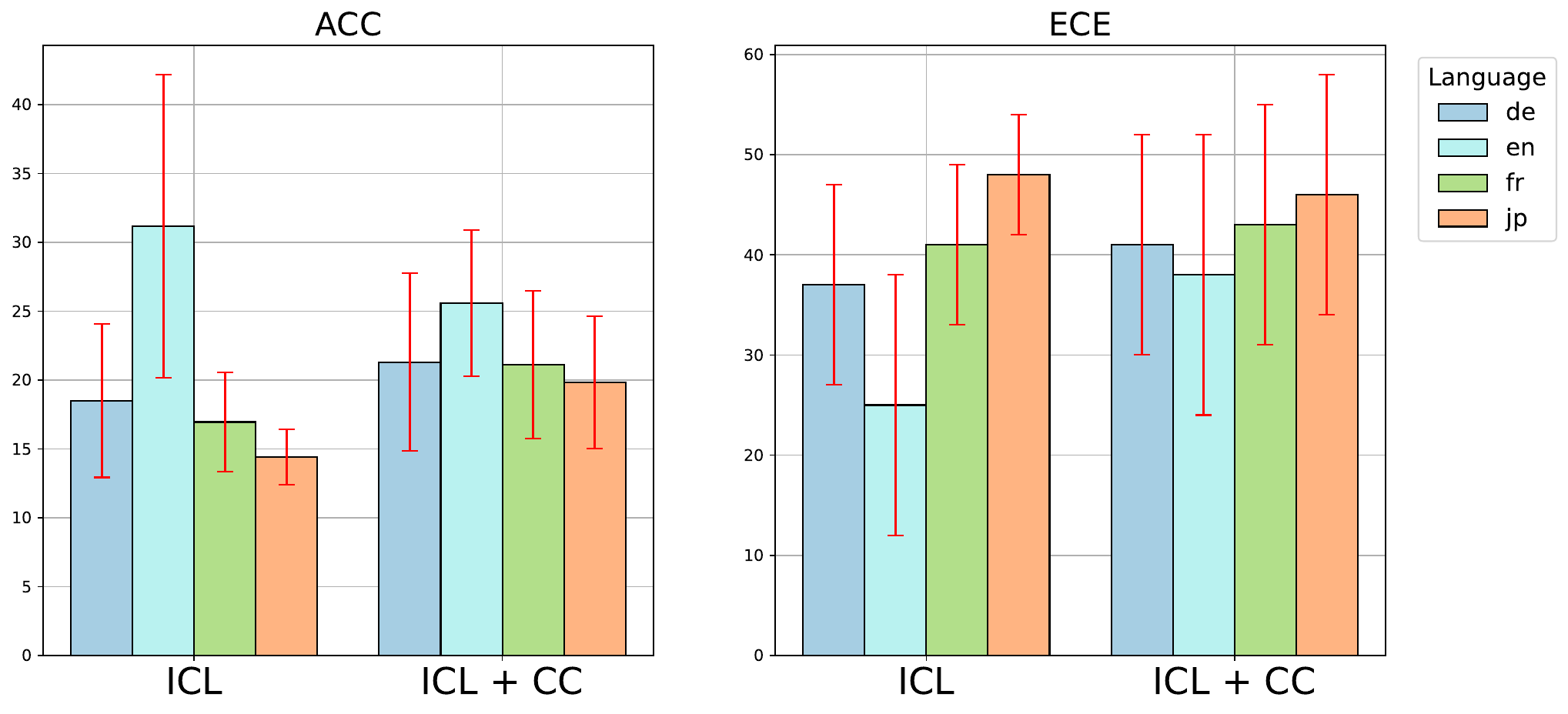}
    \end{center}
    \caption{Comparison the performance of the  cross-lingual ICL  and Contextual Calibration (ICL+CC)  across 4 languages with $\textbf{Accuracy}\uparrow(\%)$ (left) and $\textbf{ECE}\downarrow(\%)$ (right). }
    \label{conf_1}
\vspace{-0.5cm}
\end{figure}

\smallskip
\inlineSubsection{Ensemble-based Cross-lingual ICL}
\label{sec_ens}
\vic{We mainly use the XLM-RoBERTa (\simon{\text{XLM-R}}) model, which is limited by the input length, so we uniformly apply an ICL ensemble method across different shot settings~\cite{jiang-etal-2020-know}. This approach involves partitioning  $D_s$ into $m>0$  demonstration sets $D_s = D_1 \cup D_2 \cup ... \cup D_m$ to create different prompt contexts and combining the predictions from these $m$ prompts. Specifically, for calculating the prediction of the target example $x_j^t$, we compute  $    \frac{1}{m}\sum_{i=1}^{m} P(y|x_{i}^{s}, y_{i}^{s}, x_{j}^t)$}.
% As we mainly use the XLM-RoBERTa (\textit{XLM-R}) model, which is limited by input length, we apply the ICL ensemble method uniformly across different shot settings. This approach, previously employed in other studies \cite{jiang-etal-2020-know}, involves dividing $D_s$ into multiple non-overlapping demonstration sets, $D_s = D_1 \cup D_2 \cup ... \cup D_K$, to create different prompt contexts and combining the predictions from these $K$ prompts. More specifically, we compute  $    \frac{1}{K}\sum_{i=1}^{K} P(y|x_{s}^{i}, y_{s}^{i}, x^{(t)})$

\smallskip
\inlineSubsection{Calibration for  Cross-lingual ICL}
% \begin{equation}
%     \sum_{i=1}^{k} P(c|x_{i}^{(s)}, y_{i}^{(s)}, x^{(q)})
% \end{equation}
%
\vic{Calibration refers to the alignment between a model's assigned probability (\emph{confidence}) for a prediction and the true measure of its correctness (\emph{accuracy}) \cite{10.1145/1102351.1102430}. In other words,  given an input $x$, the ground truth $y$
and a prediction $\hat y$, \simon{the \emph{perfectly} calibrated confidence $\textrm{conf}(x, \hat y)$ will satisfy: $\forall p \in [0, 1], P(\hat y =
y ~|~ \textrm{conf}(x,\hat y) = p) = p$.}  The \emph{expected calibration error (ECE)} is a widely used metric to asses miscalibration that quantifies the difference between the \vico{expected confidence} %expectation of confidence 
and accuracy \cite{10.5555/2888116.2888120}. Please see  Appendix \ref{app_ece} for more details.}
%Calibration refers to the alignment between a model's assigned probability (confidence) for a prediction and the true measure of its correctness (accuracy) \cite{10.1145/1102351.1102430}. In other words,  given the input $x$, the ground truth $y$
%and the prediction $\hat y$, the perfectly calibrated confidence $\rm Conf(x, \hat y)$ will satisfy: $\forall p \in [0, 1], P(\hat y =
%y | \text{Conf}(x,\hat y) = p) = p$.
%To assess miscalibration, the Expected Calibration Error (ECE) is a widely used metric that quantifies the difference between the expectation of confidence and accuracy \cite{10.5555/2888116.2888120}. ECE achieves this by dividing predictions into $M$ bins, represented as $\{B_1, . . . , B_M\}$, based on their confidence values, and computing a weighted average of the accuracy-confidence difference within each bin:
%
\begin{figure*}[t]
    \begin{center}
\includegraphics[%width=0.7\textwidth
        width=13cm, height=7cm
        ]{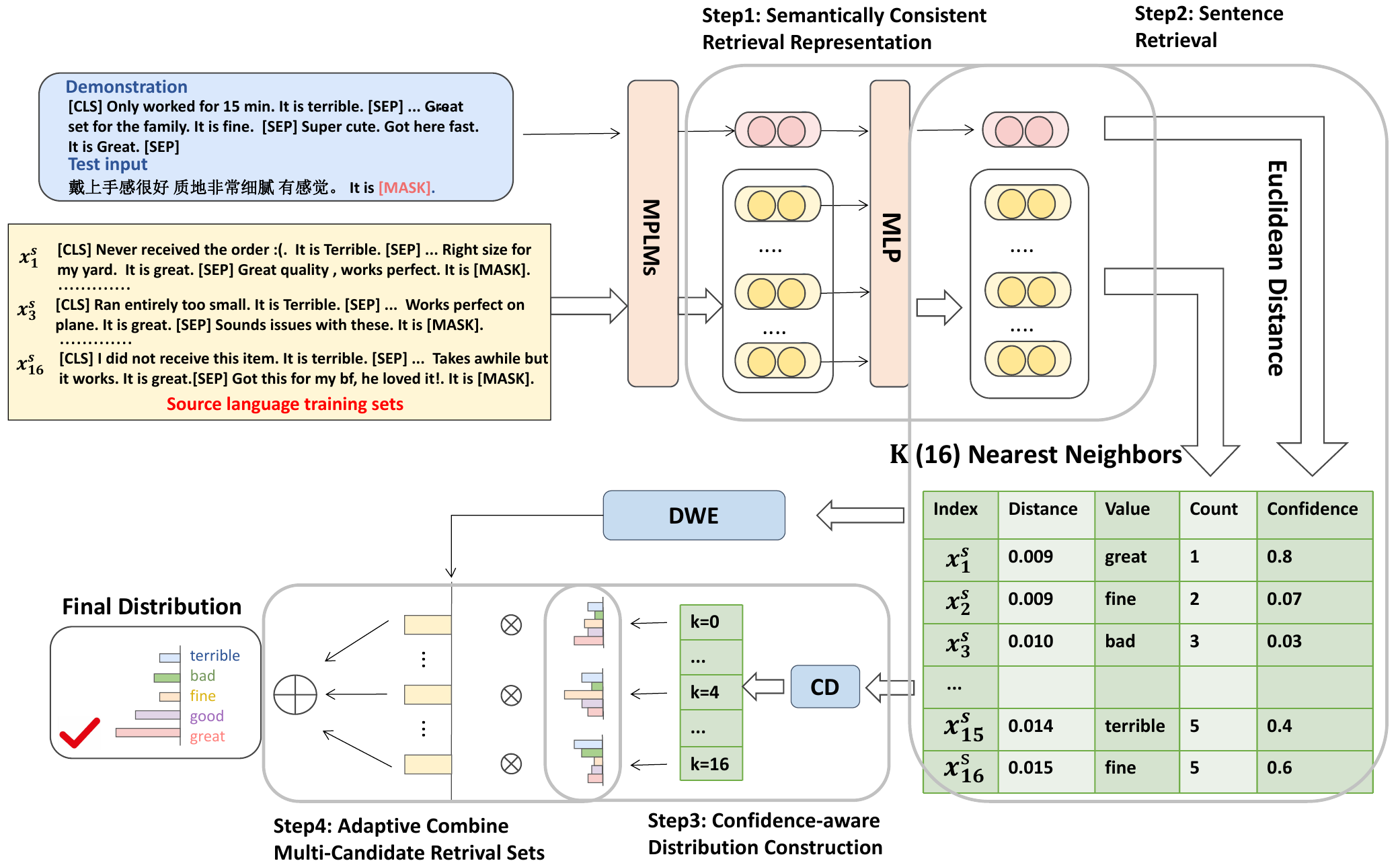}
    \end{center}
    \caption{Diagram  of  N2C2 with $k$ = 16. N2C2 first reconstructs $\text{h}_{\text{[mask]}}   (\S \ref{4.2})$ for the test example in the target language, and selects  neighbors
(\S \ref{4.1})  for it. It then consider confidence to generate multiple distributions (\S \ref{4.3}). These distributions are summed up together to form the final predicted distribution (\S \ref{adaptive combine}).
}
    \label{main}
\vspace{-0.5cm}
\end{figure*}

\vic{To assess the effectiveness of  cross-lingual ICL, we conducted preliminary experiments on the CLS dataset \cite{prettenhofer-stein-2010-cross}. These experiments involved utilizing 16 training examples per class, referred to as 16-shot. The XLMR-base model was employed as our language model. (\S \ref{4.2}). 
\simon{In the results of the preliminary experiments  shown in Figure \ref{conf_1}, while ICL  exhibits superior performance in English (with English contexts), its performance and calibration errors in other languages notably lag behind.}}
% Figure \ref{conf_1} shows that, unsurprisingly,  XLMR exhibits the highest performance on English (contexts are also in English). \simon{However, \textbf{the performance and calibration errors of XLMR in other languages are inferior.}}
%\nb{V: I removed the sentence about 30 \% as I found it uninformative and rather arbitrary. Ok? \jieh{ok}}  %In particular, ECE values surpass 30\%
%\nb{V: For English, for other languages? What 30 is telling us, good, bad? Why? \jieh{for other languages, I think for ECE, 30 is a high score, since good ece should be close to 0, no evidence show there is a boundary in ece for judging good/bad, just my thought.}}. 
% Additionally, there exists a positive relationship between accuracy and ECE \nbsimon{I can see the when $\uparrow$ accuracy, $\downarrow$ ECE. But Q1. can we say this as a "positive relationship"? Sounds a little bit ambiguous. Q2. what is the takeaway behind this finding?}. 
\simon{Moreover}, {we can see that applying \emph{contextual calibration} \cite{pmlr-v139-zhao21c} exacerbates the calibration performance \vico{for almost all languages}}. We hypothesize that this \vico{problem} %\nb{Challenge? Which challenge? \jieh{the problem that contextual calibration didt work}} 
 stems from the challenge of directly applying calibration methods that are effective in mitigating bias in monolingual scenarios to the context of cross-lingual ICL.
 \vic{To tackle these obstacles, we propose a retrieval-based approach, employing $k$-nearest neighbors ($k$NN), to enhance the performance of cross-lingual ICL and to alleviate its calibration error.}

%To assess the effectiveness of zero-shot cross-lingual ICL, we conducted initial experiments on the CLS dataset \cite{prettenhofer-stein-2010-cross}. These experiments involved utilizing 16 training examples per class ($K = 16$), referred to as 16-shot. The XLMR-base model was employed as our language model.
%The results, depicted in Figure \ref{conf_1}, indicate that the model exhibits the highest performance on English, which is expected as the training data is also in English. However, overall, the model's performance is unsatisfactory, particularly with ECE errors surpassing 30. Additionally, there exists a positive relationship between accuracy and ECE. Surprisingly, applying Contextual Calibration \cite{pmlr-v139-zhao21c} worsens the calibration performance. We hypothesize that this challenge stems from the difficulty of directly applying calibration methods, effective in reducing bias in monolingual scenarios, to the context of cross-lingual ICL.
% To address these concerns, we propose a retrieval-based approach employing k-nearest neighbors ($k$NN) to enhance the performance of cross-lingual ICL and mitigate its calibration error.
\section{Our Method: N2C2}
\textbf{Overview.} 
% We introduce a cross-lingual method that, given a test input, retrieves the top $K$ examples from the $b$-shot %\nb{I guess this k is not related to kNN or is it? If not use another letter - reviewers are annoying %\jieh{yes, I replace k with b and M with K}}
% source language. It leverages the label information from these examples to provide auxiliary predictions for the text input. 
Our method consists of \textbf{four steps}, cf.\ Figure~\ref{main}. 
\vic{\textbf{First}, the masked representations generated by multilingual pre-trained language models (MPLMs) are used to represent sentences after passing them through an MLP layer for \vico{semantically consistent} retrieval representation (\S \ref{4.2}). \textbf{Second}, we retrieve  a training set by finding  the $k$-nearest examples (\S \ref{4.1})}. \textbf{Third}, we assign higher weights to the examples with higher prediction confidence, to assist in the test input prediction (\S \ref{4.3}). \textbf{Finally}, to enhance the robustness of our method, instead of relying solely on a fixed number of top-$K$ predictions, we divide $K$ into $\{K_1, K_2, ..., K_l\}$ %\nb{This requires more detail. How do you divided it? Why is that intuitively better than just having the top-M \jieh{usually, people choose a fixed k, like 8, but we use top 4, top 8, top 12, top 16 (divisible by 4), we have multiple topk, and we can combine these results, I think it is better than a single topk}} 
and train a lightweight network to merge the results from these different $K_i$  (\S \ref{adaptive combine}). For training these lightweight networks, we split the training set into two equal parts: one  for retrieval and another one for updating the modules.

% \newtheorem{example}[]{}
% \begin{example}
We explain N2C2's steps through an example, following Fig.~\ref{main}. \textbf{Task}: Binary sentiment classification, labels 0, 1. \textbf{Input:} A target language test example $x^t$  and source examples $x_i^s$, $i \in [1,6]$
 with corresponding labels 0, 1, 1, 0, 1, 0. In Step (1), the MASK representation of the test example   is  transformed nonlinearly to obtain the representation $\mathbf{h}$.
 In Step (2), we calculate the similarities $s_i$, $i \in [1,6]$  between $\mathbf h$ and the representations $x_i^s$.
We will only use  the top 4 retrieved examples, so we only consider $s_1,s_2,s_3,s_4$ (which are the top 4 examples here)  in the next two steps. In Step (3), we consider the model's confidence in the predictions for these four retrieved examples together with their distance to refine the generated retrieval distribution: $distri_4 = [p_0,p_1]$, where $distri_4$ denotes the nearest top 4 retrieved examples and $p_i$, $i \in \{0,1\},$ a probability class. 
%using it as an influencing factor combined with  the distance in the generation of the retrieval distribution}\nb{V: after reading again, I realized I dont understand. "using it" what is "it".} $a_4 = [p_0,p_1]$\nb{What are $p_0,p_1$?}. 
In Step (4),  to enhance flexibility and robustness, we consider the top 4 examples (note that 4 is a hyperparemeter) and additionally the top 2. This yields answer distributions $dis_2$ and $dis_4$,  respectively. We additionally train aweighting module  to gauge the importance of different answer distributions. The final predicted answer distribution is obtained as $w_1 \times distri_2 + w_2 \times distri_4$.

% \end{example}

%\nb{V: Instead of saying ``better retrieval representation'', better say what is the effect of passing the sentences through and MLP layer, and why that is good

%\jieh{by passing a MLP layer, the representation can do more precise retrieval, it means it can retrieve a sentence that belongs to the same label rather than different label, since we trained this MLP layer using cross entropy loss. Maybe you can help think about writing some sentences to clarify that?}} 

%We propose N2C2, a novel zero-shot cross-lingual method. Given a test input, N2C2 retrieves the top $M$ examples from the k-shot source language, combines the label information from these examples, and provides auxiliary predictions for the text input.

%N2C2 consists of three steps (Figure 1): first, a pretrained forward network is utilized to retrieve sentences that are semantically similar to the test input (Section 4.2). Once we have a set of retrieved support sentences, we assign higher weights to the examples with higher prediction confidence to assist in predicting the test input (Section 4.3). To enhance the robustness of the method, instead of relying solely on a fixed number of top-$M$ predictions, we divide $M$ into $\{M_1, M_2, ..., M_n\}$ and train a network to merge the results from these different $M$s. For training these lightweight networks, we split the training set into two equal parts: one part is used for retrieval, while we update those modules on the other part.
\subsection{Inference with $k$NN Retrieval}
\label{4.1}
\vic{To do inference on a test example
%\nb{There is an inconsistency with the names. This was called before target example. I dont mind the inconsistency, but you know reviewers \jieh{what  names? if it is the instance, I changed it to examples.}} 
$x^t_j$, we use Equation \eqref{instance} to obtain the vector representation $\bm{h^t_j}$, corresponding to its \texttt{[MASK]} position. Subsequently, we derive the $k$NN-based prediction distribution $\bm{p_{k\text{NN}}}$ over the label space $\mathcal{Y}$ by considering the nearest neighbors:}
% To inference for a test instance $x$, we utilize equation \ref{vec} to obtain its corresponding vector representation $h_t$, which is then used for $k$NN retrieval. Subsequently, we derive the $k$NN-based prediction distribution $p_{$k$NN}$ over the label space $\mathcal{Y}$ by considering the nearest neighbors.
\begin{equation}
    p_{k\text{NN}}(y|x_{j}^{t}) \propto 
    \sum_{(h_{i}^s, y_{i}^s)\in N_{t}} \mathbb{1}_{y=y_{i}}  \text{exp}(\frac{-d(h_{i}^s,h_j^t )}{\tau})
\label{eq4}
\end{equation}
\vic{where $d$ denotes the Euclidean distance, $N_t$ denotes the set of $K$ nearest neighbors, and $\tau$ is a temperature parameter used to control the sharpness of the \emph{softmax} function. The final prediction distribution for label $y$ is obtained by interpolating two distributions, with the interpolation factor $\lambda$  being tuned within the range of [0, 1]:}
%The function $d(., .)$ represents the Euclidean distance, $N_t$ denotes the set of $M$ nearest neighbors, and $\tau$ is a temperature parameter used to control the sharpness of the softmax function. The final prediction distribution for label $y$ is obtained by interpolating two distributions, with the interpolation factor $\lambda$ tuned within the range of [0, 1].
\begin{equation*}
    p(y|x_{j}^{t}) = (1 - \lambda) \cdot p_{k\text{NN}}(y|x_{j}^{t})  
    + \lambda \cdot p_{\text{MPLM}}\left(y|x_{j}^{t}\right)
\label{eq5}
\end{equation*}
%\textcolor{red}{It should be noted that we leverage the development set to determine the suitable hyperparameters such as the interpolation factor lambda, the number of nearest neighbors k, and the temperature. I think we can delete this sentence.} 
\vic{Additionally, following the concept of ensemble ICL (\S \ref{sec_ens}), we aggregate the outcomes from various demonstrations for each test instance during the inference stage.}
\subsection{\vico{Semantically Consistent} Retrieval Representation}
\label{4.2} % $h_{\texttt{[mask]}}$
\vic{\vico{We  esttimate   the representation $\bm{h^t_j}$   of   example $x^t_j$ using Eq.\ \eqref{instance}}. 
%\nb{Is this the same $h^t_j$ as in section 4.1. I mean there is no $h^t_j$ in Eq 1. Now we are using again the term ``example'' \jieh{yes, no h in eq 1. but we got the h by pass I to MPLMs, do you have any suggestion about how to revise here?}}.
During the inference phase, for a test instance, we use its representations to retrieve the top $K$ instances from the source training language, along with their corresponding labels. \vc{However, to deal with a situation in which  retrieved instances  with similar representations  have different labels, we incorporate a straightforward linear layer that helps to distinguish such instance representations by leveraging the supervision provided by all training instances}: $h_{j}^{t'} = \mathbf{W}^{T} h_{j}^t + b$, } 
%Following the formula, we utilize $h_{\texttt{[mask]}}$ as a representation for an example. During the inference phase, for a test instance, we use its representations to retrieve in the source training language the top $M$ instances along with their corresponding labels. However, we may encounter a situation where the retrieved instances have different labels but exhibit similar representations. To address this issue, we incorporate a simple linear layer that helps to distinguish such instance representations by leveraging the supervision provided by all training instances. For each pair $(x_i, y_i)$ in all examples, we reconstruct the representation using the following approach:
% \begin{equation*}
%     h_{i}^{f} = g(\mathbf{W}_{f}h_{i} + b_{f})
% \end{equation*}
% %
\vic{where $\mathbf{W} \in \mathbb{R}^{H\times Z}$ and $b \in \mathbb{R}^{Z}$ are trainable parameters, with $Z$ representing the new dimension of the representation space. The purpose of this linear module is to ensure, by maximizing the $k$NN retrieval probability defined in Eq.\ \eqref{eq4}, that representations belonging to the same class are semantically similar. To achieve this, we optimize the linear layer by minimizing the cross-entropy loss associated with  Eq.\ \eqref{eq4}.}
%We introduce trainable parameters $\mathbf{W}_f \in \mathbb{R}^{H\times Z}$ and $b_f \in \mathbb{R}\times Z$, where $Z$ represents the new dimension of the representation space. In our experiments, we set $Z$ to 32. The purpose of the linear module is to ensure that representations belonging to the same class are semantically similar by maximizing the $k$NN retrieval probability defined in Equation \ref{eq4}. To achieve this, we optimize the linear layer by minimizing the cross-entropy loss of the Equation \ref{eq4}.
%\victor{Red text above, what is it about? ``Following formula'' Which formula? $h_{\texttt{[mask]}}$  does not seem to be ocurring anywhere else in the text.}
% \jie{can I cite equation 1?}
% \subsection{Confidence-based Distribution Reconstruction}
\subsection{Confidence-Aware Distribution Construction}
\label{4.3}
\vico{After computing the $k$NN distribution  with $\bm{h^{t'}_{j}}$} using Eq.\ \eqref{eq4}, we observe that the weights assigned to the retrieved examples are solely based on their distance to the query. However, this approach is suboptimal as it does not consider the confidence of the model. {To address this, we introduce the \emph{confidence-aware distribution (CD)} module, which assesses the importance of each retrieved pair $(x_i^s, y_i^s)$. This assessment is then used to refine the $k$NN distribution, ensuring that the probabilities are adjusted accordingly}. Specifically, the $k$NN distribution for example %\nb{Using back example \jieh{what do you mean here? don't understand}} 
$x^t_j$ is constructed through the following process:

%After calculating the $k$NN distribution using Equation \ref{eq4}, we noticed that the weights assigned to the retrieved examples are solely based on their distance from the query. However, we find this approach to be unreasonable because it does not take into account the confidence of the model. Instances with lower model confidence should be assigned lower probabilities. To address this, we introduce the Confidence-based Distribution (CD), which assesses the importance $c_i$ of each retrieved pair $(x_s^i, y_s^i)$. This assessment is then used to refine the $k$NN distribution, ensuring that the probabilities are adjusted accordingly. Specifically, the $k$NN distribution is constructed using the following process:
\begin{equation}
    p_{k\text{NN}}(y|x^{t}_j) \propto \sum_{(h_i, y_i) \in N_{t}} \mathbb{1}_{y=y_{i}}\text{exp}(\frac{-d(h_{i},h_j^{t'} )}{\tau\times T} + C)
\end{equation}
\begin{equation}
    T = \mathbf{W}_{1}(\text{tanh}(\mathbf{W}_{2}[d_{1}, \dots, d_{M}; o_{1}, \dots, o_{M}]))
    \label{eqconf}
\end{equation}
\begin{equation}
    C = \mathbf{W}_{3}(\text{tanh}(\mathbf{W}_{4}[p(y_{i}^s|x_{j}^t); p(y_{i}^s|x_{i}^s)]))
    \label{eq9}
\end{equation}
\vic{In the calculation of $C$  we consider two types of information: 1) $p(y_{i}^s| x^t_j)$, which denotes the predicted probability of $y_{i}^s$ by the multilingual PLM, given the mask representation $\bm{h_j^t}$, and 2) $p(y_{i}^s|x_{i}^s)$, which denotes the predicted probability of $y_{i}^s$ for the retrieved sample $x_i^s$. The variable $\bm{d_{i}}$ represents the L2 distance between the query $\bm{h^t_j}$ and the retrieved sample $\bm{h_i^s}$, while $\bm{o_{i}}$ denotes the number of unique values among the top $i$ neighbors. The $\mathbf{W}$s are the the parameter matrices.}

%In the calculation of $c_i$  we take into account two types of information: 1) $p(y_m|\hat h_t)$, which is the predicted probability of $y_k$ from the Multilingual pre-trained language model given the mask representation $\hat h_t$, and 2) $p(y_m|\hat h_s^i)$, which is the predicted probability of $y_m$ for the retrieved sample $h_s^i$. The value $d_m$ represents the L2 distance between the query $h_t$ and the retrieved sample $h_s^i$, while $r_m$ denotes the number of unique values among the top $M$ neighbors. The parameter matrices $\mathbf{W}$ are involved in these calculations.
\vspace{-0.2cm}
\subsection{Adaptively Combining Multi-Candidate Retrieval Sets}
\label{adaptive combine}
{Using a fixed value of $K$ in Eq.\ \eqref{eq4} can be problematic, especially when there are insufficient relevant items in the training sets. To enhance the robustness of N2C2, we adopt a technique proposed by \citet{zheng-etal-2021-adaptive}. This approach involves considering a range of $K$ values that are smaller than a predefined upper bound, $K_{max}$. Additionally, we introduce a lightweight network to assess the importance of using different selections, thereby minimizing the inclusion of irrelevant neighbors}. 
In practice, we simplify the choice of $K$ by opting for multiples of 4 (denoted as $R_{s}$), which includes $K = 0$ (corresponding to solely utilizing multilingual PLMs). 
%Concretely, we define $M$ as an element of the set $\mathcal{A}$, where $\mathcal{A} = {0} \cup {M_i \in N | M_i/4 \in N, M_i \leq M_{max}}$. 
The lightweight network evaluates the importance of  various $k$NN retrieval outcomes by using the retrieved neighbors as inputs.
%\victor{In principle, if there is an urgency for space, we could keep the paragraph above as intuition/summary in the body and refer the reader to appendix for details, i.e. the paragraphs that follow.}

%To address the potential issues caused by a fixed value of $M$ in Equation \ref{eq4}, especially when there are insufficient relevant items in the training sets, we aim to improve the robustness of our method by adopting a technique proposed by \citet{zheng-etal-2021-adaptive}. This involves considering a range of $M$ values that are smaller than a predefined upper bound, $M_{max}$. Additionally, we introduce a lightweight network to assess the importance of using different selections, thereby minimizing the inclusion of irrelevant neighbors. In practical terms, we simplify the choice of $M$ by selecting multiples of 4, including the option of $M = 0$, which corresponds to utilizing only multilingual pre-trained language models (MPLMs). Concretely, we define $M$ as an element of the set $\mathcal{A}$, where $\mathcal{A} = {0} \cup {M_i \in N | M_i/4 \in N, M_i \leq M_{max}}$. The lightweight network then evaluates the importance associated with various $k$NN retrieval outcomes by utilizing the retrieved neighbors as inputs.

\vic{More specifically, for a given test instance with a demonstration $x_j^t$, %\nb{Is this from the target language? Why i and t? What is the difference between demonstration and test instance? \jieh{ yes, i mean the i-th example, t mean target language. demonstration is like: it was a great movie. it is great. test example: it was a good night.}}, 
we begin by retrieving a maximum of $K_{max}$ neighbors from the source language $s$. Then, we compute the distances between these neighbors and the current representation, as well as the count of distinct values in the top $i$ neighbors denoted as $o_i$. The calculated distances $d = (d_1, ..., d_{K_{max}})$ and counts $c = (c_1, ..., c_{K_{max}})$ serve as inputs for determining the optimal value of $K$. The rationale behind considering the distances of each neighbor lies in the direct evidence they provide for assessing their importance. 

Furthermore, we incorporate the label counts $\bm{o_i}$, following Eq.\ \eqref{eqconf}. The distribution weight estimation network, denoted as \simon{\textbf{DWE($\cdot$)}}, comprises two feed-forward layers with a non-linear function in between. Specifically, we set the hidden size to 32. The probability of selecting a particular value of $K$ is calculated using the following formula: $ p(K|x_{j}^t) = \textit{softmax}(f_{\text{DWE}}([d,c]))$.}
%More specifically, for a given test instance with a demonstration $x^t$, we begin by retrieving a maximum of $M_{max}$ neighbors, $N_t$, from the datastore. We then compute the distances between these neighbors and the current representation, as well as the count of distinct values in the top $i$ neighbors, $c_i$. These computed distances, $d = (d_1, ..., d_{M_{max}})$, and counts, $c = (c_1, ..., c_{M_{max}})$, along with the corresponding labels $y = (y_1, ..., y_{M_{max}})$, are used as inputs to determine the optimal value of $M$. The rationale behind considering the distances of each neighbor is that it provides direct evidence when assessing their importance. Furthermore, we incorporate the label counts $c$ as same as in \ref{eq9} as model inputs to prioritize trust in multilingual pre-trained language models (MPLMs) when the retrieved labels exhibit disorder. The distribution importance estimation network, denoted as DIE(·), consists of two feed-forward layers with a non-linear function in between. Specifically, we set the hidden size to 32. The probability of selecting a particular value of $M$ is calculated using the following formula:
% \begin{equation}
%     p_{a}(M|x_{t}) = \text{softmax}(f_{\text{DIE}}([d,c]))
% \end{equation}
% \victor{Red text above, what does it mean for a neighbor to be top?}
% \jie{for example, the top 3 results are apple, apple, banana, $c_3$=2 (apple and banana), if top 5 results are apple, apple, banana, apple, orange, $c_5$=3 (apple and banana, orange)}
\vic{Instead of relying solely on the hyperparameter $\lambda$ as defined in Equation \ref{eq5}, we employ the importance estimation network to combine the outputs of PLMs and various $k$NN predictions as the final prediction:}

\begin{equation}
    p(y|x_{j}^t) = \sum_{M \in R_{s}} p(M|x^t_j) \cdot p_{\text{$k$NN}_{\text{M}}}(y|x^t_j)
\end{equation}

\section{Experiments}
%\subsection{Datasets}
\noindent \textbf{Datasets}
\vic{We evaluated N2C2 and baselines on two datasets: Multilingual Amazon Reviews Corpus (MARC) \cite{keung-etal-2020-multilingual} and Cross-language sentiment classification (CLS) \cite{prettenhofer-stein-2010-cross}. Details about the datasets and experimental setup can be found in App.~\ref{dataset}.}
% The MARC dataset is a large-scale multilingual text classification dataset from Amazon reviews (1-5). The CLS dataset consists of reviews on DVDs, music and books which are all multilingual and have four different ratings (1-4). }
% \vic{\textbf{English} is our source language, which is used for
% the training and development sets. All available languages  in the datasets are used for testing. 
% \victor{Is the same task for both datasets? As it is written gives the impression that for CLS one looks at another task}
% \simon{These two should be the same tasks, just that the number of class different, one is 5 and another is 4}

%We perform the experiment on two datasets, Multilingual Amazon Reviews Corpus (MARC) \cite{keung-etal-2020-multilingual} and Cross-language sentiment classification (CLS) \cite{prettenhofer-stein-2010-cross}. The MARC dataset is a large-scale multilingual text classification dataset from Amazon reviews, and the task is to predict the star rating given by the customers to the product based on their product review (from 1 to 5 stars, higher stars mean higher satisfaction) The CLS consists of reviews on DVD, music and books which are all multilingual and have four different ratings (1-4). Both datasets consist of reviews in German, English, French and Japanese, and the MARC dataset in addition consists of Spanish and Mandarin.

%In our $k$-shot experiments, where $k \in \{2, 4, 8, 16, 32\}$, we randomly sample $k$ cases from each category, which mean we would have $k \times n$ cases in total ($n = 5$ for MARC and $n = 4$ for CLS). 

\addtolength{\tabcolsep}{-1pt}
\begin{table*}[ht]
\centering
\tiny
\begin{tabular}{l@{\hspace{1.5\tabcolsep}}l@{\hspace{1.3\tabcolsep}}ccccccc}
\toprule
                              $b$ & \textbf{Lang} &                                                        \textbf{ICL} &                                                  \textbf{ICL + CC} &                                                         \textbf{FT} &                                                        \textbf{PT} & \textbf{X-InSTA} & \textbf{X-InSTA*} &                                                                                   \textbf{N2C2} \\
\midrule                              
\multicolumn{9}{c}{\textbf{MARC}} \\                              
\midrule
     \multirow{4}{*}{\textbf{2}}  &               De  &    $\text{28.06}_{\pm\text{3.2}}$ / $\text{28.00}_{\pm\text{8.0}}$  &   $\text{27.84}_{\pm\text{3.1}}$ / $\text{19.00}_{\pm\text{5.0}}$  &   $\text{21.28}_{\pm\text{0.4}}$ / $\text{53.92}_{\pm\text{18.4}}$  &   $\text{26.53}_{\pm\text{1.5}}$ / $\text{55.70}_{\pm\text{2.4}}$  &    20.36 / 71.65 &               27.84 / 80.54 &   $\text{\textbf{29.09}}_{\pm\text{\textbf{2.0}}}$ / $\text{\textbf{14.97}}_{\pm\text{\textbf{4.4}}}$ \\
                                  &               En  &    $\text{32.42}_{\pm\text{4.0}}$ / $\text{23.00}_{\pm\text{7.0}}$  &   $\text{31.16}_{\pm\text{3.8}}$ / $\text{22.00}_{\pm\text{5.0}}$  &   $\text{20.96}_{\pm\text{0.3}}$ / $\text{55.53}_{\pm\text{20.9}}$  &   $\text{31.70}_{\pm\text{1.9}}$ / $\text{56.72}_{\pm\text{2.0}}$  &    22.12 / 57.89 &               \textbf{43.66} / 69.51 &   $\text{32.56}_{\pm\text{2.7}}$ / $\text{\textbf{11.59}}_{\pm\text{\textbf{3.7}}}$ \\
                                  &               Zh  &    $\text{24.90}_{\pm\text{2.6}}$ / $\text{32.00}_{\pm\text{9.0}}$  &   $\text{25.57}_{\pm\text{2.2}}$ / $\text{22.00}_{\pm\text{3.0}}$  &   $\text{22.37}_{\pm\text{1.0}}$ / $\text{54.00}_{\pm\text{19.9}}$  &   $\text{22.06}_{\pm\text{0.9}}$ / $\text{59.60}_{\pm\text{2.7}}$  &    19.24 / 54.38 &               \textbf{36.14} / 39.79 &   $\text{26.79}_{\pm\text{1.8}}$ / $\text{\textbf{15.71}}_{\pm\text{\textbf{5.0}}}$ \\
                                  &          Avg.  &    $\text{27.16}_{\pm\text{2.7}}$ / $\text{30.00}_{\pm\text{4.4}}$  &   $\text{27.45}_{\pm\text{1.8}}$ / $\text{20.50}_{\pm\text{1.3}}$  &    $\text{21.37}_{\pm\text{0.7}}$ / $\text{54.57}_{\pm\text{2.6}}$  &   $\text{26.48}_{\pm\text{2.8}}$ / $\text{57.03}_{\pm\text{1.3}}$  &    20.51 / 60.82 &                \textbf{33.90} / 52.25 &   $\text{28.74}_{\pm\text{1.9}}$ / $\text{\textbf{15.57}}_{\pm\text{\textbf{2.7}}}$ \\
\midrule
     \multirow{4}{*}{\textbf{4}}  &               De  &    $\text{26.58}_{\pm\text{3.3}}$ / $\text{32.00}_{\pm\text{9.0}}$  &   $\text{28.64}_{\pm\text{2.5}}$ / $\text{22.00}_{\pm\text{6.0}}$  &   $\text{27.98}_{\pm\text{1.9}}$ / $\text{53.03}_{\pm\text{15.7}}$  &   $\text{26.57}_{\pm\text{1.5}}$ / $\text{62.82}_{\pm\text{6.1}}$  &    19.58 / 73.72 &               27.12 / 78.07 &    $\text{\textbf{34.13}}_{\pm\text{\textbf{3.0}}}$ / $\text{\textbf{7.94}}_{\pm\text{\textbf{2.8}}}$ \\
                                  &               En  &    $\text{31.58}_{\pm\text{4.8}}$ / $\text{25.00}_{\pm\text{8.0}}$  &   $\text{32.04}_{\pm\text{3.4}}$ / $\text{24.00}_{\pm\text{6.0}}$  &   $\text{26.15}_{\pm\text{1.4}}$ / $\text{51.00}_{\pm\text{23.7}}$  &   $\text{27.80}_{\pm\text{1.7}}$ / $\text{62.56}_{\pm\text{6.0}}$  &    22.24 / 56.39 &               \textbf{45.24} / 73.73 &    $\text{36.65}_{\pm\text{3.4}}$ / $\text{\textbf{5.73}}_{\pm\text{\textbf{2.4}}}$ \\
                                  &               Zh  &   $\text{23.33}_{\pm\text{2.3}}$ / $\text{36.00}_{\pm\text{10.0}}$  &   $\text{27.97}_{\pm\text{2.0}}$ / $\text{25.00}_{\pm\text{6.0}}$  &   $\text{22.45}_{\pm\text{1.3}}$ / $\text{48.96}_{\pm\text{24.3}}$  &   $\text{23.67}_{\pm\text{2.3}}$ / $\text{65.64}_{\pm\text{7.3}}$  &    20.48 / 50.64 &               \textbf{36.68} / 39.34 &    $\text{32.50}_{\pm\text{2.9}}$ / $\text{\textbf{7.80}}_{\pm\text{\textbf{3.1}}}$ \\
                                  &          Avg.  &    $\text{25.59}_{\pm\text{3.0}}$ / $\text{33.67}_{\pm\text{4.5}}$  &   $\text{29.32}_{\pm\text{1.3}}$ / $\text{22.67}_{\pm\text{1.6}}$  &    $\text{24.39}_{\pm\text{2.6}}$ / $\text{53.56}_{\pm\text{4.4}}$  &   $\text{25.17}_{\pm\text{1.6}}$ / $\text{64.32}_{\pm\text{1.2}}$  &    20.62 / 60.33 &               \textbf{34.07} / 52.51 &    $\text{33.31}_{\pm\text{1.7}}$ / $\text{\textbf{8.09}}_{\pm\text{\textbf{1.6}}}$ \\
                                  \midrule
     \multirow{4}{*}{\textbf{8}}  &               De  &   $\text{27.38}_{\pm\text{4.0}}$ / $\text{29.00}_{\pm\text{10.0}}$  &   $\text{28.91}_{\pm\text{2.4}}$ / $\text{23.00}_{\pm\text{8.0}}$  &   $\text{25.70}_{\pm\text{0.7}}$ / $\text{53.83}_{\pm\text{17.2}}$  &   $\text{31.04}_{\pm\text{1.9}}$ / $\text{62.19}_{\pm\text{5.5}}$  &    19.68 / 71.66 &               26.48 / 76.32 &   $\text{\textbf{35.08}}_{\pm\text{\textbf{1.8}}}$ / $\text{\textbf{15.07}}_{\pm\text{\textbf{3.2}}}$ \\
                                  &               En  &    $\text{32.37}_{\pm\text{4.4}}$ / $\text{22.00}_{\pm\text{8.0}}$  &   $\text{32.81}_{\pm\text{2.8}}$ / $\text{24.00}_{\pm\text{8.0}}$  &    $\text{30.62}_{\pm\text{1.2}}$ / $\text{57.99}_{\pm\text{7.8}}$  &   $\text{33.10}_{\pm\text{1.1}}$ / $\text{61.34}_{\pm\text{4.4}}$  &     22.20 / 57.01 &               \textbf{45.14} / 77.55 &    $\text{39.18}_{\pm\text{1.3}}$ / $\text{\textbf{9.87}}_{\pm\text{\textbf{2.2}}}$ \\
                                  &               Zh  &   $\text{23.82}_{\pm\text{2.8}}$ / $\text{34.00}_{\pm\text{11.0}}$  &   $\text{27.97}_{\pm\text{1.2}}$ / $\text{27.00}_{\pm\text{8.0}}$  &    $\text{26.53}_{\pm\text{0.8}}$ / $\text{59.77}_{\pm\text{6.7}}$  &   $\text{28.48}_{\pm\text{1.4}}$ / $\text{64.80}_{\pm\text{5.1}}$  &     22.30 / 52.66 &                \textbf{37.10} / 39.07 &   $\text{33.05}_{\pm\text{1.7}}$ / $\text{\textbf{13.69}}_{\pm\text{\textbf{2.6}}}$ \\
                                  &          Avg.  &    $\text{26.20}_{\pm\text{3.1}}$ / $\text{31.00}_{\pm\text{4.8}}$  &   $\text{29.61}_{\pm\text{1.5}}$ / $\text{23.33}_{\pm\text{2.0}}$  &    $\text{28.59}_{\pm\text{1.9}}$ / $\text{55.20}_{\pm\text{3.3}}$  &   $\text{30.18}_{\pm\text{1.7}}$ / $\text{63.25}_{\pm\text{1.3}}$  &    20.91 / 60.33 &               33.82 / 52.54 &   $\text{\textbf{34.36}}_{\pm\text{\textbf{2.4}}}$ / $\text{\textbf{15.09}}_{\pm\text{\textbf{2.9}}}$ \\
                                  \midrule
    \multirow{4}{*}{\textbf{16}}  &               De  &   $\text{27.36}_{\pm\text{4.0}}$ / $\text{27.00}_{\pm\text{10.0}}$  &   $\text{30.61}_{\pm\text{3.3}}$ / $\text{21.00}_{\pm\text{7.0}}$  &   $\text{36.92}_{\pm\text{2.6}}$ / $\text{48.69}_{\pm\text{17.4}}$  &   $\text{36.46}_{\pm\text{2.0}}$ / $\text{56.39}_{\pm\text{4.8}}$  &    19.68 / 64.07 &               25.62 / 76.07 &   $\text{\textbf{37.75}}_{\pm\text{\textbf{2.3}}}$ / $\text{\textbf{10.94}}_{\pm\text{\textbf{5.9}}}$ \\
                                  &               En  &    $\text{32.20}_{\pm\text{4.3}}$ / $\text{21.00}_{\pm\text{7.0}}$  &   $\text{32.71}_{\pm\text{2.6}}$ / $\text{17.00}_{\pm\text{5.0}}$  &   $\text{28.42}_{\pm\text{1.0}}$ / $\text{50.99}_{\pm\text{25.4}}$  &   $\text{37.00}_{\pm\text{1.5}}$ / $\text{56.78}_{\pm\text{3.9}}$  &    21.34 / 52.38 &                \textbf{44.00} / 76.48 &    $\text{42.85}_{\pm\text{1.4}}$ / $\text{\textbf{5.69}}_{\pm\text{\textbf{2.7}}}$ \\
                                  &               Zh  &   $\text{23.64}_{\pm\text{2.5}}$ / $\text{31.00}_{\pm\text{10.0}}$  &   $\text{28.28}_{\pm\text{1.8}}$ / $\text{25.00}_{\pm\text{7.0}}$  &    $\text{32.53}_{\pm\text{1.8}}$ / $\text{57.10}_{\pm\text{6.5}}$  &   $\text{31.82}_{\pm\text{2.3}}$ / $\text{59.84}_{\pm\text{5.7}}$  &    23.16 / 55.45 &               \textbf{36.96} / 38.88 &    $\text{35.10}_{\pm\text{\textbf{2.1}}}$ / $\text{\textbf{9.49}}_{\pm\text{\textbf{4.5}}}$ \\
                                  &          Avg.  &    $\text{26.16}_{\pm\text{3.1}}$ / $\text{28.67}_{\pm\text{4.1}}$  &   $\text{30.23}_{\pm\text{1.4}}$ / $\text{20.17}_{\pm\text{2.6}}$  &    $\text{31.78}_{\pm\text{2.7}}$ / $\text{53.05}_{\pm\text{3.2}}$  &   $\text{33.79}_{\pm\text{2.2}}$ / $\text{57.94}_{\pm\text{1.3}}$  &    20.91 / 58.94 &               33.05 / 51.66 &   $\text{\textbf{37.07}}_{\pm\text{\textbf{3.2}}}$ / $\text{\textbf{10.89}}_{\pm\text{\textbf{3.5}}}$ \\
                                  \midrule
    \multirow{4 }{*}{\textbf{32}}  &               De  &   $\text{27.55}_{\pm\text{4.0}}$ / $\text{19.00}_{\pm\text{10.0}}$  &   $\text{22.53}_{\pm\text{3.2}}$ / $\text{21.00}_{\pm\text{6.0}}$  &    $\text{38.78}_{\pm\text{1.5}}$ / $\text{55.72}_{\pm\text{2.7}}$  &   $\text{39.16}_{\pm\text{2.4}}$ / $\text{54.94}_{\pm\text{3.2}}$  &    19.92 / 60.24 &                24.54 / 75.80 &    $\text{\textbf{41.59}}_{\pm\text{\textbf{1.8}}}$ / $\text{\textbf{4.35}}_{\pm\text{\textbf{2.2}}}$ \\
                                  &               En  &    $\text{32.15}_{\pm\text{4.5}}$ / $\text{19.00}_{\pm\text{8.0}}$  &   $\text{22.57}_{\pm\text{2.4}}$ / $\text{18.00}_{\pm\text{4.0}}$  &    $\text{39.31}_{\pm\text{1.0}}$ / $\text{54.45}_{\pm\text{4.3}}$  &   $\text{40.51}_{\pm\text{1.9}}$ / $\text{54.56}_{\pm\text{2.5}}$  &    19.92 / 60.24 &               43.42 / 75.22 &    $\text{\textbf{44.79}}_{\pm\text{\textbf{1.2}}}$ / $\text{\textbf{2.72}}_{\pm\text{\textbf{1.3}}}$ \\
                                  &               Zh  &    $\text{23.81}_{\pm\text{2.6}}$ / $\text{21.00}_{\pm\text{9.0}}$  &   $\text{21.31}_{\pm\text{1.9}}$ / $\text{25.00}_{\pm\text{7.0}}$  &    $\text{36.99}_{\pm\text{2.1}}$ / $\text{53.03}_{\pm\text{6.4}}$  &   $\text{35.59}_{\pm\text{2.6}}$ / $\text{58.23}_{\pm\text{3.7}}$  &    18.32 / 69.26 &               37.06 / 39.02 &    $\text{\textbf{39.87}}_{\pm\text{\textbf{2.1}}}$ / $\text{\textbf{4.12}}_{\pm\text{\textbf{2.6}}}$ \\
                                  &          Avg.  &    $\text{26.20}_{\pm\text{3.0}}$ / $\text{19.00}_{\pm\text{1.5}}$  &   $\text{22.38}_{\pm\text{0.6}}$ / $\text{20.17}_{\pm\text{2.4}}$  &    $\text{36.90}_{\pm\text{1.7}}$ / $\text{54.08}_{\pm\text{1.8}}$  &   $\text{37.50}_{\pm\text{1.8}}$ / $\text{55.84}_{\pm\text{1.3}}$  &    22.04 / 49.73 &               32.57 / 51.16 &    $\text{\textbf{40.72}}_{\pm\text{\textbf{2.3}}}$ / $\text{\textbf{4.57}}_{\pm\text{\textbf{1.0}}}$ \\
\midrule
\multicolumn{9}{c}{\textbf{CLS}} \\
\midrule
\multirow{3}{*}{\textbf{2}}                                 &               En  &   $\text{29.55}_{\pm\text{8.2}}$ / $\text{27.00}_{\pm\text{10.0}}$  &   $\text{30.23}_{\pm\text{7.4}}$ / $\text{37.00}_{\pm\text{11.0}}$  &    $\text{25.03}_{\pm\text{6.6}}$ / $\text{40.57}_{\pm\text{21.7}}$  &    $\text{33.18}_{\pm\text{8.2}}$ / $\text{62.13}_{\pm\text{11.7}}$  &     15.40 / 49.19 &               \textbf{52.15} / 73.01 &                     $\text{30.46}_{\pm\text{5.3}}$ / $\text{\textbf{18.50}}_{\pm\text{\textbf{5.9}}}$ \\
                                 &               Fr  &    $\text{18.41}_{\pm\text{3.6}}$ / $\text{40.00}_{\pm\text{7.0}}$  &    $\text{22.49}_{\pm\text{5.2}}$ / $\text{43.00}_{\pm\text{7.0}}$  &    $\text{27.53}_{\pm\text{5.2}}$ / $\text{40.99}_{\pm\text{24.4}}$  &    $\text{31.85}_{\pm\text{9.2}}$ / $\text{62.41}_{\pm\text{14.2}}$  &     \textbf{34.20} / 50.61 &                33.10 / 44.36 &                     $\text{20.51}_{\pm\text{3.1}}$ / $\text{\textbf{29.43}}_{\pm\text{\textbf{6.2}}}$ \\
                                 &          Avg.  &    $\text{21.72}_{\pm\text{4.6}}$ / $\text{36.25}_{\pm\text{6.1}}$  &    $\text{24.20}_{\pm\text{3.5}}$ / $\text{42.75}_{\pm\text{3.9}}$  &     $\text{24.85}_{\pm\text{2.1}}$ / $\text{42.15}_{\pm\text{1.8}}$  &     $\text{31.83}_{\pm\text{1.2}}$ / $\text{62.49}_{\pm\text{1.3}}$  &    24.72 / 54.64 &               \textbf{38.12} / 53.82 &                     $\text{22.96}_{\pm\text{4.3}}$ / $\text{\textbf{26.66}}_{\pm\text{\textbf{4.7}}}$ \\
                                 \midrule
\multirow{3}{*}{\textbf{4}}                                 &               En  &    $\text{29.60}_{\pm\text{7.6}}$ / $\text{22.00}_{\pm\text{9.0}}$  &   $\text{27.10}_{\pm\text{4.0}}$ / $\text{38.00}_{\pm\text{13.0}}$  &    $\text{40.85}_{\pm\text{2.3}}$ / $\text{40.72}_{\pm\text{12.0}}$  &     $\text{37.59}_{\pm\text{4.8}}$ / $\text{56.13}_{\pm\text{6.5}}$  &     15.25 / 47.5 &                \textbf{54.05} / 75.57 &                     $\text{35.34}_{\pm\text{4.2}}$ / $\text{\textbf{15.88}}_{\pm\text{\textbf{5.8}}}$ \\
                                 &               Fr  &    $\text{21.78}_{\pm\text{3.0}}$ / $\text{26.00}_{\pm\text{6.0}}$  &   $\text{24.02}_{\pm\text{3.4}}$ / $\text{40.00}_{\pm\text{13.0}}$  &     $\text{41.18}_{\pm\text{2.8}}$ / $\text{46.07}_{\pm\text{7.5}}$  &     $\textbf{39.44}_{\pm\textbf{4.7}}$ / $\text{52.89}_{\pm\text{6.5}}$  &     34.90 / 53.75 &                34.85 / 42.69 &                     $\text{27.29}_{\pm\text{2.0}}$ / $\text{\textbf{22.55}}_{\pm\text{\textbf{2.6}}}$ \\
                                 &          Avg.  &    $\text{23.50}_{\pm\text{3.6}}$ / $\text{24.75}_{\pm\text{3.0}}$  &    $\text{24.50}_{\pm\text{1.6}}$ / $\text{40.50}_{\pm\text{1.8}}$  &     $\text{39.24}_{\pm\text{1.9}}$ / $\text{44.82}_{\pm\text{2.5}}$  &     $\text{38.52}_{\pm\text{1.0}}$ / $\text{53.98}_{\pm\text{1.3}}$  &    24.91 / 56.98 &               \textbf{39.11} / 53.56 &                     $\text{28.68}_{\pm\text{4.0}}$ / $\text{\textbf{20.84}}_{\pm\text{\textbf{3.0}}}$ \\
                                                                                                   \midrule

\multirow{3}{*}{\textbf{8}}                                 &               En  &    $\text{27.57}_{\pm\text{7.1}}$ / $\text{17.00}_{\pm\text{8.0}}$  &   $\text{27.23}_{\pm\text{5.1}}$ / $\text{39.00}_{\pm\text{18.0}}$  &     $\text{37.24}_{\pm\text{2.9}}$ / $\text{38.95}_{\pm\text{9.5}}$  &         $\text{38.69}_{\pm\text{2.3}}$ / $\text{48.51}_{\pm\text{1.4}}$  &    15.15 / 46.36 &                 \textbf{52.60} / 76.30 &                  $\text{42.85}_{\pm\text{4.7}}$ / $\text{\textbf{9.30}}_{\pm\text{\textbf{4.1}}}$ \\
                                 &               Fr  &    $\text{18.53}_{\pm\text{3.3}}$ / $\text{26.00}_{\pm\text{5.0}}$  &   $\text{24.20}_{\pm\text{6.0}}$ / $\text{44.00}_{\pm\text{18.0}}$  &    $\text{34.69}_{\pm\text{2.0}}$ / $\text{44.83}_{\pm\text{19.4}}$  &         $\text{33.09}_{\pm\text{2.7}}$ / $\text{49.12}_{\pm\text{2.5}}$  &    35.25 / 55.33 &                \textbf{35.90} / 42.65 &                 $\text{35.58}_{\pm\text{3.3}}$ / $\text{\textbf{13.22}}_{\pm\text{\textbf{4.3}}}$ \\

                                 &          Avg.  &    $\text{21.54}_{\pm\text{3.7}}$ / $\text{23.00}_{\pm\text{4.3}}$  &    $\text{25.56}_{\pm\text{1.1}}$ / $\text{42.00}_{\pm\text{2.1}}$  &     $\text{31.57}_{\pm\text{4.6}}$ / $\text{41.94}_{\pm\text{2.6}}$  &         $\text{33.04}_{\pm\text{1.6}}$ / $\text{49.76}_{\pm\text{1.2}}$  &     24.35 / 57.50 &               \textbf{38.75} / 53.02 &                 $\text{36.78}_{\pm\text{3.7}}$ / $\text{\textbf{12.46}}_{\pm\text{\textbf{1.9}}}$ \\
                                                                                                   \midrule
\multirow{3}{*}{\textbf{16}}                                 &               En  &   $\text{27.32}_{\pm\text{7.2}}$ / $\text{31.00}_{\pm\text{10.0}}$  &   $\text{25.10}_{\pm\text{4.3}}$ / $\text{39.00}_{\pm\text{10.0}}$  &     $\text{42.03}_{\pm\text{2.0}}$ / $\text{50.20}_{\pm\text{4.6}}$  &         $\text{44.83}_{\pm\text{4.7}}$ / $\text{51.44}_{\pm\text{5.0}}$  &    15.15 / 46.41 &               \textbf{52.55} / 74.96 &    $\text{\text{47.64}}_{\pm\text{\text{2.5}}}$ / $\text{\textbf{6.56}}_{\pm\text{\textbf{2.6}}}$ \\
                                 &               Fr  &    $\text{18.67}_{\pm\text{3.2}}$ / $\text{40.00}_{\pm\text{7.0}}$  &   $\text{21.80}_{\pm\text{5.0}}$ / $\text{43.00}_{\pm\text{11.0}}$  &   $\text{39.31}_{\pm\text{10.1}}$ / $\text{49.11}_{\pm\text{22.9}}$  &         $\text{38.48}_{\pm\text{4.4}}$ / $\text{57.19}_{\pm\text{5.1}}$  &      35.60 / 55.20 &                34.90 / 42.62 &    $\text{\textbf{41.64}}_{\pm\text{\textbf{5.0}}}$ / $\text{\textbf{8.00}}_{\pm\text{\textbf{4.6}}}$ \\
                                 &          Avg.  &    $\text{21.37}_{\pm\text{3.5}}$ / $\text{37.25}_{\pm\text{4.6}}$  &    $\text{22.65}_{\pm\text{1.5}}$ / $\text{41.75}_{\pm\text{1.6}}$  &    $\text{36.00}_{\pm\text{5.4}}$ / $\text{37.48}_{\pm\text{17.6}}$  &         $\text{40.85}_{\pm\text{2.4}}$ / $\text{54.06}_{\pm\text{2.3}}$  &    24.92 / 57.52 &                38.56 / 52.60 &    $\text{\textbf{43.99}}_{\pm\text{\textbf{3.0}}}$ / $\text{\textbf{7.61}}_{\pm\text{\textbf{0.6}}}$ \\
                                 \midrule
\multirow{3}{*}{\textbf{32}}                                 &               En  &    $\text{26.69}_{\pm\text{6.9}}$ / $\text{28.00}_{\pm\text{9.0}}$  &    $\text{22.53}_{\pm\text{4.3}}$ / $\text{31.00}_{\pm\text{6.0}}$  &    $\text{51.27}_{\pm\text{5.5}}$ / $\text{36.07}_{\pm\text{11.2}}$  &         $\text{48.98}_{\pm\text{2.6}}$ / $\text{46.62}_{\pm\text{3.3}}$  &    15.15 / 45.12 &               51.45 / 74.91 &    $\text{\textbf{55.63}}_{\pm\text{\textbf{2.0}}}$ / $\text{\textbf{4.50}}_{\pm\text{\textbf{1.3}}}$ \\
                                 &               Fr  &    $\text{18.67}_{\pm\text{3.3}}$ / $\text{37.00}_{\pm\text{7.0}}$  &    $\text{20.24}_{\pm\text{3.9}}$ / $\text{32.00}_{\pm\text{7.0}}$  &    $\text{27.04}_{\pm\text{14.6}}$ / $\text{14.07}_{\pm\text{2.6}}$  &         $\text{40.54}_{\pm\text{3.3}}$ / $\text{51.42}_{\pm\text{4.1}}$  &    35.65 / 55.61 &                35.30 / 40.61 &    $\text{\textbf{43.73}}_{\pm\text{\textbf{3.7}}}$ / $\text{\textbf{4.83}}_{\pm\text{\textbf{3.4}}}$ \\
                                 &          Avg.  &    $\text{21.25}_{\pm\text{3.2}}$ / $\text{34.00}_{\pm\text{4.3}}$  &    $\text{20.92}_{\pm\text{1.3}}$ / $\text{31.75}_{\pm\text{0.8}}$  &    $\text{39.95}_{\pm\text{8.9}}$ / $\text{35.38}_{\pm\text{13.1}}$  &         $\text{44.42}_{\pm\text{3.1}}$ / $\text{49.33}_{\pm\text{1.8}}$  &    24.48 / 57.32 &               38.29 / 51.86 &    $\text{\textbf{47.27}}_{\pm\text{\textbf{4.9}}}$ / $\text{\textbf{5.31}}_{\pm\text{\textbf{0.7}}}$ \\
\bottomrule
\end{tabular}

   \caption{Main results for the baselines and our method, reported as $\textbf{Accuracy}\uparrow(\%)$ / $\textbf{ECE}\downarrow(\%)$. $b$ is the number of training samples per class (i.e. $b$-shot). ``Avg.'' is the average result for all languages. We report the mean results across 20 runs with
random restarts. The subscript represents the corresponding standard deviation.}
    \label{tab1}
\vspace{-0.3cm}
\end{table*}
\addtolength{\tabcolsep}{1pt}

%\victor{The experimental setup should go to the appendix}

% \subsection{Experimental Setups}
% Our method is based on \texttt{XLM-RoberTa-base} model \cite{conneau-etal-2020-unsupervised}, which is a widely used multilingual pretrained language model. All the experiments are run 5 times with different random seeds (\{1, 10, 100, 1000, 10000\}).

\smallskip \noindent \textbf{Baselines}
\vic{We compared N2C2 with the following cross-lingual language models: 
(1) \textbf{In-Context Learning (ICL)} \cite{Brown2020LanguageMA}:{
this method utilizes $b$ input-label pairs from the training data and employs in-context learning.} 
(2) \textbf{Contextual Calibration (ICL+CC)} \cite{pmlr-v139-zhao21c}: this approach addresses prediction bias in ICL by introducing a content-free input ``N/A''; (3) \textbf{Fine Tuning (FT)}:  this method uses a classifier head that takes the [CLS] token as input and fine-tunes over MPLMs with the classification head; (4) \textbf{Prompt Tuning (PT)} \cite{schick-schutze-2021-exploiting}: a prompt-based fine-tuning method that utilizes manual prompts and fine-tunes over MPLMs.} \simon{(5) \textbf{Cross-lingual In-context
Source-Target Alignment (X-InSTA)} \cite{tanwar2023multilingual}: it prepends the top-${K}$ similar in-context samples which are retrieved from the training samples augmented with
task alignment. 

To ensure fairness, we used XLM-R as the base model, but we also present results for the variant  X-InSTA$^{*}$, using XGLM-7.5B \cite{lin-etal-2022-shot}.}

%\subsection{Baselines}
%We compared our model with the following cross-lingual language models: 
% (1) \textbf{PV-Zero}: we take the constructed prompts and verbalizers to obtain prediction of PLMs without any fine-tuning; 
%(1) \textbf{In-Context Learning (ICL)} \cite{Brown2020LanguageMA}: we apply the same training paradigm with the same $k$-shot learning examples from our main model, where the backbone model still use \texttt{XLM-RoBERTa-base}; (2) \textbf{Contextual Calibration (ICL+CC)} \cite{pmlr-v139-zhao21c}: remove the prediction bias in ICL by introducing content-free input "N/A"; (3) \textbf{Fine-tuning (FT)}: following the standard fine-tuning paradigm, by introducing a classifier taking the \texttt{[CLS]} token as input; (4) \textbf{Prompt-tuning (PET)} \cite{schick-schutze-2021-exploiting}: the prompt-based fine-tuning method that employs manual prompts and fine-tunes over PLMs. 
\renewcommand{\arraystretch}{1.02}
\begin{table*}[]
\centering
\tiny
\begin{tabular}{lcccc}
\toprule
\textbf{Language} &                                                              \textbf{N2C2} &                                                 \textbf{(1) W/o CD} &                                   \textbf{(2) W/o retrieval representation shaping} &                          \textbf{(3) W/o DWE} \\
\midrule
\multicolumn{5}{c}{\textbf{MARC}} \\
\midrule
De      &  $\text{37.75}_{\pm\text{2.3}}$ / $\text{10.94}_{\pm\text{5.9}}$ &    $\text{37.07}_{\pm\text{2.4}}$ / $\text{12.35}_{\pm\text{5.9}}$ &  $\text{36.16}_{\pm\text{1.2}}$ / $\text{4.82}_{\pm\text{1.0}}$ &  $\text{37.99}_{\pm\text{2.5}}$ / $\text{16.18}_{\pm\text{6.4}}$ \\
En      &   $\text{42.85}_{\pm\text{1.4}}$ / $\text{5.69}_{\pm\text{2.7}}$ &     $\text{42.10}_{\pm\text{1.6}}$ / $\text{6.90}_{\pm\text{1.8}}$ &  $\text{38.09}_{\pm\text{1.9}}$ / $\text{4.91}_{\pm\text{1.2}}$ &  $\text{43.33}_{\pm\text{1.4}}$ / $\text{11.06}_{\pm\text{3.6}}$ \\
Es      &  $\text{34.50}_{\pm\text{2.0}}$ / $\text{12.88}_{\pm\text{5.4}}$ &    $\text{33.76}_{\pm\text{1.9}}$ / $\text{14.45}_{\pm\text{5.0}}$ &  $\text{33.95}_{\pm\text{1.6}}$ / $\text{4.49}_{\pm\text{1.6}}$ &  $\text{34.90}_{\pm\text{2.2}}$ / $\text{17.45}_{\pm\text{6.1}}$ \\
Fr      &  $\text{33.51}_{\pm\text{2.0}}$ / $\text{17.01}_{\pm\text{6.4}}$ &    $\text{32.92}_{\pm\text{1.7}}$ / $\text{18.50}_{\pm\text{6.3}}$ &  $\text{33.17}_{\pm\text{1.9}}$ / $\text{5.72}_{\pm\text{1.9}}$ &  $\text{34.06}_{\pm\text{2.0}}$ / $\text{22.30}_{\pm\text{7.1}}$ \\
Ja      &   $\text{38.72}_{\pm\text{2.3}}$ / $\text{9.36}_{\pm\text{4.8}}$ &    $\text{37.98}_{\pm\text{1.8}}$ / $\text{10.37}_{\pm\text{3.6}}$ &  $\text{34.85}_{\pm\text{2.1}}$ / $\text{5.26}_{\pm\text{2.2}}$ &  $\text{39.37}_{\pm\text{2.1}}$ / $\text{14.92}_{\pm\text{6.0}}$ \\
Zh      &   $\text{35.10}_{\pm\text{2.1}}$ / $\text{9.49}_{\pm\text{4.5}}$ &    $\text{34.39}_{\pm\text{1.5}}$ / $\text{10.62}_{\pm\text{3.5}}$ &  $\text{33.45}_{\pm\text{1.4}}$ / $\text{4.82}_{\pm\text{1.4}}$ &  $\text{35.74}_{\pm\text{2.1}}$ / $\text{14.17}_{\pm\text{5.3}}$ \\
Avg. &  $\text{37.07}_{\pm\text{3.2}}$ / $\text{10.89}_{\pm\text{3.5}}$ &    $\text{36.37}_{\pm\text{3.4}}$ / $\text{12.20}_{\pm\text{4.0}}$ &  $\text{34.94}_{\pm\text{1.9}}$ / $\text{5.00}_{\pm\text{0.4}}$ &  $\text{37.56}_{\pm\text{3.5}}$ / $\text{16.01}_{\pm\text{3.8}}$ \\
\rowcolor{lightgray}\textit{p-value} & - & 8.6e-01 / 1.0e-02 & 3.4e-02 / 1.1e-02 & 5.8e-04 / 6.1e-07 \\
\midrule
\multicolumn{5}{c}{\textbf{CLS}} \\
\midrule
De      &  $\text{46.15}_{\pm\text{6.9}}$ / $\text{7.84}_{\pm\text{4.0}}$ &    $\text{45.02}_{\pm\text{7.2}}$ / $\text{8.40}_{\pm\text{4.3}}$ &  $\text{44.21}_{\pm\text{6.6}}$ / $\text{7.19}_{\pm\text{3.6}}$ &  $\text{45.85}_{\pm\text{7.0}}$ / $\text{11.79}_{\pm\text{5.4}}$ \\
En      &  $\text{47.64}_{\pm\text{2.5}}$ / $\text{6.56}_{\pm\text{2.6}}$ &    $\text{46.37}_{\pm\text{2.3}}$ / $\text{7.88}_{\pm\text{2.6}}$ &  $\text{43.41}_{\pm\text{4.1}}$ / $\text{5.19}_{\pm\text{2.5}}$ &  $\text{48.99}_{\pm\text{2.4}}$ / $\text{11.52}_{\pm\text{4.5}}$ \\
Fr      &  $\text{41.64}_{\pm\text{5.0}}$ / $\text{8.00}_{\pm\text{4.6}}$ &    $\text{40.71}_{\pm\text{5.4}}$ / $\text{9.09}_{\pm\text{4.6}}$ &  $\text{41.20}_{\pm\text{5.2}}$ / $\text{5.88}_{\pm\text{2.1}}$ &   $\text{41.01}_{\pm\text{4.8}}$ / $\text{9.79}_{\pm\text{5.5}}$ \\
Jp      &  $\text{40.52}_{\pm\text{3.2}}$ / $\text{8.05}_{\pm\text{2.9}}$ &    $\text{39.66}_{\pm\text{3.1}}$ / $\text{8.82}_{\pm\text{2.4}}$ &  $\text{33.65}_{\pm\text{2.3}}$ / $\text{6.25}_{\pm\text{2.5}}$ &  $\text{40.87}_{\pm\text{3.0}}$ / $\text{14.66}_{\pm\text{4.4}}$ \\
Avg. &  $\text{43.99}_{\pm\text{3.0}}$ / $\text{7.61}_{\pm\text{0.6}}$ &    $\text{42.94}_{\pm\text{3.3}}$ / $\text{8.55}_{\pm\text{0.5}}$ &  $\text{40.62}_{\pm\text{4.8}}$ / $\text{6.13}_{\pm\text{0.8}}$ &  $\text{44.18}_{\pm\text{3.0}}$ / $\text{11.94}_{\pm\text{2.0}}$ \\
\rowcolor{lightgray}
\textit{p-value} & - & 1.7e-01 / 3.6e-02 & 9.6e-02 / 1.9e-02 & 6.9e-01 / 2.3e-02 \\
\bottomrule
\end{tabular}
   \caption{\textbf{Ablation study results for N2C2.} Results are reported as $\textbf{Accuracy}\uparrow(\%)$ / $\textbf{ECE}\downarrow (\%)$. The p-value is calculated by two-tailed t-tests.}
    \label{ablation}
\vspace{-0.2cm}
\end{table*}

\subsection{Main Results}

Table \ref{tab1} %\nb{Please write Table~\ref{bla}. It is not good practice to put the numbers as they can change if tables are added or deleted}
offers an overview of the performance of N2C2 alongside the baselines. Detailed results including all other languages are available in App.\ \ref{app:experiments}, Tables~\ref{appendix:marc_results} and \ref{appendix:cls_result}. From Table \ref{tab1}, it is evident that N2C2 outperforms all baselines in terms of accuracy. N2C2 performs very well across all datasets, achieving an average 4.2\%, 3.24\% accuracy improvement  over the strong baseline (PT) on the MARC  and CLS datasets, respectively, with only 8, 16 and 32 shots.
In addition, it is noteworthy that that the effectiveness of the X-InSTA model does not increase with the number of shots, whereas our method shows a continuous improvement as the number of shots increases, even surpassing  X-InSTA$^{*}$ when $b = 32$.

% Nevertheless, as demonstrated at 16 and 32 shots, the performance of our methods improves and still surpasses all other baselines, including fine-tuning and prompt-tuning methods.

Particularly remarkable is N2C2's  performance in terms of ECE, with an average decrease of 10.53\% and 16.47\%  when compared to the strongest baseline on the MARC and CLS datasets, respectively. Another important observation is that while the baseline methods show varying degrees of improvement in accuracy with the increase of training data, their ECE does not consistently decrease and in some cases even exhibits a slight increase. In contrast, our proposed method demonstrates a consistent decrease in ECE. For a more detailed  analysis, please see Appendix~\ref{app:experiments}. 

Only when compared to X-InSTA$^{*}$, which is 60 times larger than $\text{XLM-RoBERTa}_{\text{base}}$, and when compared to fine-tuning and prompt-tuning methods with limited data ($b = 2$ or 4), our method does not perform better. However, it is important to note that the CLS dataset suffers from class imbalance, where the number of instances belonging to two classes is approximately half of those in the other two classes. 
% It is conceivable that  due to the class imbalance, both fine tuning and prompt tuning methods outperform ours at 2 or 4 shots. 

Overall, N2C2 exhibits superior performance over baselines in both accuracy and ECE across all datasets, demonstrating consistent improvement with as the training data increases, although facing challenges with class imbalance in some cases.

\begin{figure}[t]
    \centering
    \includegraphics[width=0.4\textwidth]{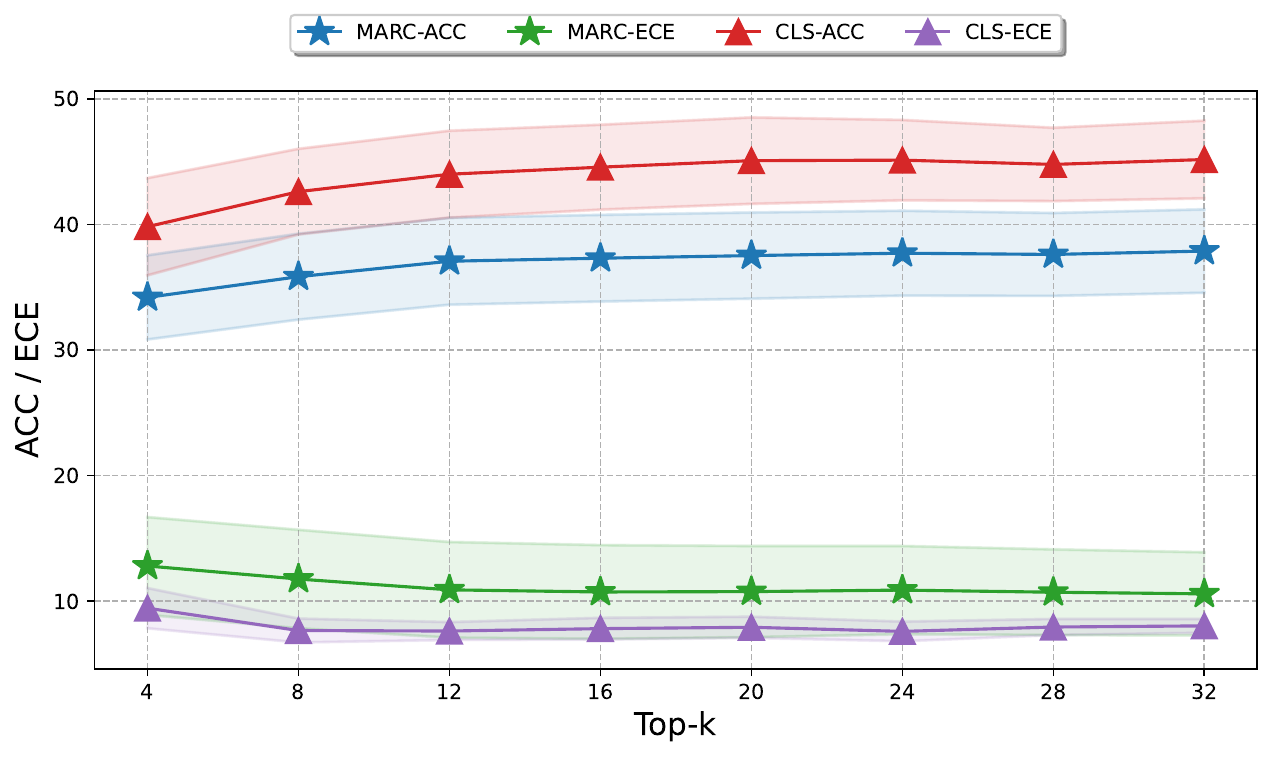}
    \caption{Top-$k$ Ablation Study}
    \label{fig:topK}
\vspace{-0.4cm}
\end{figure}
\subsection{Ablation Study}

Table \ref{ablation} outlines ablation experiments for N2C2 (XLMR-base).  In Variant (1), removing the confidence module (\ref{4.2}) results in a loss of 0.7\% and 1\% accuracy on  MARC and CLS, respectively, along with an increase of 1.3\% and 1.1\% in ECE. Variant (2) involves $k$NN retrieval in original feature spaces (\ref{4.1}), leading to a significant 3\% accuracy drop for both datasets, yet lower ECE.%, suggesting more accurate retrieval disrupts perfect calibration. 
In Variant (3), fixing retrieval results to the top 8 examples maintains accuracy but notably increases ECE by 5\% and 4\% on MARC and CLS, respectively. Overall, the confidence-aware distribution construction module  can improve accuracy and calibration, while retrieval representation shaping and the weight estimation network (DWE) can aid calibration.

%Table 5 presents the results of ablation experiments for N2C2 (XLMR-base). In Variant (1), we remove the confidence module, which means we do not modify the $k$NN distribution based on the confidence of the examples. The results show a loss of 0.7\% and 1\% in accuracy for the MARC and CLS, respectively. However, there is an improvement of 1.3\% and 1.1\% in ECE. This indicates that the usage of the confidence leads to better overall performance.

%In Variant (2), we omit the retrieval representation shaping. The results show a significant drop in accuracy, with a decrease of 3\% for both datasets. However, the ECE error is lower. We speculate that for our method, more accurate retrieval means retrieving more examples consistent with the test labels for both correct and incorrect predictions. This violates the definition of perfect calibration, where the probability should be 1 for correct predictions and 0 for incorrect predictions, resulting in higher ECE.

%In Variant (3), we omit the combination of multiple retrieval result sets and only use the top 8 retrieved examples. We find that the accuracy remains the same as our method, but there is an improvement of 5\% and 4\% in ECE for the MARC and CLS, respectively. This suggests that using multiple candidate retrieval sets plays a crucial role in boosting the model's confidence. \jieh{I'm not sure how to say about this.}

\smallskip \noindent \textbf{Influence of Top $K_{max}$}
\label{in_topk}
\vic{We also investigate the impact of $K_{max}$ %\nb{Check in the entire draft top k vs top M (or whatever letter was used before) \jieh{checked}} 
mentioned in Section \ref{adaptive combine} on the performance under 16 shots. We compute the average accuracy and ECE across all language results. Our analysis reveals that accuracy and ECE reach a saturation point, beyond which they do not significantly improve with an increase in $K_{max}$. This might be because retrieving more neighbors can introduce noise to the $k$NN distribution, and the influence of neighbors decreases as their distance from the query increases. Based on the results shown in Figure~\ref{fig:topK}, we use $K_{max}=16$ for the MARC dataset and $K_{max}=12$ for the CLS dataset.} 

%In this section, we analyze the impact of setting the size of $k_{max}$ in Section \ref{adaptive combine}. We calculate the average accuracy and ECE across all language results. We find that accuracy and ECE have nearly the same saturation point, beyond which they do not significantly improve with an increase in $k_{max}$. We speculate that this is because retrieving more neighbors can introduce noise to the $k$NN distribution, and the influence of neighbors decreases as their distance from the query increases. Based on the results shown in the graph, we select $k_{max}=16$ for the MARC dataset and $k_{max}=12$ for the CLS dataset.

\subsection{Comparisons With Other Calibration Methods}
\vic{
Given the effectiveness of N2C2 in reducing ECE, we are interested in examining the performance of classic calibration methods  for cross-lingual scenarios. Specifically, we consider the \emph{temperature scaling (TS)} technique~\cite{pmlr-v70-guo17a} and \emph{label smoothing (LS)}~\cite{Szegedy_2016_CVPR}. As both methods require training, we solely focus on fine tuning and prompt tuning, and calculate the average results across all languages under 16 shots. 

We can observe in Table~\ref{tab:ececomp} that both methods significantly reduce the ECE. An exception is the application of temperature scaling during prompt tuning, where its effectiveness is limited. This occurs probably because in prompt tuning (before applying \emph{softmax}) the probabilities of the retrieved labels are extremely small. Despite the effectiveness of these two classical calibration methods, N2C2 consistently outperforms them by a considerable margin. This indicates the robustness and superior performance of N2C2 in achieving enhanced calibration compared to other approaches. }

\begin{table}
\centering
\tiny
\begin{tabular}{clll|l}
\toprule
\multirow{1}{*}{\textbf{Dataset}} & \multirow{1}{*}{\textbf{Methods}} & \multicolumn{1}{c}{\textbf{FT}} & \multicolumn{1}{c}{\textbf{PT}} & \textbf{N2C2} \\
\midrule
% {} & {} & \multicolumn{1}{c}{ECE} & \multicolumn{1}{c}{Acc} & \multicolumn{1}{c}{ECE} \\
\multirow{4}{*}{\textbf{MARC}} & Vanilla & $\text{53.05}_{\pm \text{3.2}}$ & $\text{57.94}_{\pm \text{1.3}}$ & \multirow{4}{*}{$\textbf{10.89}_{\pm \textbf{3.5}}$}\\
{} & LS & $\text{45.36}_{\pm \text{5.01}}$&  $\text{42.04}_{\pm \text{0.87}}$\\
{} & TS & $\text{43.87}_{\pm \text{20.82}}$&  $\text{57.94}_{\pm \text{1.41}}$\\
{} & LS + TS & $\text{35.93}_{\pm \text{18.0}}$ &  $\text{42.04}_{\pm \text{0.87}}$\\
% {} & Ours & $\text{37.75}_{\pm \text{}}$ & $\text{37.75}_{\pm \text{}}$ & $\text{37.75}_{\pm \text{}}$ & $\text{37.75}_{\pm \text{}}$ \\
\midrule
\multirow{4}{*}{\textbf{CLS}} & Vanilla & $\text{37.48}_{\pm \text{17.6}}$ & $\text{54.06}_{\pm \text{2.3}}$ & \multirow{4}{*}{$\textbf{7.61}_{\pm \textbf{0.6}}$}\\
{} & LS  & $\text{29.91}_{\pm \text{18.17}}$ & $\text{40.24}_{\pm \text{2.36}}$\\
{} & TS  & $\text{30.62}_{\pm \text{6.60}}$ & $\text{54.06}_{\pm \text{2.69}}$ \\
{} & LS + TS & $\text{27.17}_{\pm \text{3.17}}$ & $\text{40.24}_{\pm \text{2.36}}$\\
\bottomrule

\end{tabular}
\caption{Calibration Errors for $\text{XLM-RoBERTa}_{\text{base}}$ when using different methods for calibration.}\label{tab:ececomp}
\vspace{-0.4cm}
\end{table}

\begin{figure}[!pt]
    \centering
    \includegraphics[width=0.4\textwidth]{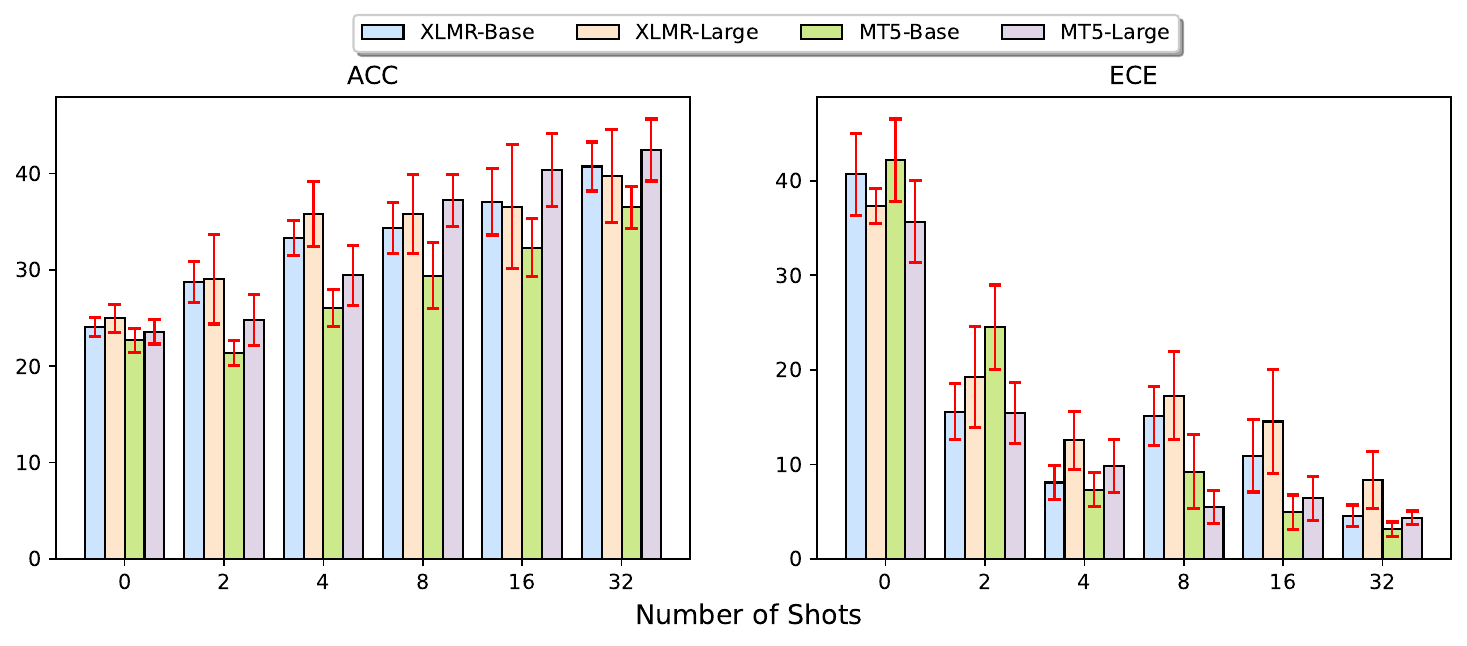}
    \caption{Comparsion with different model sizes and architectures on MARC. We provide the average scores over all languages.}
\label{marc_model}
\vspace{-0.35cm}
\end{figure}
\subsection{Effects of Model Size and Architecture }
\vic{Further questions are whether N2C2 is effective on larger models and how it performs in different multilingual frameworks. To answer these questions, we implement our methods using the XLMR-large and MT5 models \cite{xue-etal-2021-mt5}. 

\vc{Figure \ref{marc_model} (and Figure~\ref{cls_model} in App.\ \ref{app:experiments}), shows that on average, larger models exhibit superior performance than their base counterparts, with lower ECE and higher accuracy. This  proves that our method also works on larger models.} In terms of accuracy, MT5-base is significantly inferior than XLMR-base, while MT5-large performs similarly to XLMR-large in terms of ECE. We believe that this is because  MT5  itself is not as effective as the XLMR series, but the increase in the parameter size of MT5-large (1.2B).} %compensates for its inherent shortcomings compared to XLMR-large with 560M parameters.}

\subsection{Qualitative Analysis}
\begin{figure}
    \centering
    \includegraphics[width=0.46\textwidth]{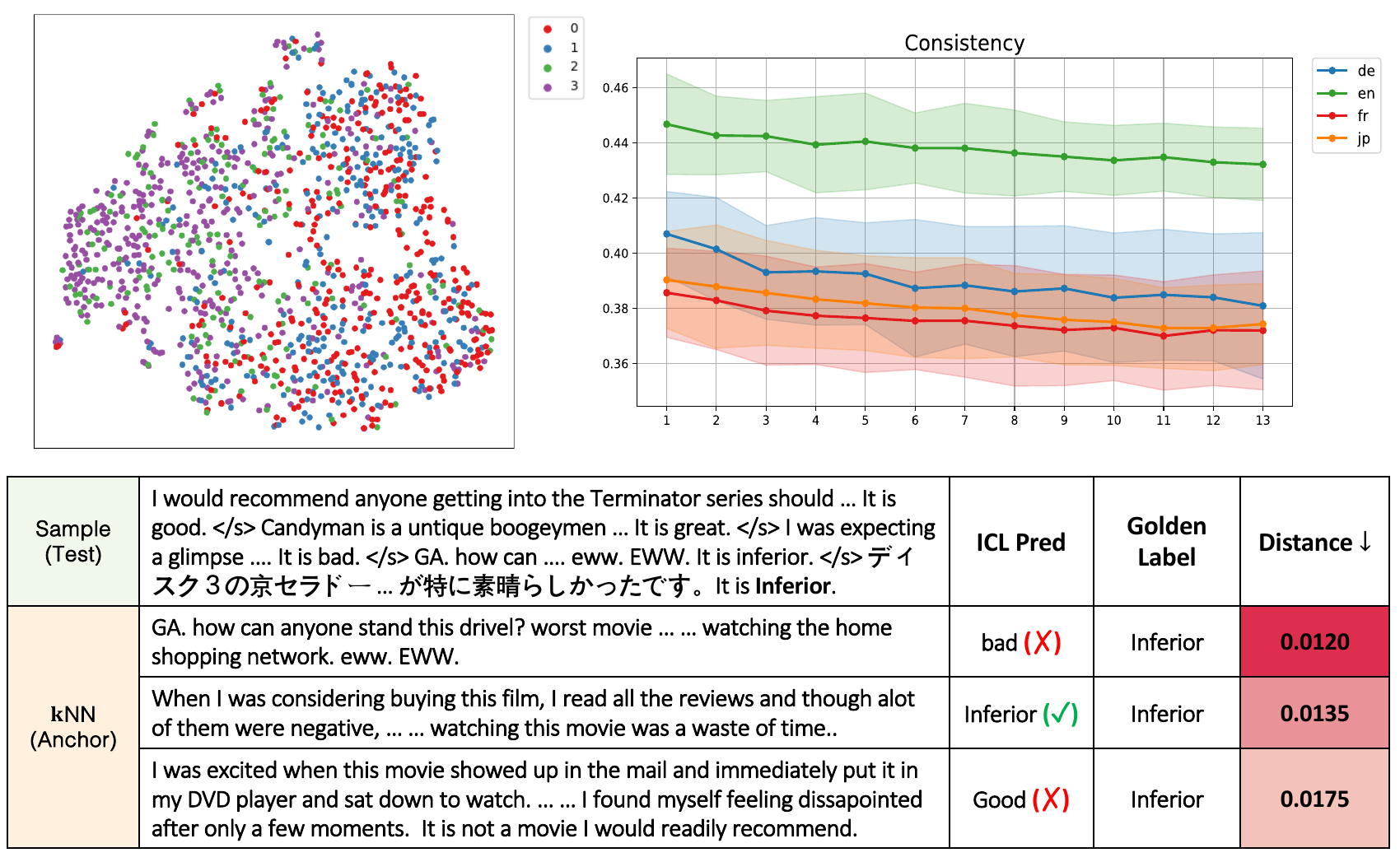}
    \caption{The scatter plot is the t-SNE for test cases in CLS dataset. In the top right figure, consistency scores are computed based on the alignment between the gold label of the top $k$ retrieved examples and the gold label of the anchor example. The bottom figure displays one randomly selected example that our method predict correctly and the three examples retrieved from the source language.}
    \label{case}
\vspace{-0.4cm}
\end{figure}

\vic{We conducted  qualitative analysis, as shown in  Figure \ref{case}}. First, from the scatter plot depicting the distribution of test cases, it is evident that the LM distribution is relatively poor for direct prediction during cross-lingual in context learning.  Secondly, despite a decrease in consistency with distance, the consistency remains relatively high, as depicted in the top right subfigure, which illustrates whether the top 12 examples retrieved for each language in CLS are consistent with the gold labels of the test examples. This observation suggests a potential reason why N2C2 effectively reduces calibration errors. Additionally, our method correctly predicts the randomly selected example and the three examples retrieved from the source language, all sharing the same label as the test example. In contrast, the ICL method (Baseline) fails to predict accurately in similar scenarios. This further shows the superiority of our method.
\vspace{-0.7em}
\section{Conclusion}
\vic{In our study, we have investigated the performance of multilingual models in cross-lingual ICL scenarios, revealing a notable deficiency in both performance and calibration. To address this, we propose N2C2, a cross-lingual ICL technique that leverages  examples from the source  language. We conduct experiments on two multilingual sentiment classification datasets, comparing our method with strong baselines and popular calibration methods. The results show N2C2 significantly improves the performance in terms of accuracy and expected calibration errors. Furthermore, our ablation studies demonstrate the contributions of each module within our framework, providing deeper insights into their role and impact. Importantly, our approach exhibits scalability, proving effective even with larger models. Our proposed method not only provides a substantial improvement in cross-lingual ICL, but also offers insights for future research in more effective cross-lingual learning strategies.}

%We introduced N2C2, a zero-shot cross-lingual in-context learning method that retrieves support examples from the source training language. Our method has the following features: (1) It retrieves more relevant text from the source training language using a representation reconstruction module, effectively improving the accuracy of retrieval. (2) It dynamically adjusts the weights of the retrieval neighbors based on confidence and combines multiple retrieval candidate sets. Evaluation on two multilingual classification datasets demonstrates that N2C2 significantly outperforms the previous baseline and performs comparably with different K-shot settings. Overall, N2C2 demonstrates the possibility of achieving significantly higher zero-shot cross-lingual performance using multilingual pre-trained language models. This opens up a new research direction towards effectively reducing the calibration error in cross-lingual in-context learning.

\section*{Limitations}

% \textbf{Beyond multilingual classification.}\quad
Our experiments focused on multilingual classification tasks. However, extending our approach to tasks such as multilingual multi-choice or multilingual generation tasks would require exploring beyond a fixed set of verbalizations. %It would be particularly interesting to study the calibration performance of cross-lingual in-context generation tasks. We leave the investigation of extensions to non-classification tasks for future work.

% \inlineSubsection{Extension to multilingual generative models.} 
Due to computational resource limitations, we did not conduct experiments on decoder-only multilingual models, such as XGLM \cite{lin-etal-2022-shot}, \simon{BLOOM \cite{Scao2022BLOOMA1}} or even large language models like ChatGPT. It would be interesting to examine the performance of these models in cross-lingual ICL.

%Due to computational resource limitations, we did not conduct experiments on decoder-only multilingual models, such as XGLM or even large language models like ChatGPT. It would be interesting to examine the performance of these models in cross-lingual in-context learning. We leave this as an avenue for future research.

% \inlineSubsection{More extensive demonstrations.} 
Additionally, due to length constraints, we only used 1-shot per class demonstrations for each test input and averaged the predictions using ensemble ICL. However, ICL often benefits from longer and richer demonstrations. It would be valuable to explore how N2C2 could incorporate existing retrieval demonstration approaches and investigate the calibration error performance when longer demonstrations are used.
% \section*{Ethics Statement}
% Scientific work published at EMNLP 2023 must comply with the \href{https://www.aclweb.org/portal/content/acl-code-ethics}{ACL Ethics Policy}. We encourage all authors to include an explicit ethics statement on the broader impact of the work, or other ethical considerations after the conclusion but before the references. The ethics statement will not count toward the page limit (8 pages for long, 4 pages for short papers).

\bibliography{emnlp2023,custom}

\begin{thebibliography}{49}
\expandafter\ifx\csname natexlab\endcsname\relax\def\natexlab#1{#1}\fi

\bibitem[{Ahuja et~al.(2022)Ahuja, Sitaram, Dandapat, and Choudhury}]{ahuja-etal-2022-calibration}
Kabir Ahuja, Sunayana Sitaram, Sandipan Dandapat, and Monojit Choudhury. 2022.
\newblock \href {https://aclanthology.org/2022.emnlp-main.290} {On the calibration of massively multilingual language models}.
\newblock In \emph{Proceedings of the 2022 Conference on Empirical Methods in Natural Language Processing}, pages 4310--4323, Abu Dhabi, United Arab Emirates. Association for Computational Linguistics.

\bibitem[{Bose et~al.(2022)Bose, Aletras, Illina, and Fohr}]{bose-etal-2022-dynamically}
Tulika Bose, Nikolaos Aletras, Irina Illina, and Dominique Fohr. 2022.
\newblock \href {https://doi.org/10.18653/v1/2022.findings-acl.32} {Dynamically refined regularization for improving cross-corpora hate speech detection}.
\newblock In \emph{Findings of the Association for Computational Linguistics: ACL 2022}, pages 372--382, Dublin, Ireland. Association for Computational Linguistics.

\bibitem[{Brown et~al.(2020{\natexlab{a}})Brown, Mann, Ryder, Subbiah, Kaplan, Dhariwal, Neelakantan, Shyam, Sastry, Askell, Agarwal, Herbert-Voss, Krueger, Henighan, Child, Ramesh, Ziegler, Wu, Winter, Hesse, Chen, Sigler, Litwin, Gray, Chess, Clark, Berner, McCandlish, Radford, Sutskever, and Amodei}]{NEURIPS2020_1457c0d6}
Tom Brown, Benjamin Mann, Nick Ryder, Melanie Subbiah, Jared~D Kaplan, Prafulla Dhariwal, Arvind Neelakantan, Pranav Shyam, Girish Sastry, Amanda Askell, Sandhini Agarwal, Ariel Herbert-Voss, Gretchen Krueger, Tom Henighan, Rewon Child, Aditya Ramesh, Daniel Ziegler, Jeffrey Wu, Clemens Winter, Chris Hesse, Mark Chen, Eric Sigler, Mateusz Litwin, Scott Gray, Benjamin Chess, Jack Clark, Christopher Berner, Sam McCandlish, Alec Radford, Ilya Sutskever, and Dario Amodei. 2020{\natexlab{a}}.
\newblock \href {https://proceedings.neurips.cc/paper_files/paper/2020/file/1457c0d6bfcb4967418bfb8ac142f64a-Paper.pdf} {Language models are few-shot learners}.
\newblock In \emph{Advances in Neural Information Processing Systems}, volume~33, pages 1877--1901. Curran Associates, Inc.

\bibitem[{Brown et~al.(2020{\natexlab{b}})Brown, Mann, Ryder, Subbiah, Kaplan, Dhariwal, Neelakantan, Shyam, Sastry, Askell, Agarwal, Herbert-Voss, Krueger, Henighan, Child, Ramesh, Ziegler, Wu, Winter, Hesse, Chen, Sigler, Litwin, Gray, Chess, Clark, Berner, McCandlish, Radford, Sutskever, and Amodei}]{Brown2020LanguageMA}
Tom~B. Brown, Benjamin Mann, Nick Ryder, Melanie Subbiah, Jared Kaplan, Prafulla Dhariwal, Arvind Neelakantan, Pranav Shyam, Girish Sastry, Amanda Askell, Sandhini Agarwal, Ariel Herbert-Voss, Gretchen Krueger, T.~J. Henighan, Rewon Child, Aditya Ramesh, Daniel~M. Ziegler, Jeff Wu, Clemens Winter, Christopher Hesse, Mark Chen, Eric Sigler, Mateusz Litwin, Scott Gray, Benjamin Chess, Jack Clark, Christopher Berner, Sam McCandlish, Alec Radford, Ilya Sutskever, and Dario Amodei. 2020{\natexlab{b}}.
\newblock Language models are few-shot learners.
\newblock \emph{ArXiv}, abs/2005.14165.

\bibitem[{Chen et~al.(2023)Chen, Yuan, Cui, Liu, and Ji}]{chen2023close}
Yangyi Chen, Lifan Yuan, Ganqu Cui, Zhiyuan Liu, and Heng Ji. 2023.
\newblock \href {http://arxiv.org/abs/2211.00151} {A close look into the calibration of pre-trained language models}.

\bibitem[{Conneau et~al.(2020)Conneau, Khandelwal, Goyal, Chaudhary, Wenzek, Guzm{\'a}n, Grave, Ott, Zettlemoyer, and Stoyanov}]{conneau-etal-2020-unsupervised}
Alexis Conneau, Kartikay Khandelwal, Naman Goyal, Vishrav Chaudhary, Guillaume Wenzek, Francisco Guzm{\'a}n, Edouard Grave, Myle Ott, Luke Zettlemoyer, and Veselin Stoyanov. 2020.
\newblock \href {https://doi.org/10.18653/v1/2020.acl-main.747} {Unsupervised cross-lingual representation learning at scale}.
\newblock In \emph{Proceedings of the 58th Annual Meeting of the Association for Computational Linguistics}, pages 8440--8451, Online. Association for Computational Linguistics.

\bibitem[{Desai and Durrett(2020)}]{desai-durrett-2020-calibration}
Shrey Desai and Greg Durrett. 2020.
\newblock \href {https://doi.org/10.18653/v1/2020.emnlp-main.21} {Calibration of pre-trained transformers}.
\newblock In \emph{Proceedings of the 2020 Conference on Empirical Methods in Natural Language Processing (EMNLP)}, pages 295--302, Online. Association for Computational Linguistics.

\bibitem[{Gao et~al.(2021)Gao, Fisch, and Chen}]{gao-etal-2021-making}
Tianyu Gao, Adam Fisch, and Danqi Chen. 2021.
\newblock \href {https://doi.org/10.18653/v1/2021.acl-long.295} {Making pre-trained language models better few-shot learners}.
\newblock In \emph{Proceedings of the 59th Annual Meeting of the Association for Computational Linguistics and the 11th International Joint Conference on Natural Language Processing (Volume 1: Long Papers)}, pages 3816--3830, Online. Association for Computational Linguistics.

\bibitem[{Guo et~al.(2017)Guo, Pleiss, Sun, and Weinberger}]{pmlr-v70-guo17a}
Chuan Guo, Geoff Pleiss, Yu~Sun, and Kilian~Q. Weinberger. 2017.
\newblock \href {https://proceedings.mlr.press/v70/guo17a.html} {On calibration of modern neural networks}.
\newblock In \emph{Proceedings of the 34th International Conference on Machine Learning}, volume~70 of \emph{Proceedings of Machine Learning Research}, pages 1321--1330. PMLR.

\bibitem[{He et~al.(2021)He, McCann, Xiong, and Hosseini-Asl}]{he-etal-2021-joint}
Tianxing He, Bryan McCann, Caiming Xiong, and Ehsan Hosseini-Asl. 2021.
\newblock \href {https://doi.org/10.18653/v1/2021.eacl-main.151} {Joint energy-based model training for better calibrated natural language understanding models}.
\newblock In \emph{Proceedings of the 16th Conference of the European Chapter of the Association for Computational Linguistics: Main Volume}, pages 1754--1761, Online. Association for Computational Linguistics.

\bibitem[{Hendrycks et~al.(2019)Hendrycks, Lee, and Mazeika}]{pmlr-v97-hendrycks19a}
Dan Hendrycks, Kimin Lee, and Mantas Mazeika. 2019.
\newblock \href {https://proceedings.mlr.press/v97/hendrycks19a.html} {Using pre-training can improve model robustness and uncertainty}.
\newblock In \emph{Proceedings of the 36th International Conference on Machine Learning}, volume~97 of \emph{Proceedings of Machine Learning Research}, pages 2712--2721. PMLR.

\bibitem[{Hu et~al.(2022)Hu, Lee, Xie, Yu, Smith, and Ostendorf}]{hu-etal-2022-context}
Yushi Hu, Chia-Hsuan Lee, Tianbao Xie, Tao Yu, Noah~A. Smith, and Mari Ostendorf. 2022.
\newblock \href {https://aclanthology.org/2022.findings-emnlp.193} {In-context learning for few-shot dialogue state tracking}.
\newblock In \emph{Findings of the Association for Computational Linguistics: EMNLP 2022}, pages 2627--2643, Abu Dhabi, United Arab Emirates. Association for Computational Linguistics.

\bibitem[{Huang et~al.(2022)Huang, Ma, Zhang, Wei, and Wang}]{huang-etal-2022-zero}
Lianzhe Huang, Shuming Ma, Dongdong Zhang, Furu Wei, and Houfeng Wang. 2022.
\newblock \href {https://aclanthology.org/2022.emnlp-main.790} {Zero-shot cross-lingual transfer of prompt-based tuning with a unified multilingual prompt}.
\newblock In \emph{Proceedings of the 2022 Conference on Empirical Methods in Natural Language Processing}, pages 11488--11497, Abu Dhabi, United Arab Emirates. Association for Computational Linguistics.

\bibitem[{Jiang et~al.(2020)Jiang, Xu, Araki, and Neubig}]{jiang-etal-2020-know}
Zhengbao Jiang, Frank~F. Xu, Jun Araki, and Graham Neubig. 2020.
\newblock \href {https://doi.org/10.1162/tacl_a_00324} {How can we know what language models know?}
\newblock \emph{Transactions of the Association for Computational Linguistics}, 8:423--438.

\bibitem[{Jiang et~al.(2022)Jiang, Liu, and Van~Durme}]{jiang-etal-2022-calibrating}
Zhengping Jiang, Anqi Liu, and Benjamin Van~Durme. 2022.
\newblock \href {https://aclanthology.org/2022.emnlp-main.170} {Calibrating zero-shot cross-lingual (un-)structured predictions}.
\newblock In \emph{Proceedings of the 2022 Conference on Empirical Methods in Natural Language Processing}, pages 2648--2674, Abu Dhabi, United Arab Emirates. Association for Computational Linguistics.

\bibitem[{Jung et~al.(2020)Jung, Kang, Cheng, Mentch, and Schaaf}]{jung-etal-2020-posterior}
Taehee Jung, Dongyeop Kang, Hua Cheng, Lucas Mentch, and Thomas Schaaf. 2020.
\newblock \href {https://doi.org/10.18653/v1/2020.acl-main.242} {Posterior calibrated training on sentence classification tasks}.
\newblock In \emph{Proceedings of the 58th Annual Meeting of the Association for Computational Linguistics}, pages 2723--2730, Online. Association for Computational Linguistics.

\bibitem[{Keung et~al.(2020)Keung, Lu, Szarvas, and Smith}]{keung-etal-2020-multilingual}
Phillip Keung, Yichao Lu, Gy{\"o}rgy Szarvas, and Noah~A. Smith. 2020.
\newblock \href {https://doi.org/10.18653/v1/2020.emnlp-main.369} {The multilingual {A}mazon reviews corpus}.
\newblock In \emph{Proceedings of the 2020 Conference on Empirical Methods in Natural Language Processing (EMNLP)}, pages 4563--4568, Online. Association for Computational Linguistics.

\bibitem[{Kim et~al.(2023)Kim, Ki, Kim, and Lee}]{kim2023boosting}
Sunkyoung Kim, Dayeon Ki, Yireun Kim, and Jinsik Lee. 2023.
\newblock \href {http://arxiv.org/abs/2305.15233} {Boosting cross-lingual transferability in multilingual models via in-context learning}.

\bibitem[{Kingma and Ba(2014)}]{kingma2014adam}
Diederik~P Kingma and Jimmy Ba. 2014.
\newblock Adam: A method for stochastic optimization.
\newblock \emph{arXiv preprint arXiv:1412.6980}.

\bibitem[{Li et~al.(2023)Li, Yan, Li, Qian, He, and Gui}]{li-etal-2023-distinguishability}
Hongjing Li, Hanqi Yan, Yanran Li, Li~Qian, Yulan He, and Lin Gui. 2023.
\newblock \href {https://aclanthology.org/2023.findings-eacl.102} {Distinguishability calibration to in-context learning}.
\newblock In \emph{Findings of the Association for Computational Linguistics: EACL 2023}, pages 1385--1397, Dubrovnik, Croatia. Association for Computational Linguistics.

\bibitem[{Lin et~al.(2022)Lin, Mihaylov, Artetxe, Wang, Chen, Simig, Ott, Goyal, Bhosale, Du, Pasunuru, Shleifer, Koura, Chaudhary, O{'}Horo, Wang, Zettlemoyer, Kozareva, Diab, Stoyanov, and Li}]{lin-etal-2022-shot}
Xi~Victoria Lin, Todor Mihaylov, Mikel Artetxe, Tianlu Wang, Shuohui Chen, Daniel Simig, Myle Ott, Naman Goyal, Shruti Bhosale, Jingfei Du, Ramakanth Pasunuru, Sam Shleifer, Punit~Singh Koura, Vishrav Chaudhary, Brian O{'}Horo, Jeff Wang, Luke Zettlemoyer, Zornitsa Kozareva, Mona Diab, Veselin Stoyanov, and Xian Li. 2022.
\newblock \href {https://aclanthology.org/2022.emnlp-main.616} {Few-shot learning with multilingual generative language models}.
\newblock In \emph{Proceedings of the 2022 Conference on Empirical Methods in Natural Language Processing}, pages 9019--9052, Abu Dhabi, United Arab Emirates. Association for Computational Linguistics.

\bibitem[{Liu et~al.(2022{\natexlab{a}})Liu, Shen, Zhang, Dolan, Carin, and Chen}]{liu-etal-2022-makes}
Jiachang Liu, Dinghan Shen, Yizhe Zhang, Bill Dolan, Lawrence Carin, and Weizhu Chen. 2022{\natexlab{a}}.
\newblock \href {https://doi.org/10.18653/v1/2022.deelio-1.10} {What makes good in-context examples for {GPT}-3?}
\newblock In \emph{Proceedings of Deep Learning Inside Out (DeeLIO 2022): The 3rd Workshop on Knowledge Extraction and Integration for Deep Learning Architectures}, pages 100--114, Dublin, Ireland and Online. Association for Computational Linguistics.

\bibitem[{Liu et~al.(2022{\natexlab{b}})Liu, Schick, and Schütze}]{liu2022semanticoriented}
Yanchen Liu, Timo Schick, and Hinrich Schütze. 2022{\natexlab{b}}.
\newblock \href {http://arxiv.org/abs/2202.06133} {Semantic-oriented unlabeled priming for large-scale language models}.

\bibitem[{Long et~al.(2020{\natexlab{a}})Long, Cai, Reid, Webber, and Xiong}]{long-etal-2020-shallow}
Wanqiu Long, Xinyi Cai, James Reid, Bonnie Webber, and Deyi Xiong. 2020{\natexlab{a}}.
\newblock \href {https://aclanthology.org/2020.lrec-1.129} {Shallow discourse annotation for {C}hinese {TED} talks}.
\newblock In \emph{Proceedings of the Twelfth Language Resources and Evaluation Conference}, pages 1025--1032, Marseille, France. European Language Resources Association.

\bibitem[{Long et~al.(2024)Long, N, and Webber}]{long-etal-2024-multi}
Wanqiu Long, Siddharth N, and Bonnie Webber. 2024.
\newblock \href {https://doi.org/10.18653/v1/2024.findings-acl.500} {Multi-label classification for implicit discourse relation recognition}.
\newblock In \emph{Findings of the Association for Computational Linguistics: ACL 2024}, pages 8437--8451, Bangkok, Thailand. Association for Computational Linguistics.

\bibitem[{Long and Webber(2022)}]{long-webber-2022-facilitating}
Wanqiu Long and Bonnie Webber. 2022.
\newblock \href {https://aclanthology.org/2022.emnlp-main.734} {Facilitating contrastive learning of discourse relational senses by exploiting the hierarchy of sense relations}.
\newblock In \emph{Proceedings of the 2022 Conference on Empirical Methods in Natural Language Processing}, pages 10704--10716, Abu Dhabi, United Arab Emirates. Association for Computational Linguistics.

\bibitem[{Long and Webber(2024)}]{long2024leveraginghierarchicalprototypesverbalizer}
Wanqiu Long and Bonnie Webber. 2024.
\newblock \href {http://arxiv.org/abs/2411.14880} {Leveraging hierarchical prototypes as the verbalizer for implicit discourse relation recognition}.

\bibitem[{Long et~al.(2020{\natexlab{b}})Long, Webber, and Xiong}]{long-etal-2020-ted}
Wanqiu Long, Bonnie Webber, and Deyi Xiong. 2020{\natexlab{b}}.
\newblock \href {https://doi.org/10.18653/v1/2020.emnlp-main.223} {{TED}-{CDB}: A large-scale {C}hinese discourse relation dataset on {TED} talks}.
\newblock In \emph{Proceedings of the 2020 Conference on Empirical Methods in Natural Language Processing (EMNLP)}, pages 2793--2803, Online. Association for Computational Linguistics.

\bibitem[{Lyu et~al.(2023)Lyu, Min, Beltagy, Zettlemoyer, and Hajishirzi}]{lyu2023zicl}
Xinxi Lyu, Sewon Min, Iz~Beltagy, Luke Zettlemoyer, and Hannaneh Hajishirzi. 2023.
\newblock \href {http://arxiv.org/abs/2212.09865} {Z-icl: Zero-shot in-context learning with pseudo-demonstrations}.

\bibitem[{Naeini et~al.(2015)Naeini, Cooper, and Hauskrecht}]{10.5555/2888116.2888120}
Mahdi~Pakdaman Naeini, Gregory~F. Cooper, and Milos Hauskrecht. 2015.
\newblock Obtaining well calibrated probabilities using bayesian binning.
\newblock In \emph{Proceedings of the Twenty-Ninth AAAI Conference on Artificial Intelligence}, AAAI'15, page 2901–2907. AAAI Press.

\bibitem[{Niculescu-Mizil and Caruana(2005)}]{10.1145/1102351.1102430}
Alexandru Niculescu-Mizil and Rich Caruana. 2005.
\newblock \href {https://doi.org/10.1145/1102351.1102430} {Predicting good probabilities with supervised learning}.
\newblock In \emph{Proceedings of the 22nd International Conference on Machine Learning}, ICML '05, page 625–632, New York, NY, USA. Association for Computing Machinery.

\bibitem[{Nie et~al.(2023)Nie, Liang, Schmid, and Schütze}]{nie2023crosslingual}
Ercong Nie, Sheng Liang, Helmut Schmid, and Hinrich Schütze. 2023.
\newblock \href {http://arxiv.org/abs/2212.09651} {Cross-lingual retrieval augmented prompt for low-resource languages}.

\bibitem[{Nie et~al.(2022)Nie, Chen, Zhang, and Cheng}]{nie2022improving}
Feng Nie, Meixi Chen, Zhirui Zhang, and Xu~Cheng. 2022.
\newblock \href {http://arxiv.org/abs/2212.02216} {Improving few-shot performance of language models via nearest neighbor calibration}.

\bibitem[{Park and Caragea(2022)}]{park-caragea-2022-calibration}
Seo~Yeon Park and Cornelia Caragea. 2022.
\newblock \href {https://doi.org/10.18653/v1/2022.acl-long.368} {On the calibration of pre-trained language models using mixup guided by area under the margin and saliency}.
\newblock In \emph{Proceedings of the 60th Annual Meeting of the Association for Computational Linguistics (Volume 1: Long Papers)}, pages 5364--5374, Dublin, Ireland. Association for Computational Linguistics.

\bibitem[{Prettenhofer and Stein(2010)}]{prettenhofer-stein-2010-cross}
Peter Prettenhofer and Benno Stein. 2010.
\newblock \href {https://aclanthology.org/P10-1114} {Cross-language text classification using structural correspondence learning}.
\newblock In \emph{Proceedings of the 48th Annual Meeting of the Association for Computational Linguistics}, pages 1118--1127, Uppsala, Sweden. Association for Computational Linguistics.

\bibitem[{Schick and Sch{\"u}tze(2021)}]{schick-schutze-2021-exploiting}
Timo Schick and Hinrich Sch{\"u}tze. 2021.
\newblock \href {https://doi.org/10.18653/v1/2021.eacl-main.20} {Exploiting cloze-questions for few-shot text classification and natural language inference}.
\newblock In \emph{Proceedings of the 16th Conference of the European Chapter of the Association for Computational Linguistics: Main Volume}, pages 255--269, Online. Association for Computational Linguistics.

\bibitem[{Shi et~al.(2022{\natexlab{a}})Shi, Zhang, Bai, and Lin}]{shi-etal-2022-xricl}
Peng Shi, Rui Zhang, He~Bai, and Jimmy Lin. 2022{\natexlab{a}}.
\newblock \href {https://aclanthology.org/2022.findings-emnlp.384} {{XRICL}: Cross-lingual retrieval-augmented in-context learning for cross-lingual text-to-{SQL} semantic parsing}.
\newblock In \emph{Findings of the Association for Computational Linguistics: EMNLP 2022}, pages 5248--5259, Abu Dhabi, United Arab Emirates. Association for Computational Linguistics.

\bibitem[{Shi et~al.(2022{\natexlab{b}})Shi, Michael, Gururangan, and Zettlemoyer}]{shi-etal-2022-nearest}
Weijia Shi, Julian Michael, Suchin Gururangan, and Luke Zettlemoyer. 2022{\natexlab{b}}.
\newblock \href {https://aclanthology.org/2022.emnlp-main.214} {Nearest neighbor zero-shot inference}.
\newblock In \emph{Proceedings of the 2022 Conference on Empirical Methods in Natural Language Processing}, pages 3254--3265, Abu Dhabi, United Arab Emirates. Association for Computational Linguistics.

\bibitem[{Szegedy et~al.(2016)Szegedy, Vanhoucke, Ioffe, Shlens, and Wojna}]{Szegedy_2016_CVPR}
Christian Szegedy, Vincent Vanhoucke, Sergey Ioffe, Jon Shlens, and Zbigniew Wojna. 2016.
\newblock Rethinking the inception architecture for computer vision.
\newblock In \emph{Proceedings of the IEEE Conference on Computer Vision and Pattern Recognition (CVPR)}.

\bibitem[{Tanwar et~al.(2023)Tanwar, Dutta, Borthakur, and Chakraborty}]{tanwar2023multilingual}
Eshaan Tanwar, Subhabrata Dutta, Manish Borthakur, and Tanmoy Chakraborty. 2023.
\newblock \href {http://arxiv.org/abs/2305.05940} {Multilingual llms are better cross-lingual in-context learners with alignment}.

\bibitem[{Wang et~al.(2022)Wang, Wang, Luo, Tan, Qiu, Yang, Shi, Huang, and Gao}]{wang-etal-2022-towards-unified}
Jianing Wang, Chengyu Wang, Fuli Luo, Chuanqi Tan, Minghui Qiu, Fei Yang, Qiuhui Shi, Songfang Huang, and Ming Gao. 2022.
\newblock \href {https://aclanthology.org/2022.findings-emnlp.37} {Towards unified prompt tuning for few-shot text classification}.
\newblock In \emph{Findings of the Association for Computational Linguistics: EMNLP 2022}, pages 524--536, Abu Dhabi, United Arab Emirates. Association for Computational Linguistics.

\bibitem[{Webson and Pavlick(2022)}]{webson-pavlick-2022-prompt}
Albert Webson and Ellie Pavlick. 2022.
\newblock \href {https://doi.org/10.18653/v1/2022.naacl-main.167} {Do prompt-based models really understand the meaning of their prompts?}
\newblock In \emph{Proceedings of the 2022 Conference of the North American Chapter of the Association for Computational Linguistics: Human Language Technologies}, pages 2300--2344, Seattle, United States. Association for Computational Linguistics.

\bibitem[{Winata et~al.(2021)Winata, Madotto, Lin, Liu, Yosinski, and Fung}]{winata-etal-2021-language}
Genta~Indra Winata, Andrea Madotto, Zhaojiang Lin, Rosanne Liu, Jason Yosinski, and Pascale Fung. 2021.
\newblock \href {https://doi.org/10.18653/v1/2021.mrl-1.1} {Language models are few-shot multilingual learners}.
\newblock In \emph{Proceedings of the 1st Workshop on Multilingual Representation Learning}, pages 1--15, Punta Cana, Dominican Republic. Association for Computational Linguistics.

\bibitem[{Workshop et~al.(2022)Workshop, Scao, Fan, Akiki, Pavlick, Ili{\'c}, Hesslow, Castagn{\'e}, Luccioni, Yvon et~al.}]{Scao2022BLOOMA1}
BigScience Workshop, Teven~Le Scao, Angela Fan, Christopher Akiki, Ellie Pavlick, Suzana Ili{\'c}, Daniel Hesslow, Roman Castagn{\'e}, Alexandra~Sasha Luccioni, Fran{\c{c}}ois Yvon, et~al. 2022.
\newblock \href {https://api.semanticscholar.org/CorpusID:253420279} {Bloom: A 176b-parameter open-access multilingual language model}.
\newblock \emph{ArXiv}, abs/2211.05100.

\bibitem[{Xu et~al.(2023)Xu, Wang, Mao, Lyu, She, and Zhang}]{xu2023kNN}
Benfeng Xu, Quan Wang, Zhendong Mao, Yajuan Lyu, Qiaoqiao She, and Yongdong Zhang. 2023.
\newblock \href {https://openreview.net/forum?id=fe2S7736sNS} {\$k\${NN} prompting: Beyond-context learning with calibration-free nearest neighbor inference}.
\newblock In \emph{The Eleventh International Conference on Learning Representations}.

\bibitem[{Xue et~al.(2021)Xue, Constant, Roberts, Kale, Al-Rfou, Siddhant, Barua, and Raffel}]{xue-etal-2021-mt5}
Linting Xue, Noah Constant, Adam Roberts, Mihir Kale, Rami Al-Rfou, Aditya Siddhant, Aditya Barua, and Colin Raffel. 2021.
\newblock \href {https://doi.org/10.18653/v1/2021.naacl-main.41} {m{T}5: A massively multilingual pre-trained text-to-text transformer}.
\newblock In \emph{Proceedings of the 2021 Conference of the North American Chapter of the Association for Computational Linguistics: Human Language Technologies}, pages 483--498, Online. Association for Computational Linguistics.

\bibitem[{Zhao and Sch{\"u}tze(2021)}]{zhao-schutze-2021-discrete}
Mengjie Zhao and Hinrich Sch{\"u}tze. 2021.
\newblock \href {https://doi.org/10.18653/v1/2021.emnlp-main.672} {Discrete and soft prompting for multilingual models}.
\newblock In \emph{Proceedings of the 2021 Conference on Empirical Methods in Natural Language Processing}, pages 8547--8555, Online and Punta Cana, Dominican Republic. Association for Computational Linguistics.

\bibitem[{Zhao et~al.(2021)Zhao, Wallace, Feng, Klein, and Singh}]{pmlr-v139-zhao21c}
Zihao Zhao, Eric Wallace, Shi Feng, Dan Klein, and Sameer Singh. 2021.
\newblock \href {https://proceedings.mlr.press/v139/zhao21c.html} {Calibrate before use: Improving few-shot performance of language models}.
\newblock In \emph{Proceedings of the 38th International Conference on Machine Learning}, volume 139 of \emph{Proceedings of Machine Learning Research}, pages 12697--12706. PMLR.

\bibitem[{Zheng et~al.(2021)Zheng, Zhang, Guo, Huang, Chen, Luo, and Chen}]{zheng-etal-2021-adaptive}
Xin Zheng, Zhirui Zhang, Junliang Guo, Shujian Huang, Boxing Chen, Weihua Luo, and Jiajun Chen. 2021.
\newblock \href {https://doi.org/10.18653/v1/2021.acl-short.47} {Adaptive nearest neighbor machine translation}.
\newblock In \emph{Proceedings of the 59th Annual Meeting of the Association for Computational Linguistics and the 11th International Joint Conference on Natural Language Processing (Volume 2: Short Papers)}, pages 368--374, Online. Association for Computational Linguistics.

\end{thebibliography}
\clearpage
\appendix

\addtolength{\tabcolsep}{-1pt}
\begin{table*}[h]
\setlength{\belowcaptionskip}{-0.5cm} 

\centering
\tiny
\begin{tabular}{l@{\hspace{1.5\tabcolsep}}l@{\hspace{1.3\tabcolsep}}ccccccc}
\toprule
                              $b$ & \textbf{Lang} &                                                        \textbf{ICL} &                                                  \textbf{ICL + CC} &                                                         \textbf{FT} &                                                        \textbf{PT} & \textbf{X-InSTA} & \textbf{X-InSTA*} &                                                                                   \textbf{N2C2} \\
\midrule
     \multirow{7}{*}{\textbf{2}}  &               De  &    $\text{28.06}_{\pm\text{3.2}}$ / $\text{28.00}_{\pm\text{8.0}}$  &   $\text{27.84}_{\pm\text{3.1}}$ / $\text{19.00}_{\pm\text{5.0}}$  &   $\text{21.28}_{\pm\text{0.4}}$ / $\text{53.92}_{\pm\text{18.4}}$  &   $\text{26.53}_{\pm\text{1.5}}$ / $\text{55.70}_{\pm\text{2.4}}$  &    20.36 / 71.65 &               27.84 / 80.54 &   $\text{\textbf{29.09}}_{\pm\text{\textbf{2.0}}}$ / $\text{\textbf{14.97}}_{\pm\text{\textbf{4.4}}}$ \\
                                  &               En  &    $\text{32.42}_{\pm\text{4.0}}$ / $\text{23.00}_{\pm\text{7.0}}$  &   $\text{31.16}_{\pm\text{3.8}}$ / $\text{22.00}_{\pm\text{5.0}}$  &   $\text{20.96}_{\pm\text{0.3}}$ / $\text{55.53}_{\pm\text{20.9}}$  &   $\text{31.70}_{\pm\text{1.9}}$ / $\text{56.72}_{\pm\text{2.0}}$  &    22.12 / 57.89 &               \textbf{43.66} / 69.51 &   $\text{32.56}_{\pm\text{2.7}}$ / $\text{\textbf{11.59}}_{\pm\text{\textbf{3.7}}}$ \\
                                  &               Es  &    $\text{27.49}_{\pm\text{2.6}}$ / $\text{28.00}_{\pm\text{9.0}}$  &   $\text{27.17}_{\pm\text{3.0}}$ / $\text{19.00}_{\pm\text{3.0}}$  &   $\text{22.24}_{\pm\text{0.6}}$ / $\text{55.54}_{\pm\text{18.6}}$  &   $\text{26.53}_{\pm\text{1.0}}$ / $\text{57.38}_{\pm\text{3.0}}$  &     21.50 / 75.59 &               \textbf{37.02} / 50.27 &   $\text{29.29}_{\pm\text{1.6}}$ / $\text{\textbf{13.45}}_{\pm\text{\textbf{4.5}}}$ \\
                                  &               Fr  &    $\text{25.90}_{\pm\text{2.5}}$ / $\text{32.00}_{\pm\text{9.0}}$  &   $\text{27.03}_{\pm\text{0.8}}$ / $\text{20.00}_{\pm\text{2.0}}$  &   $\text{20.97}_{\pm\text{0.9}}$ / $\text{58.55}_{\pm\text{15.9}}$  &   $\text{26.77}_{\pm\text{0.8}}$ / $\text{56.65}_{\pm\text{2.5}}$  &    20.06 / 55.56 &               \textbf{31.94} / 51.55 &   $\text{27.67}_{\pm\text{2.0}}$ / $\text{\textbf{17.95}}_{\pm\text{\textbf{5.3}}}$ \\
                                  &               Ja  &    $\text{24.18}_{\pm\text{1.8}}$ / $\text{37.00}_{\pm\text{8.0}}$  &   $\text{25.95}_{\pm\text{1.5}}$ / $\text{21.00}_{\pm\text{2.0}}$  &   $\text{20.42}_{\pm\text{0.4}}$ / $\text{49.85}_{\pm\text{26.5}}$  &   $\text{25.28}_{\pm\text{1.4}}$ / $\text{56.11}_{\pm\text{2.7}}$  &     19.80 / 49.84 &               26.82 / 21.85 &   $\text{\textbf{27.04}}_{\pm\text{\textbf{1.8}}}$ / $\text{\textbf{19.75}}_{\pm\text{\textbf{4.7}}}$ \\
                                  &               Zh  &    $\text{24.90}_{\pm\text{2.6}}$ / $\text{32.00}_{\pm\text{9.0}}$  &   $\text{25.57}_{\pm\text{2.2}}$ / $\text{22.00}_{\pm\text{3.0}}$  &   $\text{22.37}_{\pm\text{1.0}}$ / $\text{54.00}_{\pm\text{19.9}}$  &   $\text{22.06}_{\pm\text{0.9}}$ / $\text{59.60}_{\pm\text{2.7}}$  &    19.24 / 54.38 &               \textbf{36.14} / 39.79 &   $\text{26.79}_{\pm\text{1.8}}$ / $\text{\textbf{15.71}}_{\pm\text{\textbf{5.0}}}$ \\
                                  &          Avg.  &    $\text{27.16}_{\pm\text{2.7}}$ / $\text{30.00}_{\pm\text{4.4}}$  &   $\text{27.45}_{\pm\text{1.8}}$ / $\text{20.50}_{\pm\text{1.3}}$  &    $\text{21.37}_{\pm\text{0.7}}$ / $\text{54.57}_{\pm\text{2.6}}$  &   $\text{26.48}_{\pm\text{2.8}}$ / $\text{57.03}_{\pm\text{1.3}}$  &    20.51 / 60.82 &                \textbf{33.90} / 52.25 &   $\text{28.74}_{\pm\text{1.9}}$ / $\text{\textbf{15.57}}_{\pm\text{\textbf{2.7}}}$ \\
\midrule
     \multirow{7}{*}{\textbf{4}}  &               De  &    $\text{26.58}_{\pm\text{3.3}}$ / $\text{32.00}_{\pm\text{9.0}}$  &   $\text{28.64}_{\pm\text{2.5}}$ / $\text{22.00}_{\pm\text{6.0}}$  &   $\text{27.98}_{\pm\text{1.9}}$ / $\text{53.03}_{\pm\text{15.7}}$  &   $\text{26.57}_{\pm\text{1.5}}$ / $\text{62.82}_{\pm\text{6.1}}$  &    19.58 / 73.72 &               27.12 / 78.07 &    $\text{\textbf{34.13}}_{\pm\text{\textbf{3.0}}}$ / $\text{\textbf{7.94}}_{\pm\text{\textbf{2.8}}}$ \\
                                  &               En  &    $\text{31.58}_{\pm\text{4.8}}$ / $\text{25.00}_{\pm\text{8.0}}$  &   $\text{32.04}_{\pm\text{3.4}}$ / $\text{24.00}_{\pm\text{6.0}}$  &   $\text{26.15}_{\pm\text{1.4}}$ / $\text{51.00}_{\pm\text{23.7}}$  &   $\text{27.80}_{\pm\text{1.7}}$ / $\text{62.56}_{\pm\text{6.0}}$  &    22.24 / 56.39 &               \textbf{45.24} / 73.73 &    $\text{36.65}_{\pm\text{3.4}}$ / $\text{\textbf{5.73}}_{\pm\text{\textbf{2.4}}}$ \\
                                  &               Es  &   $\text{25.17}_{\pm\text{2.9}}$ / $\text{33.00}_{\pm\text{10.0}}$  &   $\text{29.41}_{\pm\text{2.6}}$ / $\text{22.00}_{\pm\text{4.0}}$  &   $\text{26.17}_{\pm\text{1.2}}$ / $\text{56.76}_{\pm\text{10.3}}$  &   $\text{24.65}_{\pm\text{2.2}}$ / $\text{65.00}_{\pm\text{6.7}}$  &    22.42 / 76.55 &               \textbf{37.08} / 49.88 &    $\text{32.13}_{\pm\text{2.8}}$ / $\text{\textbf{6.88}}_{\pm\text{\textbf{3.4}}}$ \\
                                  &               Fr  &    $\text{24.10}_{\pm\text{2.3}}$ / $\text{37.00}_{\pm\text{9.0}}$  &   $\text{28.85}_{\pm\text{1.7}}$ / $\text{20.00}_{\pm\text{4.0}}$  &   $\text{23.09}_{\pm\text{0.9}}$ / $\text{50.00}_{\pm\text{22.6}}$  &   $\text{25.18}_{\pm\text{1.8}}$ / $\text{64.33}_{\pm\text{6.4}}$  &     19.40 / 53.96 &               31.16 / 51.94 &    $\text{\textbf{31.84}}_{\pm\text{\textbf{3.1}}}$ / $\text{\textbf{9.66}}_{\pm\text{\textbf{4.2}}}$ \\
                                  &               Ja  &    $\text{22.77}_{\pm\text{1.6}}$ / $\text{39.00}_{\pm\text{8.0}}$  &   $\text{28.99}_{\pm\text{1.6}}$ / $\text{23.00}_{\pm\text{5.0}}$  &   $\text{20.51}_{\pm\text{0.2}}$ / $\text{61.60}_{\pm\text{13.0}}$  &   $\text{23.17}_{\pm\text{1.2}}$ / $\text{65.59}_{\pm\text{7.2}}$  &    19.62 / 50.73 &               27.14 / 22.11 &   $\text{\textbf{32.61}}_{\pm\text{\textbf{3.2}}}$ / $\text{\textbf{10.53}}_{\pm\text{\textbf{4.2}}}$ \\
                                  &               Zh  &   $\text{23.33}_{\pm\text{2.3}}$ / $\text{36.00}_{\pm\text{10.0}}$  &   $\text{27.97}_{\pm\text{2.0}}$ / $\text{25.00}_{\pm\text{6.0}}$  &   $\text{22.45}_{\pm\text{1.3}}$ / $\text{48.96}_{\pm\text{24.3}}$  &   $\text{23.67}_{\pm\text{2.3}}$ / $\text{65.64}_{\pm\text{7.3}}$  &    20.48 / 50.64 &               \textbf{36.68} / 39.34 &    $\text{32.50}_{\pm\text{2.9}}$ / $\text{\textbf{7.80}}_{\pm\text{\textbf{3.1}}}$ \\
                                  &          Avg.  &    $\text{25.59}_{\pm\text{3.0}}$ / $\text{33.67}_{\pm\text{4.5}}$  &   $\text{29.32}_{\pm\text{1.3}}$ / $\text{22.67}_{\pm\text{1.6}}$  &    $\text{24.39}_{\pm\text{2.6}}$ / $\text{53.56}_{\pm\text{4.4}}$  &   $\text{25.17}_{\pm\text{1.6}}$ / $\text{64.32}_{\pm\text{1.2}}$  &    20.62 / 60.33 &               \textbf{34.07} / 52.51 &    $\text{33.31}_{\pm\text{1.7}}$ / $\text{\textbf{8.09}}_{\pm\text{\textbf{1.6}}}$ \\
                                  \midrule
     \multirow{7}{*}{\textbf{8}}  &               De  &   $\text{27.38}_{\pm\text{4.0}}$ / $\text{29.00}_{\pm\text{10.0}}$  &   $\text{28.91}_{\pm\text{2.4}}$ / $\text{23.00}_{\pm\text{8.0}}$  &   $\text{25.70}_{\pm\text{0.7}}$ / $\text{53.83}_{\pm\text{17.2}}$  &   $\text{31.04}_{\pm\text{1.9}}$ / $\text{62.19}_{\pm\text{5.5}}$  &    19.68 / 71.66 &               26.48 / 76.32 &   $\text{\textbf{35.08}}_{\pm\text{\textbf{1.8}}}$ / $\text{\textbf{15.07}}_{\pm\text{\textbf{3.2}}}$ \\
                                  &               En  &    $\text{32.37}_{\pm\text{4.4}}$ / $\text{22.00}_{\pm\text{8.0}}$  &   $\text{32.81}_{\pm\text{2.8}}$ / $\text{24.00}_{\pm\text{8.0}}$  &    $\text{30.62}_{\pm\text{1.2}}$ / $\text{57.99}_{\pm\text{7.8}}$  &   $\text{33.10}_{\pm\text{1.1}}$ / $\text{61.34}_{\pm\text{4.4}}$  &     22.20 / 57.01 &               \textbf{45.14} / 77.55 &    $\text{39.18}_{\pm\text{1.3}}$ / $\text{\textbf{9.87}}_{\pm\text{\textbf{2.2}}}$ \\
                                  &               Es  &   $\text{25.92}_{\pm\text{3.5}}$ / $\text{30.00}_{\pm\text{11.0}}$  &   $\text{29.39}_{\pm\text{1.9}}$ / $\text{23.00}_{\pm\text{7.0}}$  &   $\text{29.20}_{\pm\text{1.7}}$ / $\text{51.60}_{\pm\text{20.9}}$  &   $\text{30.88}_{\pm\text{1.5}}$ / $\text{62.73}_{\pm\text{4.8}}$  &    22.22 / 65.56 &               \textbf{36.32} / 49.29 &   $\text{32.20}_{\pm\text{2.2}}$ / $\text{\textbf{17.79}}_{\pm\text{\textbf{5.4}}}$ \\
                                  &               Fr  &   $\text{24.71}_{\pm\text{2.6}}$ / $\text{34.00}_{\pm\text{10.0}}$  &   $\text{29.04}_{\pm\text{0.9}}$ / $\text{20.00}_{\pm\text{7.0}}$  &   $\text{30.49}_{\pm\text{1.5}}$ / $\text{50.95}_{\pm\text{22.4}}$  &   $\text{29.70}_{\pm\text{2.0}}$ / $\text{63.75}_{\pm\text{5.7}}$  &    19.84 / 60.13 &               31.04 / 51.19 &   $\text{\textbf{32.21}}_{\pm\text{\textbf{1.8}}}$ / $\text{\textbf{18.62}}_{\pm\text{\textbf{4.2}}}$ \\
                                  &               Ja  &    $\text{23.02}_{\pm\text{1.9}}$ / $\text{37.00}_{\pm\text{9.0}}$  &   $\text{29.51}_{\pm\text{1.4}}$ / $\text{23.00}_{\pm\text{6.0}}$  &   $\text{29.01}_{\pm\text{1.0}}$ / $\text{57.06}_{\pm\text{11.0}}$  &   $\text{27.88}_{\pm\text{2.1}}$ / $\text{64.67}_{\pm\text{6.0}}$  &    19.24 / 54.95 &               26.82 / 21.85 &   $\text{\textbf{34.43}}_{\pm\text{\textbf{1.5}}}$ / $\text{\textbf{15.49}}_{\pm\text{\textbf{2.3}}}$ \\
                                  &               Zh  &   $\text{23.82}_{\pm\text{2.8}}$ / $\text{34.00}_{\pm\text{11.0}}$  &   $\text{27.97}_{\pm\text{1.2}}$ / $\text{27.00}_{\pm\text{8.0}}$  &    $\text{26.53}_{\pm\text{0.8}}$ / $\text{59.77}_{\pm\text{6.7}}$  &   $\text{28.48}_{\pm\text{1.4}}$ / $\text{64.80}_{\pm\text{5.1}}$  &     22.30 / 52.66 &                \textbf{37.10} / 39.07 &   $\text{33.05}_{\pm\text{1.7}}$ / $\text{\textbf{13.69}}_{\pm\text{\textbf{2.6}}}$ \\
                                  &          Avg.  &    $\text{26.20}_{\pm\text{3.1}}$ / $\text{31.00}_{\pm\text{4.8}}$  &   $\text{29.61}_{\pm\text{1.5}}$ / $\text{23.33}_{\pm\text{2.0}}$  &    $\text{28.59}_{\pm\text{1.9}}$ / $\text{55.20}_{\pm\text{3.3}}$  &   $\text{30.18}_{\pm\text{1.7}}$ / $\text{63.25}_{\pm\text{1.3}}$  &    20.91 / 60.33 &               33.82 / 52.54 &   $\text{\textbf{34.36}}_{\pm\text{\textbf{2.4}}}$ / $\text{\textbf{15.09}}_{\pm\text{\textbf{2.9}}}$ \\
                                  \midrule
    \multirow{7}{*}{\textbf{16}}  &               De  &   $\text{27.36}_{\pm\text{4.0}}$ / $\text{27.00}_{\pm\text{10.0}}$  &   $\text{30.61}_{\pm\text{3.3}}$ / $\text{21.00}_{\pm\text{7.0}}$  &   $\text{36.92}_{\pm\text{2.6}}$ / $\text{48.69}_{\pm\text{17.4}}$  &   $\text{36.46}_{\pm\text{2.0}}$ / $\text{56.39}_{\pm\text{4.8}}$  &    19.68 / 64.07 &               25.62 / 76.07 &   $\text{\textbf{37.75}}_{\pm\text{\textbf{2.3}}}$ / $\text{\textbf{10.94}}_{\pm\text{\textbf{5.9}}}$ \\
                                  &               En  &    $\text{32.20}_{\pm\text{4.3}}$ / $\text{21.00}_{\pm\text{7.0}}$  &   $\text{32.71}_{\pm\text{2.6}}$ / $\text{17.00}_{\pm\text{5.0}}$  &   $\text{28.42}_{\pm\text{1.0}}$ / $\text{50.99}_{\pm\text{25.4}}$  &   $\text{37.00}_{\pm\text{1.5}}$ / $\text{56.78}_{\pm\text{3.9}}$  &    21.34 / 52.38 &                \textbf{44.00} / 76.48 &    $\text{42.85}_{\pm\text{1.4}}$ / $\text{\textbf{5.69}}_{\pm\text{\textbf{2.7}}}$ \\
                                  &               Es  &   $\text{26.07}_{\pm\text{3.4}}$ / $\text{28.00}_{\pm\text{10.0}}$  &   $\text{30.98}_{\pm\text{1.2}}$ / $\text{19.00}_{\pm\text{5.0}}$  &   $\text{29.56}_{\pm\text{1.7}}$ / $\text{57.26}_{\pm\text{11.7}}$  &   $\text{33.52}_{\pm\text{2.0}}$ / $\text{57.41}_{\pm\text{4.2}}$  &     22.90 / 54.18 &               \textbf{35.34} / 48.62 &   $\text{34.50}_{\pm\text{2.0}}$ / $\text{\textbf{12.88}}_{\pm\text{\textbf{5.4}}}$ \\
                                  &               Fr  &   $\text{24.70}_{\pm\text{2.5}}$ / $\text{31.00}_{\pm\text{10.0}}$  &   $\text{29.02}_{\pm\text{3.1}}$ / $\text{21.00}_{\pm\text{8.0}}$  &   $\text{32.09}_{\pm\text{1.6}}$ / $\text{51.14}_{\pm\text{16.4}}$  &   $\text{32.70}_{\pm\text{2.3}}$ / $\text{57.78}_{\pm\text{5.2}}$  &     19.50 / 63.03 &                30.52 / 47.9 &   $\text{\textbf{33.51}}_{\pm\text{\textbf{2.0}}}$ / $\text{\textbf{17.01}}_{\pm\text{\textbf{6.4}}}$ \\
                                  &               Ja  &    $\text{22.99}_{\pm\text{1.9}}$ / $\text{34.00}_{\pm\text{9.0}}$  &   $\text{29.78}_{\pm\text{2.9}}$ / $\text{18.00}_{\pm\text{8.0}}$  &   $\text{31.17}_{\pm\text{1.5}}$ / $\text{53.10}_{\pm\text{15.5}}$  &   $\text{31.25}_{\pm\text{3.1}}$ / $\text{59.47}_{\pm\text{7.2}}$  &     18.90 / 64.53 &               25.88 / 21.98 &    $\text{\textbf{38.72}}_{\pm\text{\textbf{2.3}}}$ / $\text{\textbf{9.36}}_{\pm\text{\textbf{4.8}}}$ \\
                                  &               Zh  &   $\text{23.64}_{\pm\text{2.5}}$ / $\text{31.00}_{\pm\text{10.0}}$  &   $\text{28.28}_{\pm\text{1.8}}$ / $\text{25.00}_{\pm\text{7.0}}$  &    $\text{32.53}_{\pm\text{1.8}}$ / $\text{57.10}_{\pm\text{6.5}}$  &   $\text{31.82}_{\pm\text{2.3}}$ / $\text{59.84}_{\pm\text{5.7}}$  &    23.16 / 55.45 &               \textbf{36.96} / 38.88 &    $\text{35.10}_{\pm\text{\textbf{2.1}}}$ / $\text{\textbf{9.49}}_{\pm\text{\textbf{4.5}}}$ \\
                                  &          Avg.  &    $\text{26.16}_{\pm\text{3.1}}$ / $\text{28.67}_{\pm\text{4.1}}$  &   $\text{30.23}_{\pm\text{1.4}}$ / $\text{20.17}_{\pm\text{2.6}}$  &    $\text{31.78}_{\pm\text{2.7}}$ / $\text{53.05}_{\pm\text{3.2}}$  &   $\text{33.79}_{\pm\text{2.2}}$ / $\text{57.94}_{\pm\text{1.3}}$  &    20.91 / 58.94 &               33.05 / 51.66 &   $\text{\textbf{37.07}}_{\pm\text{\textbf{3.2}}}$ / $\text{\textbf{10.89}}_{\pm\text{\textbf{3.5}}}$ \\
                                  \midrule
    \multirow{7}{*}{\textbf{32}}  &               De  &   $\text{27.55}_{\pm\text{4.0}}$ / $\text{19.00}_{\pm\text{10.0}}$  &   $\text{22.53}_{\pm\text{3.2}}$ / $\text{21.00}_{\pm\text{6.0}}$  &    $\text{38.78}_{\pm\text{1.5}}$ / $\text{55.72}_{\pm\text{2.7}}$  &   $\text{39.16}_{\pm\text{2.4}}$ / $\text{54.94}_{\pm\text{3.2}}$  &    19.92 / 60.24 &                24.54 / 75.80 &    $\text{\textbf{41.59}}_{\pm\text{\textbf{1.8}}}$ / $\text{\textbf{4.35}}_{\pm\text{\textbf{2.2}}}$ \\
                                  &               En  &    $\text{32.15}_{\pm\text{4.5}}$ / $\text{19.00}_{\pm\text{8.0}}$  &   $\text{22.57}_{\pm\text{2.4}}$ / $\text{18.00}_{\pm\text{4.0}}$  &    $\text{39.31}_{\pm\text{1.0}}$ / $\text{54.45}_{\pm\text{4.3}}$  &   $\text{40.51}_{\pm\text{1.9}}$ / $\text{54.56}_{\pm\text{2.5}}$  &    19.92 / 60.24 &               43.42 / 75.22 &    $\text{\textbf{44.79}}_{\pm\text{\textbf{1.2}}}$ / $\text{\textbf{2.72}}_{\pm\text{\textbf{1.3}}}$ \\
                                  &               Es  &    $\text{26.09}_{\pm\text{3.6}}$ / $\text{19.00}_{\pm\text{5.0}}$  &   $\text{22.90}_{\pm\text{1.2}}$ / $\text{18.00}_{\pm\text{5.0}}$  &    $\text{35.41}_{\pm\text{2.4}}$ / $\text{51.76}_{\pm\text{1.5}}$  &   $\text{37.68}_{\pm\text{2.1}}$ / $\text{55.28}_{\pm\text{3.1}}$  &     20.46 / 45.80 &                34.76 / 48.90 &    $\text{\textbf{38.09}}_{\pm\text{\textbf{2.2}}}$ / $\text{\textbf{4.95}}_{\pm\text{\textbf{3.2}}}$ \\
                                  &               Fr  &    $\text{24.59}_{\pm\text{2.8}}$ / $\text{16.00}_{\pm\text{7.0}}$  &   $\text{23.07}_{\pm\text{3.3}}$ / $\text{20.00}_{\pm\text{6.0}}$  &    $\text{36.19}_{\pm\text{1.9}}$ / $\text{57.00}_{\pm\text{4.5}}$  &   $\text{36.15}_{\pm\text{2.1}}$ / $\text{55.12}_{\pm\text{2.9}}$  &    23.68 / 50.68 &               30.24 / 45.78 &    $\text{\textbf{38.22}}_{\pm\text{\textbf{1.8}}}$ / $\text{\textbf{6.05}}_{\pm\text{\textbf{3.3}}}$ \\
                                  &               Ja  &    $\text{23.01}_{\pm\text{1.9}}$ / $\text{20.00}_{\pm\text{6.0}}$  &   $\text{21.88}_{\pm\text{3.1}}$ / $\text{19.00}_{\pm\text{7.0}}$  &   $\text{34.71}_{\pm\text{2.0}}$ / $\text{52.50}_{\pm\text{11.3}}$  &   $\text{35.91}_{\pm\text{3.0}}$ / $\text{56.89}_{\pm\text{4.0}}$  &    19.16 / 64.54 &                25.40 / 22.22 &    $\text{\textbf{41.77}}_{\pm\text{\textbf{1.9}}}$ / $\text{\textbf{5.23}}_{\pm\text{\textbf{2.9}}}$ \\
                                  &               Zh  &    $\text{23.81}_{\pm\text{2.6}}$ / $\text{21.00}_{\pm\text{9.0}}$  &   $\text{21.31}_{\pm\text{1.9}}$ / $\text{25.00}_{\pm\text{7.0}}$  &    $\text{36.99}_{\pm\text{2.1}}$ / $\text{53.03}_{\pm\text{6.4}}$  &   $\text{35.59}_{\pm\text{2.6}}$ / $\text{58.23}_{\pm\text{3.7}}$  &    18.32 / 69.26 &               37.06 / 39.02 &    $\text{\textbf{39.87}}_{\pm\text{\textbf{2.1}}}$ / $\text{\textbf{4.12}}_{\pm\text{\textbf{2.6}}}$ \\
                                  &          Avg.  &    $\text{26.20}_{\pm\text{3.0}}$ / $\text{19.00}_{\pm\text{1.5}}$  &   $\text{22.38}_{\pm\text{0.6}}$ / $\text{20.17}_{\pm\text{2.4}}$  &    $\text{36.90}_{\pm\text{1.7}}$ / $\text{54.08}_{\pm\text{1.8}}$  &   $\text{37.50}_{\pm\text{1.8}}$ / $\text{55.84}_{\pm\text{1.3}}$  &    22.04 / 49.73 &               32.57 / 51.16 &    $\text{\textbf{40.72}}_{\pm\text{\textbf{2.3}}}$ / $\text{\textbf{4.57}}_{\pm\text{\textbf{1.0}}}$ \\
\bottomrule
\end{tabular}

   \caption{Complete Main Results for MARC on all 6 languages. The X-InSTA* refer to the version with XGLM-7.5B as the base model.}
    \label{appendix:marc_results}
\end{table*}
\addtolength{\tabcolsep}{1pt}

\section{Measuring Calibration}
\label{app_ece}
\textbf{Expected Calibration Error} (ECE) is achieved by dividing predictions into $M$ bins $\{B_1, . . . , B_M\}$, based on their confidence values, and computing a weighted average of the accuracy-confidence difference within each bin:
\begin{equation}
\text{ECE} = \sum_{m=1}^{M} \frac{|B_{m}|}{N}\left\lvert \text{acc}(B_{m}) - \text{conf}(B_{m})\right\rvert
\end{equation}
where $N$ represents the number of prediction samples and $|B_m|$ denotes the count of samples in the $m$-th bin.
\section{Choice of Patterns and Verbalizers} \label{sec:appendix}

For both MARC and CLS datasets, we used 4 different patterns and verbalizers. Given a sentence $sent$, the used patterns are as follows:

\begin{tcolorbox}[colback=gray!20]
$sent$. It is <MASK>.
\end{tcolorbox}

\begin{tcolorbox}[colback=gray!20]
$sent$. A <MASK> one. 
\end{tcolorbox}

\begin{tcolorbox}[colback=gray!20]
$sent$. All in all, it is <MASK>.
\end{tcolorbox}

\begin{tcolorbox}[colback=gray!20]
$sent$. A <MASK> piece.
\end{tcolorbox}
\noindent with ``inferior'', ``bad'', ``good'' and ``great'' as the verbalizer for rating 1-4 in CLS. In case of MARC with five labels, we expand it by imposing ``fine'' between the ``bad'' and 
``good'' labels.
\section{Datasets and Experimental Setup}
\label{dataset}
\textbf{Datasets}  
The MARC dataset is a large-scale multilingual text classification dataset from Amazon reviews (1-5). The CLS dataset consists of reviews on DVDs, music and books which are all multilingual and have four different ratings (1-4). 
% \vic{\textbf{English} is our source language, which is used for
% the training and development sets. All available languages  in the datasets are used for testing. 
For these datasets, the task is to predict the star rating given by the customers to a product based on reviews of the product (from 1 to 5 stars, higher stars mean higher satisfaction). For MARC, each  language has a test set of size 5k, and CLS has 2k test examples for each language. Both datasets consist of reviews in German, English, French and Japanese; the MARC dataset in addition includes Spanish and Mandarin.
For the $b$-shot experiments, where $b \in \{2, 4, 8, 16, 32\}$, we randomly sample $b$ cases from each category, i.e. we will have $b \times n$ cases in total ($n = 5$ for MARC and $n = 4$ for CLS).

\smallskip\noindent \textbf{Experimental Setup}
\vic{N2C2 is based on \texttt{XLM-RoberTa-base} model \cite{conneau-etal-2020-unsupervised}, which is a widely used multilingual pretrained language model. All  experiments were run 5 times with different random seeds \{1, 10, 100, 1000, 10000\} } \vic{During the training process of N2C2, we utilized the Adam optimizer \cite{kingma2014adam}  with a learning rate of 1e-3. The batch size was set to 32, and the total number of epochs was set to 10. We leveraged the development set to select the best checkpoints for testing. The temperature parameter $\tau$ was set to 5, and the maximum number of retrieved neighbors $K$ was determined based on the experimental results in Section \ref{in_topk}. For the MARC dataset, $K_{max}$ was set to 16, while for the CLS dataset, $K_{max}$ was 12. This means $R_{s} = \{0, 4, 8, \dots, K_{max}\}$}.   The size of Z was informed by the nature of few-shot experiments, where the training dataset is relatively small, necessitating a lighter network design to avoid severe overfitting. Thus, we selected Z=32 to reduce the number of parameters needed to be trained.

\section{ Experiment Results} \label{app:experiments}
\noindent

\smallskip\noindent \textbf{Main Results: Detailed Analysis}
\vic{We next provide a detailed analysis with each of the baselines. \textbf{Comparing the ICL and ICL+CC methods}, it can be observed that ICL+CC achieves higher accuracy and lower ECE than ICL in the MARC dataset. However, in the CLS dataset, ICL+CC has a significantly higher ECE compared to the ICL method. This indicates that ICL is not universally effective in cross-lingual scenarios.  Additionally, we notice that ICL's performance in CLS languages other than English is almost random. %This could be due to the model's inability to perform ICL in a cross-lingual manner for these languages. 
We note that previous  studies already observed such phenomena in monolingual ICL \cite{webson-pavlick-2022-prompt}.
%; cross-lingual ICL has its added nuances that make it even more difficult.
However, N2C2 continues to improve performance based on ICL as the data size increases. Furthermore, we observe that the performance of ICL and ICL+CC remains relatively stable as the data increases. This may be attributed to the input length limitation and the 1-shot ensemble approach in ICL, where increasing the data does not provide significant assistance to cross-lingual ICL. 
A key difference is that N2C2 benefits from an increased number of examples belonging to the same class, so it is more likely that the predicted instances are examples from the same class. This results in consistent  improvements of N2C2 in accuracy and ECE.
}
\begin{figure}[t]
\setlength{\belowcaptionskip}{-0.5cm} 

    \centering
    \includegraphics[width=1.0\linewidth]{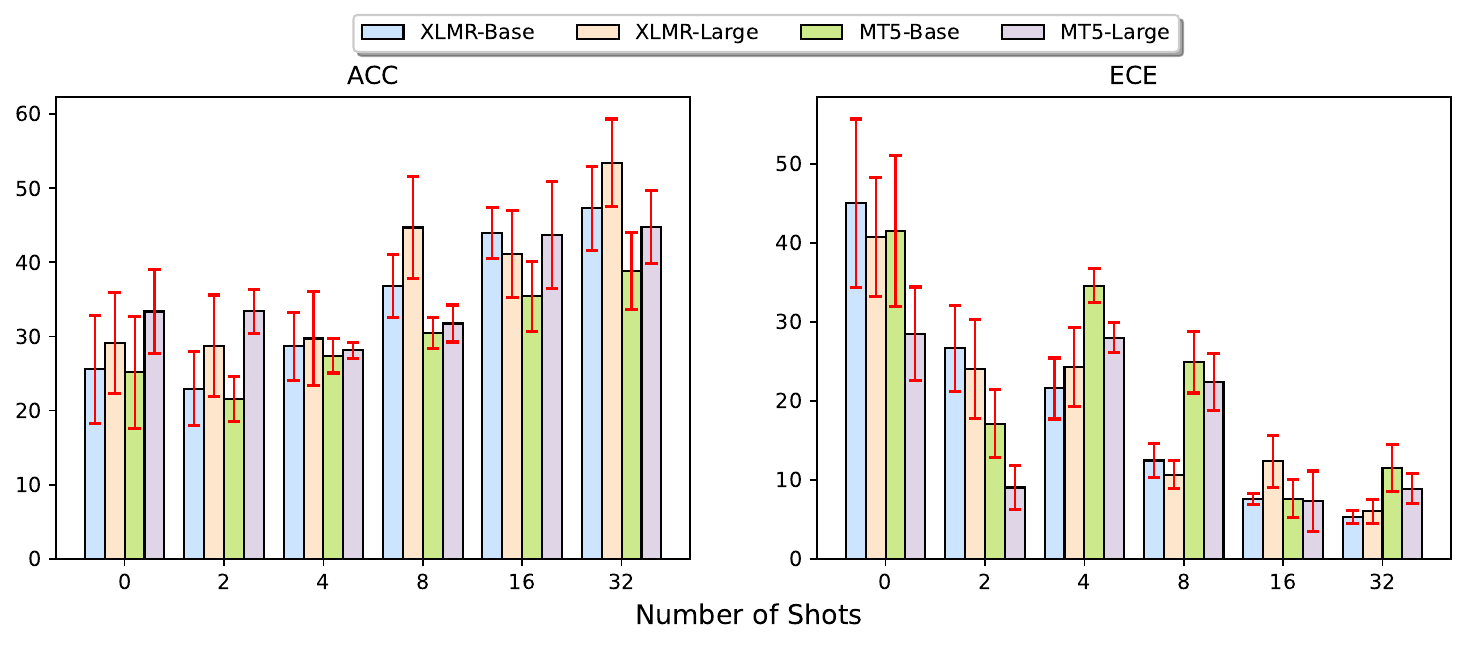}
    \caption{Comparison with different model sizes and architectures in CLS.}
\label{cls_model}
\end{figure}

\inlineSubsection{Comparing prompt tuning and fine tuning methods,}involves updating the parameters of the multilingual pretrained language model itself, prompt tuning performs better in terms of accuracy compared to fine tuning. This is not surprising as prompt tuning benefits from having labelled  information, which gives it an advantage over randomly initialized fine tuning methods, especially on tasks with relatively small training data \cite{wang-etal-2022-towards-unified}. However, surprisingly, prompt tuning exhibits a higher calibration error compared to fine tuning methods. We believe that this might be because prompt tuning, in the case of few-shot updates, may not fully understand the meaning of the verbalizer words and updating them might introduce noise. Regarding fine tuning and prompt tuning, both methods have high ECE values, averaging higher than our proposed method by 33.89\%.
\smallskip
\inlineSubsection{Effect of  Model Size and Architecture} Figure~\ref{cls_model} presents additional results for the CLS dataset.
\smallskip
% \inlineSubsection{Cross-Domain Learning Evaluation} 
\section{Cross-Domain Learning Evaluation}
\vic{Due to insufficient examples in the book and music domains within the CLS dataset to reach 32 shots across all languages, we focused solely on the DVD domain for our main experiments. We evaluated the performance of N2C2 when \simon{the in-context samples and datastores are crafted} from the DVD domain, and \simon{test it on data} from the book and music domains under 16 shots. In this setup, we compared N2C2 with baselines that are trained on the \simon{in-domain data, from music and book domains}.  

Table~\ref{tlcls_music} presents the results on the music domain. It shows that ICL and ICL+CC perform poorly as expected. \simon{Nonetheless}, N2C2 showcases notable superiority, outperforming fine-tuning and prompt tuning by 13\% and 15\% respectively. Moreover, N2C2 consistently exhibits significantly better ECE performance compared to other baselines. %We believe this is because the music domain has less content in the multilingual pretraining language model corpus than the book domain.  
These results show the robustness of N2C2 in cross-domain scenarios. Results for the book domain can be seen in App.\ \ref{app:experiments}, Table~\ref{tlcls_book}.}

Table~\ref{tlcls_book} presents results for the book domain. In this domain, fine-tuning and prompt tuning show better results than our method by 3\% and 5\%, respectively.  Recall that on the music domain, our method outperforms them by 13\% and 15\%.  We speculate that this  difference is because the music domain has less content in the multilingual pretraining language model corpus, while the book domain has more.

\begin{table}[]
\setlength{\belowcaptionskip}{-0.5cm} 

    \small
    \centering
    \begin{tabular}{c|cc}
\bottomrule
       $b$ shots  & W/o 
$k$NN training &$k$NN training \\
\hline
        2 & 0.021s & 0.0248s \\
\hline
        4 &	0.045s & 0.0543s\\ 
\hline
        8 & 0.102s& 0.106s\\ 
\hline
        16 & 0.178s	&0.189s \\ 
\hline

        32& 0.236s	&0.252s \\
\toprule
    \end{tabular}
    \caption{Run time comparison when b = 2, 4, 8, 16, 32}
    \label{tab:eff}
\end{table}

\section{Efficiency}
Since we are dealing with a few-shot problem where the training set is small, we directly use the L2 distance for calculation without employing efficient retrieval methods like FAISS. We only add some parameters that are merely simple linear transformation layers during the retrieval process. Table \ref{tab:eff}  illustrates the average time required for inference for each example during the inference stage.

\renewcommand{\arraystretch}{1.02}
\addtolength{\tabcolsep}{-1pt}
\begin{table*}[]
\centering
\tiny
\begin{tabular}{l@{\hspace{1.3\tabcolsep}}l@{\hspace{1.3\tabcolsep}}ccccccc}
\toprule
                             $b$ & \textbf{Lang} &                                                        \textbf{ICL} &                                                   \textbf{ICL + CC} &                                                          \textbf{FT} &                                                              \textbf{PT} & \textbf{X-InSTA} & \textbf{X-InSTA*} &                                                                                   \textbf{N2C2} \\
                                                              \midrule

    \multirow{5}{*}{\textbf{2}}  &               De  &    $\text{20.07}_{\pm\text{4.1}}$ / $\text{35.00}_{\pm\text{7.0}}$  &   $\text{22.76}_{\pm\text{5.4}}$ / $\text{43.00}_{\pm\text{10.0}}$  &    $\text{21.60}_{\pm\text{3.1}}$ / $\text{41.93}_{\pm\text{25.2}}$  &   $\text{29.87}_{\pm\text{11.3}}$ / $\text{64.47}_{\pm\text{16.1}}$  &     21.30 / 65.58 &               \textbf{31.25} / 71.05 &                     $\text{20.89}_{\pm\text{3.1}}$ / $\text{\textbf{28.77}}_{\pm\text{\textbf{5.5}}}$ \\
                                 &               En  &   $\text{29.55}_{\pm\text{8.2}}$ / $\text{27.00}_{\pm\text{10.0}}$  &   $\text{30.23}_{\pm\text{7.4}}$ / $\text{37.00}_{\pm\text{11.0}}$  &    $\text{25.03}_{\pm\text{6.6}}$ / $\text{40.57}_{\pm\text{21.7}}$  &    $\text{33.18}_{\pm\text{8.2}}$ / $\text{62.13}_{\pm\text{11.7}}$  &     15.40 / 49.19 &               \textbf{52.15} / 73.01 &                     $\text{30.46}_{\pm\text{5.3}}$ / $\text{\textbf{18.50}}_{\pm\text{\textbf{5.9}}}$ \\
                                 &               Fr  &    $\text{18.41}_{\pm\text{3.6}}$ / $\text{40.00}_{\pm\text{7.0}}$  &    $\text{22.49}_{\pm\text{5.2}}$ / $\text{43.00}_{\pm\text{7.0}}$  &    $\text{27.53}_{\pm\text{5.2}}$ / $\text{40.99}_{\pm\text{24.4}}$  &    $\text{31.85}_{\pm\text{9.2}}$ / $\text{62.41}_{\pm\text{14.2}}$  &     \textbf{34.20} / 50.61 &                33.10 / 44.36 &                     $\text{20.51}_{\pm\text{3.1}}$ / $\text{\textbf{29.43}}_{\pm\text{\textbf{6.2}}}$ \\
                                 &               Jp  &    $\text{18.86}_{\pm\text{2.0}}$ / $\text{43.00}_{\pm\text{4.0}}$  &    $\text{21.33}_{\pm\text{3.3}}$ / $\text{48.00}_{\pm\text{6.0}}$  &    $\text{25.25}_{\pm\text{2.3}}$ / $\text{45.11}_{\pm\text{26.9}}$  &    $\text{32.42}_{\pm\text{6.1}}$ / $\text{60.95}_{\pm\text{11.6}}$  &     28.00 / 53.19 &                \textbf{36.00} / 26.84 &                     $\text{19.99}_{\pm\text{1.8}}$ / $\text{\textbf{29.96}}_{\pm\text{\textbf{3.5}}}$ \\
                                 &          Avg.  &    $\text{21.72}_{\pm\text{4.6}}$ / $\text{36.25}_{\pm\text{6.1}}$  &    $\text{24.20}_{\pm\text{3.5}}$ / $\text{42.75}_{\pm\text{3.9}}$  &     $\text{24.85}_{\pm\text{2.1}}$ / $\text{42.15}_{\pm\text{1.8}}$  &     $\text{31.83}_{\pm\text{1.2}}$ / $\text{62.49}_{\pm\text{1.3}}$  &    24.72 / 54.64 &               \textbf{38.12} / 53.82 &                     $\text{22.96}_{\pm\text{4.3}}$ / $\text{\textbf{26.66}}_{\pm\text{\textbf{4.7}}}$ \\
                                 \midrule
    \multirow{5}{*}{\textbf{4}}  &               De  &    $\text{22.61}_{\pm\text{3.9}}$ / $\text{22.00}_{\pm\text{7.0}}$  &   $\text{24.24}_{\pm\text{4.5}}$ / $\text{41.00}_{\pm\text{14.0}}$  &    $\text{36.50}_{\pm\text{4.7}}$ / $\text{47.59}_{\pm\text{18.6}}$  &     $\textbf{39.56}_{\pm\textbf{6.0}}$ / $\text{52.83}_{\pm\text{7.6}}$  &     21.20 / 67.93 &                 30.40 / 67.37 &                     $\text{27.07}_{\pm\text{1.6}}$ / $\text{\textbf{21.15}}_{\pm\text{\textbf{3.2}}}$ \\
                                 &               En  &    $\text{29.60}_{\pm\text{7.6}}$ / $\text{22.00}_{\pm\text{9.0}}$  &   $\text{27.10}_{\pm\text{4.0}}$ / $\text{38.00}_{\pm\text{13.0}}$  &    $\text{40.85}_{\pm\text{2.3}}$ / $\text{40.72}_{\pm\text{12.0}}$  &     $\text{37.59}_{\pm\text{4.8}}$ / $\text{56.13}_{\pm\text{6.5}}$  &     15.25 / 47.5 &                \textbf{54.05} / 75.57 &                     $\text{35.34}_{\pm\text{4.2}}$ / $\text{\textbf{15.88}}_{\pm\text{\textbf{5.8}}}$ \\
                                 &               Fr  &    $\text{21.78}_{\pm\text{3.0}}$ / $\text{26.00}_{\pm\text{6.0}}$  &   $\text{24.02}_{\pm\text{3.4}}$ / $\text{40.00}_{\pm\text{13.0}}$  &     $\text{41.18}_{\pm\text{2.8}}$ / $\text{46.07}_{\pm\text{7.5}}$  &     $\textbf{39.44}_{\pm\textbf{4.7}}$ / $\text{52.89}_{\pm\text{6.5}}$  &     34.90 / 53.75 &                34.85 / 42.69 &                     $\text{27.29}_{\pm\text{2.0}}$ / $\text{\textbf{22.55}}_{\pm\text{\textbf{2.6}}}$ \\
                                 &               Jp  &    $\text{19.99}_{\pm\text{2.0}}$ / $\text{29.00}_{\pm\text{4.0}}$  &   $\text{22.66}_{\pm\text{3.4}}$ / $\text{43.00}_{\pm\text{13.0}}$  &    $\text{38.42}_{\pm\text{0.8}}$ / $\text{44.89}_{\pm\text{16.4}}$  &     $\textbf{37.48}_{\pm\textbf{4.6}}$ / $\text{54.07}_{\pm\text{7.2}}$  &     28.30 / 58.76 &               37.15 / 26.59 &                     $\text{25.02}_{\pm\text{2.3}}$ / $\text{\textbf{23.76}}_{\pm\text{\textbf{3.9}}}$ \\
                                 &          Avg.  &    $\text{23.50}_{\pm\text{3.6}}$ / $\text{24.75}_{\pm\text{3.0}}$  &    $\text{24.50}_{\pm\text{1.6}}$ / $\text{40.50}_{\pm\text{1.8}}$  &     $\text{39.24}_{\pm\text{1.9}}$ / $\text{44.82}_{\pm\text{2.5}}$  &     $\text{38.52}_{\pm\text{1.0}}$ / $\text{53.98}_{\pm\text{1.3}}$  &    24.91 / 56.98 &               \textbf{39.11} / 53.56 &                     $\text{28.68}_{\pm\text{4.0}}$ / $\text{\textbf{20.84}}_{\pm\text{\textbf{3.0}}}$ \\
\midrule

    \multirow{5}{*}{\textbf{8}}  &               De  &    $\text{21.60}_{\pm\text{5.2}}$ / $\text{21.00}_{\pm\text{7.0}}$  &   $\text{25.57}_{\pm\text{6.1}}$ / $\text{41.00}_{\pm\text{18.0}}$  &    $\text{25.89}_{\pm\text{4.7}}$ / $\text{44.15}_{\pm\text{18.0}}$  &         $\text{29.05}_{\pm\text{2.1}}$ / $\text{49.69}_{\pm\text{1.2}}$  &     21.05 / 68.20 &                30.00 / 67.08 &                 $\textbf{35.76}_{\pm\textbf{3.1}}$ / $\text{\textbf{13.24}}_{\pm\text{\textbf{4.7}}}$ \\
                                 &               En  &    $\text{27.57}_{\pm\text{7.1}}$ / $\text{17.00}_{\pm\text{8.0}}$  &   $\text{27.23}_{\pm\text{5.1}}$ / $\text{39.00}_{\pm\text{18.0}}$  &     $\text{37.24}_{\pm\text{2.9}}$ / $\text{38.95}_{\pm\text{9.5}}$  &         $\text{38.69}_{\pm\text{2.3}}$ / $\text{48.51}_{\pm\text{1.4}}$  &    15.15 / 46.36 &                 \textbf{52.60} / 76.30 &                  $\text{42.85}_{\pm\text{4.7}}$ / $\text{\textbf{9.30}}_{\pm\text{\textbf{4.1}}}$ \\
                                 &               Fr  &    $\text{18.53}_{\pm\text{3.3}}$ / $\text{26.00}_{\pm\text{5.0}}$  &   $\text{24.20}_{\pm\text{6.0}}$ / $\text{44.00}_{\pm\text{18.0}}$  &    $\text{34.69}_{\pm\text{2.0}}$ / $\text{44.83}_{\pm\text{19.4}}$  &         $\text{33.09}_{\pm\text{2.7}}$ / $\text{49.12}_{\pm\text{2.5}}$  &    35.25 / 55.33 &                \textbf{35.90} / 42.65 &                 $\text{35.58}_{\pm\text{3.3}}$ / $\text{\textbf{13.22}}_{\pm\text{\textbf{4.3}}}$ \\
                                 &               Jp  &    $\text{18.47}_{\pm\text{2.0}}$ / $\text{28.00}_{\pm\text{3.0}}$  &   $\text{25.25}_{\pm\text{5.5}}$ / $\text{44.00}_{\pm\text{19.0}}$  &    $\text{28.48}_{\pm\text{4.5}}$ / $\text{39.82}_{\pm\text{18.2}}$  &         $\text{31.31}_{\pm\text{1.1}}$ / $\text{51.73}_{\pm\text{1.7}}$  &    25.95 / 60.12 &                \textbf{36.50} / 26.07 &                 $\text{32.92}_{\pm\text{2.0}}$ / $\text{\textbf{14.08}}_{\pm\text{\textbf{4.9}}}$ \\

                                 &          Avg.  &    $\text{21.54}_{\pm\text{3.7}}$ / $\text{23.00}_{\pm\text{4.3}}$  &    $\text{25.56}_{\pm\text{1.1}}$ / $\text{42.00}_{\pm\text{2.1}}$  &     $\text{31.57}_{\pm\text{4.6}}$ / $\text{41.94}_{\pm\text{2.6}}$  &         $\text{33.04}_{\pm\text{1.6}}$ / $\text{49.76}_{\pm\text{1.2}}$  &     24.35 / 57.50 &               \textbf{38.75} / 53.02 &                 $\text{36.78}_{\pm\text{3.7}}$ / $\text{\textbf{12.46}}_{\pm\text{\textbf{1.9}}}$ \\
                                                                                                   \midrule

   \multirow{5}{*}{\textbf{16}}  &               De  &    $\text{20.57}_{\pm\text{4.5}}$ / $\text{35.00}_{\pm\text{8.0}}$  &   $\text{21.31}_{\pm\text{4.7}}$ / $\text{42.00}_{\pm\text{11.0}}$  &     $\text{27.66}_{\pm\text{10.2}}$ / $\text{7.30}_{\pm\text{1.9}}$  &         $\text{40.23}_{\pm\text{4.8}}$ / $\text{55.37}_{\pm\text{5.1}}$  &     21.20 / 67.88 &               30.35 / 66.74 &    $\text{\textbf{46.15}}_{\pm\text{\textbf{6.9}}}$ / $\text{\textbf{7.84}}_{\pm\text{\textbf{4.0}}}$ \\
                                 &               En  &   $\text{27.32}_{\pm\text{7.2}}$ / $\text{31.00}_{\pm\text{10.0}}$  &   $\text{25.10}_{\pm\text{4.3}}$ / $\text{39.00}_{\pm\text{10.0}}$  &     $\text{42.03}_{\pm\text{2.0}}$ / $\text{50.20}_{\pm\text{4.6}}$  &         $\text{44.83}_{\pm\text{4.7}}$ / $\text{51.44}_{\pm\text{5.0}}$  &    15.15 / 46.41 &               \textbf{52.55} / 74.96 &    $\text{\text{47.64}}_{\pm\text{\text{2.5}}}$ / $\text{\textbf{6.56}}_{\pm\text{\textbf{2.6}}}$ \\
                                 &               Fr  &    $\text{18.67}_{\pm\text{3.2}}$ / $\text{40.00}_{\pm\text{7.0}}$  &   $\text{21.80}_{\pm\text{5.0}}$ / $\text{43.00}_{\pm\text{11.0}}$  &   $\text{39.31}_{\pm\text{10.1}}$ / $\text{49.11}_{\pm\text{22.9}}$  &         $\text{38.48}_{\pm\text{4.4}}$ / $\text{57.19}_{\pm\text{5.1}}$  &      35.60 / 55.20 &                34.90 / 42.62 &    $\text{\textbf{41.64}}_{\pm\text{\textbf{5.0}}}$ / $\text{\textbf{8.00}}_{\pm\text{\textbf{4.6}}}$ \\
                                 &               Jp  &    $\text{18.90}_{\pm\text{2.9}}$ / $\text{43.00}_{\pm\text{6.0}}$  &   $\text{22.41}_{\pm\text{4.2}}$ / $\text{43.00}_{\pm\text{12.0}}$  &    $\text{35.00}_{\pm\text{4.7}}$ / $\text{43.31}_{\pm\text{21.1}}$  &         $\text{39.86}_{\pm\text{4.3}}$ / $\text{52.23}_{\pm\text{4.7}}$  &    27.75 / 60.58 &               36.45 / 26.07 &    $\text{\textbf{40.52}}_{\pm\text{\textbf{3.2}}}$ / $\text{\textbf{8.05}}_{\pm\text{\textbf{2.9}}}$ \\
                                 &          Avg.  &    $\text{21.37}_{\pm\text{3.5}}$ / $\text{37.25}_{\pm\text{4.6}}$  &    $\text{22.65}_{\pm\text{1.5}}$ / $\text{41.75}_{\pm\text{1.6}}$  &    $\text{36.00}_{\pm\text{5.4}}$ / $\text{37.48}_{\pm\text{17.6}}$  &         $\text{40.85}_{\pm\text{2.4}}$ / $\text{54.06}_{\pm\text{2.3}}$  &    24.92 / 57.52 &                38.56 / 52.60 &    $\text{\textbf{43.99}}_{\pm\text{\textbf{3.0}}}$ / $\text{\textbf{7.61}}_{\pm\text{\textbf{0.6}}}$ \\
                                 \midrule
   \multirow{5}{*}{\textbf{32}}  &               De  &    $\text{20.59}_{\pm\text{4.8}}$ / $\text{32.00}_{\pm\text{8.0}}$  &    $\text{19.24}_{\pm\text{3.3}}$ / $\text{33.00}_{\pm\text{7.0}}$  &    $\text{44.19}_{\pm\text{5.7}}$ / $\text{43.00}_{\pm\text{20.1}}$  &         $\text{45.40}_{\pm\text{4.2}}$ / $\text{49.23}_{\pm\text{5.2}}$  &    21.15 / 67.93 &                30.60 / 66.43 &    $\text{\textbf{46.12}}_{\pm\text{\textbf{3.5}}}$ / $\text{\textbf{5.76}}_{\pm\text{\textbf{2.1}}}$ \\
                                 &               En  &    $\text{26.69}_{\pm\text{6.9}}$ / $\text{28.00}_{\pm\text{9.0}}$  &    $\text{22.53}_{\pm\text{4.3}}$ / $\text{31.00}_{\pm\text{6.0}}$  &    $\text{51.27}_{\pm\text{5.5}}$ / $\text{36.07}_{\pm\text{11.2}}$  &         $\text{48.98}_{\pm\text{2.6}}$ / $\text{46.62}_{\pm\text{3.3}}$  &    15.15 / 45.12 &               51.45 / 74.91 &    $\text{\textbf{55.63}}_{\pm\text{\textbf{2.0}}}$ / $\text{\textbf{4.50}}_{\pm\text{\textbf{1.3}}}$ \\
                                 &               Fr  &    $\text{18.67}_{\pm\text{3.3}}$ / $\text{37.00}_{\pm\text{7.0}}$  &    $\text{20.24}_{\pm\text{3.9}}$ / $\text{32.00}_{\pm\text{7.0}}$  &    $\text{27.04}_{\pm\text{14.6}}$ / $\text{14.07}_{\pm\text{2.6}}$  &         $\text{40.54}_{\pm\text{3.3}}$ / $\text{51.42}_{\pm\text{4.1}}$  &    35.65 / 55.61 &                35.30 / 40.61 &    $\text{\textbf{43.73}}_{\pm\text{\textbf{3.7}}}$ / $\text{\textbf{4.83}}_{\pm\text{\textbf{3.4}}}$ \\
                                 &               Jp  &    $\text{19.03}_{\pm\text{2.9}}$ / $\text{39.00}_{\pm\text{5.0}}$  &    $\text{21.67}_{\pm\text{3.7}}$ / $\text{31.00}_{\pm\text{8.0}}$  &    $\text{37.32}_{\pm\text{6.5}}$ / $\text{48.40}_{\pm\text{19.7}}$  &         $\text{42.76}_{\pm\text{2.4}}$ / $\text{50.06}_{\pm\text{3.9}}$  &     25.95 / 60.60 &                35.80 / 25.48 &    $\text{\textbf{43.59}}_{\pm\text{\textbf{2.5}}}$ / $\text{\textbf{6.17}}_{\pm\text{\textbf{4.4}}}$ \\
                                 &          Avg.  &    $\text{21.25}_{\pm\text{3.2}}$ / $\text{34.00}_{\pm\text{4.3}}$  &    $\text{20.92}_{\pm\text{1.3}}$ / $\text{31.75}_{\pm\text{0.8}}$  &    $\text{39.95}_{\pm\text{8.9}}$ / $\text{35.38}_{\pm\text{13.1}}$  &         $\text{44.42}_{\pm\text{3.1}}$ / $\text{49.33}_{\pm\text{1.8}}$  &    24.48 / 57.32 &               38.29 / 51.86 &    $\text{\textbf{47.27}}_{\pm\text{\textbf{4.9}}}$ / $\text{\textbf{5.31}}_{\pm\text{\textbf{0.7}}}$ \\
\bottomrule
\end{tabular}
   \caption{Complete Main Results for CLS on all 4 languages. The X-InSTA* refer to the version with XGLM-7.5B as the base model.}
    \label{appendix:cls_result}
\end{table*}
\addtolength{\tabcolsep}{1pt}
\begin{table*}[]
    \centering
    \tiny
\begin{tabular}{lccccc}
\toprule
\textbf{Methods} &                                                                 \textbf{De} &                                                                 \textbf{En} &                                                                 \textbf{Fr} &                                                                
\textbf{Jp} &                                                           \textbf{Avg.} \\
\midrule
ICL         &   $\text{18.50}_{\pm\text{5.6}}$ / $\text{37.00}_{\pm\text{10.0}}$ &  $\text{31.16}_{\pm\text{11.0}}$ / $\text{25.00}_{\pm\text{13.0}}$ &    $\text{16.95}_{\pm\text{3.6}}$ / $\text{41.00}_{\pm\text{8.0}}$ &   $\text{14.42}_{\pm\text{2.0}}$ / $\text{48.00}_{\pm\text{6.0}}$ &   $\text{20.26}_{\pm\text{7.5}}$ / $\text{37.75}_{\pm\text{9.6}}$ \\
ICL + CC    &   $\text{21.30}_{\pm\text{6.5}}$ / $\text{41.00}_{\pm\text{11.0}}$ &   $\text{25.58}_{\pm\text{5.3}}$ / $\text{38.00}_{\pm\text{14.0}}$ &   $\text{21.12}_{\pm\text{5.4}}$ / $\text{43.00}_{\pm\text{12.0}}$ &  $\text{19.83}_{\pm\text{4.8}}$ / $\text{46.00}_{\pm\text{12.0}}$ &   $\text{21.96}_{\pm\text{2.5}}$ / $\text{42.00}_{\pm\text{3.4}}$ \\
FT   &   $\text{41.28}_{\pm\text{6.8}}$ / $\text{52.15}_{\pm\text{10.0}}$ &    $\text{46.78}_{\pm\text{4.0}}$ / $\text{48.69}_{\pm\text{5.7}}$ &     $\text{26.39}_{\pm\text{7.3}}$ / $\text{6.24}_{\pm\text{4.0}}$ &  $\text{29.96}_{\pm\text{4.9}}$ / $\text{48.97}_{\pm\text{26.3}}$ &  $\text{36.10}_{\pm\text{9.5}}$ / $\text{39.01}_{\pm\text{21.9}}$ \\
PT &    $\text{34.35}_{\pm\text{4.6}}$ / $\text{57.15}_{\pm\text{9.6}}$ &    $\text{35.94}_{\pm\text{3.0}}$ / $\text{57.13}_{\pm\text{8.4}}$ &    $\text{34.14}_{\pm\text{4.7}}$ / $\text{57.71}_{\pm\text{8.2}}$ &   $\text{32.31}_{\pm\text{4.1}}$ / $\text{58.26}_{\pm\text{9.9}}$ &   $\text{34.18}_{\pm\text{1.5}}$ / $\text{57.56}_{\pm\text{0.5}}$ \\
N2C2         &     $\textbf{50.28}_{\pm\textbf{5.8}}$ / $\textbf{7.55}_{\pm\textbf{2.4}}$ &     $\textbf{54.56}_{\pm\textbf{1.9}}$ / $\textbf{5.00}_{\pm\textbf{2.2}}$ &     $\textbf{46.96}_{\pm\textbf{3.4}}$ / $\textbf{6.05}_{\pm\textbf{2.4}}$ &    $\textbf{47.60}_{\pm\textbf{3.2}}$ / $\textbf{5.40}_{\pm\textbf{1.5}}$ &    $\textbf{49.85}_{\pm\textbf{3.5}}$ / $\textbf{6.00}_{\pm\textbf{1.1}}$ \\
\bottomrule
\end{tabular}
    \caption{
    Experimental results for the cross-domain condition.  Results are reported as $\textbf{Accuracy}\uparrow(\%)$ / $\textbf{ECE}\downarrow(\%)$. For N2C2, We use the DVD domain as the training and validation set, while for other baselines, we use the Music domain as the training and validation set. The test set is uniformly composed of data from the \textbf{Music} domain. }
    \label{tlcls_music}
\vspace{-0.3cm}
\end{table*}
\begin{table*}[]
    \centering
    \scriptsize
\begin{tabular}{lccccc}
\toprule
\textbf{Method} &                                                                \textbf{De} &                                                                 \textbf{En} &                                                                \textbf{Fr} &                                                                \textbf{Jp} &                                                            \textbf{Avg.} \\
\midrule
ICL         &   $\text{21.04}_{\pm\text{5.6}}$ / $\text{36.00}_{\pm\text{9.0}}$ &  $\text{30.50}_{\pm\text{10.8}}$ / $\text{26.00}_{\pm\text{14.0}}$ &   $\text{21.13}_{\pm\text{3.4}}$ / $\text{39.00}_{\pm\text{7.0}}$ &   $\text{18.16}_{\pm\text{2.3}}$ / $\text{46.00}_{\pm\text{5.0}}$ &    $\text{22.71}_{\pm\text{5.4}}$ / $\text{36.75}_{\pm\text{8.3}}$ \\
ICL + CC    &  $\text{22.92}_{\pm\text{6.1}}$ / $\text{41.00}_{\pm\text{12.0}}$ &   $\text{24.24}_{\pm\text{3.8}}$ / $\text{39.00}_{\pm\text{13.0}}$ &  $\text{24.58}_{\pm\text{5.3}}$ / $\text{41.00}_{\pm\text{13.0}}$ &  $\text{21.68}_{\pm\text{3.6}}$ / $\text{45.00}_{\pm\text{12.0}}$ &    $\text{23.36}_{\pm\text{1.3}}$ / $\text{41.50}_{\pm\text{2.5}}$ \\
FT   &   $\text{43.08}_{\pm\text{5.1}}$ / $\text{50.39}_{\pm\text{9.5}}$ &    $\text{51.56}_{\pm\text{3.6}}$ / $\text{42.59}_{\pm\text{2.8}}$ &  $\textbf{50.15}_{\pm\textbf{3.7}}$ / $\text{38.64}_{\pm\text{13.7}}$ &   $\text{39.82}_{\pm\text{3.4}}$ / $\text{51.57}_{\pm\text{9.2}}$ &    $\text{46.15}_{\pm\text{5.6}}$ / $\text{45.80}_{\pm\text{6.2}}$ \\
PT &   $\textbf{49.99}_{\pm\textbf{3.8}}$ / $\text{46.38}_{\pm\text{3.8}}$ &    $\textbf{53.01}_{\pm\textbf{3.1}}$ / $\text{43.69}_{\pm\text{2.3}}$ &   $\text{46.65}_{\pm\text{3.4}}$ / $\text{49.72}_{\pm\text{3.5}}$ &   $\textbf{42.80}_{\pm\textbf{2.5}}$ / $\text{53.06}_{\pm\text{2.5}}$ &    $\textbf{48.11}_{\pm\textbf{4.4}}$ / $\text{48.21}_{\pm\text{4.1}}$ \\
N2C2         &    $\text{46.54}_{\pm\text{6.3}}$ / $\textbf{7.89}_{\pm\textbf{4.0}}$ &     $\text{46.07}_{\pm\text{2.7}}$ / $\textbf{6.42}_{\pm\textbf{2.3}}$ &    $\text{42.77}_{\pm\text{5.0}}$ / $\textbf{8.20}_{\pm\textbf{4.6}}$ &    $\text{40.41}_{\pm\text{3.9}}$ / $\textbf{9.00}_{\pm\textbf{3.4}}$ &     $\text{43.95}_{\pm\text{2.9}}$ / $\textbf{7.88}_{\pm\textbf{1.1}}$ \\
\bottomrule
\end{tabular}
    \caption{    Experimental results of cross-domain condition.  Results are reported as $\textbf{Accuracy}\uparrow(\%)$ / $\textbf{ECE}\downarrow(\%)$. For N2C2, We use the DVD domain as the training and validation set, while for other baselines, we use the book domain as the training and validation set. The test set is uniformly composed of data from the \textbf{Book} domain. }

    \label{tlcls_book}
\end{table*}

\end{document}